\documentclass[twoside]{article}

\usepackage{graphicx}
\usepackage{subfigure}
\usepackage{booktabs}       

\usepackage{latexsym}
\usepackage{amsfonts}       
\usepackage{amsmath}
\usepackage{amssymb}
\usepackage{adjustbox}
\usepackage{multirow}
\usepackage{nicefrac}       
\usepackage{microtype}      
\usepackage[T1]{fontenc}    
\usepackage[utf8]{inputenc} 
\usepackage{comment}
\usepackage{float}
\usepackage{makecell}
\usepackage{appendix}
\usepackage[hyphens]{url}   
\usepackage[dvipsnames]{color,colortbl,xcolor}
\usepackage{caption}
\usepackage{epsfig}
\usepackage{pifont}
\usepackage{xspace}
\usepackage{multicol}

\definecolor{Gray}{gray}{0.9}
\newcommand{\model}{SoftER Teacher\xspace}
\newcommand{\xmark}{\ding{55}}
\newcommand{\cmark}{\ding{51}}
\newcommand{\ie}{\emph{i.e.}}
\newcommand{\eg}{\emph{e.g.}}
\newcommand{\vs}{\emph{vs.}}

\usepackage[pagebackref,breaklinks,bookmarks=false]{hyperref}
\hypersetup{
    colorlinks=true,
    linkcolor=red,
    citecolor=Blue,
    urlcolor=Green
}
\usepackage[capitalize]{cleveref}
\crefname{section}{Sec.}{Secs.}
\Crefname{section}{Section}{Sections}
\Crefname{table}{Table}{Tables}
\crefname{table}{Tab.}{Tabs.}

\usepackage[accepted]{aistats2024}
\usepackage[round]{natbib}

\begin{document}

%

%

\twocolumn[

\aistatstitle{LEDetection: A Simple Framework for Semi-Supervised Few-Shot Object Detection}

\aistatsauthor{ Phi Vu Tran }

\aistatsaddress{ LexisNexis Risk Solutions } ]

\begin{abstract}
  Few-shot object detection (FSOD) is a challenging problem aimed at detecting novel concepts from few exemplars. Existing approaches to FSOD all assume abundant base labels to adapt to novel objects. This paper studies the new task of \emph{semi-supervised FSOD} by considering a realistic scenario in which both base and novel labels are simultaneously scarce. We explore the utility of unlabeled data within our proposed label-efficient detection framework and discover its remarkable ability to boost semi-supervised FSOD by way of region proposals. Motivated by this finding, we introduce SoftER Teacher, a robust detector combining pseudo-labeling with consistency learning on region proposals, to harness unlabeled data for improved FSOD without relying on abundant labels. Rigorous experiments show that SoftER Teacher surpasses the novel performance of a strong supervised detector using only 10\% of required base labels, without catastrophic forgetting observed in prior approaches. Our work also sheds light on a potential relationship between semi-supervised and few-shot detection suggesting that a stronger semi-supervised detector leads to a more effective few-shot detector.
\end{abstract}

\section{INTRODUCTION}\label{sec::intro}

\begin{figure}[ht]
\centering
\includegraphics[width=0.9\columnwidth]{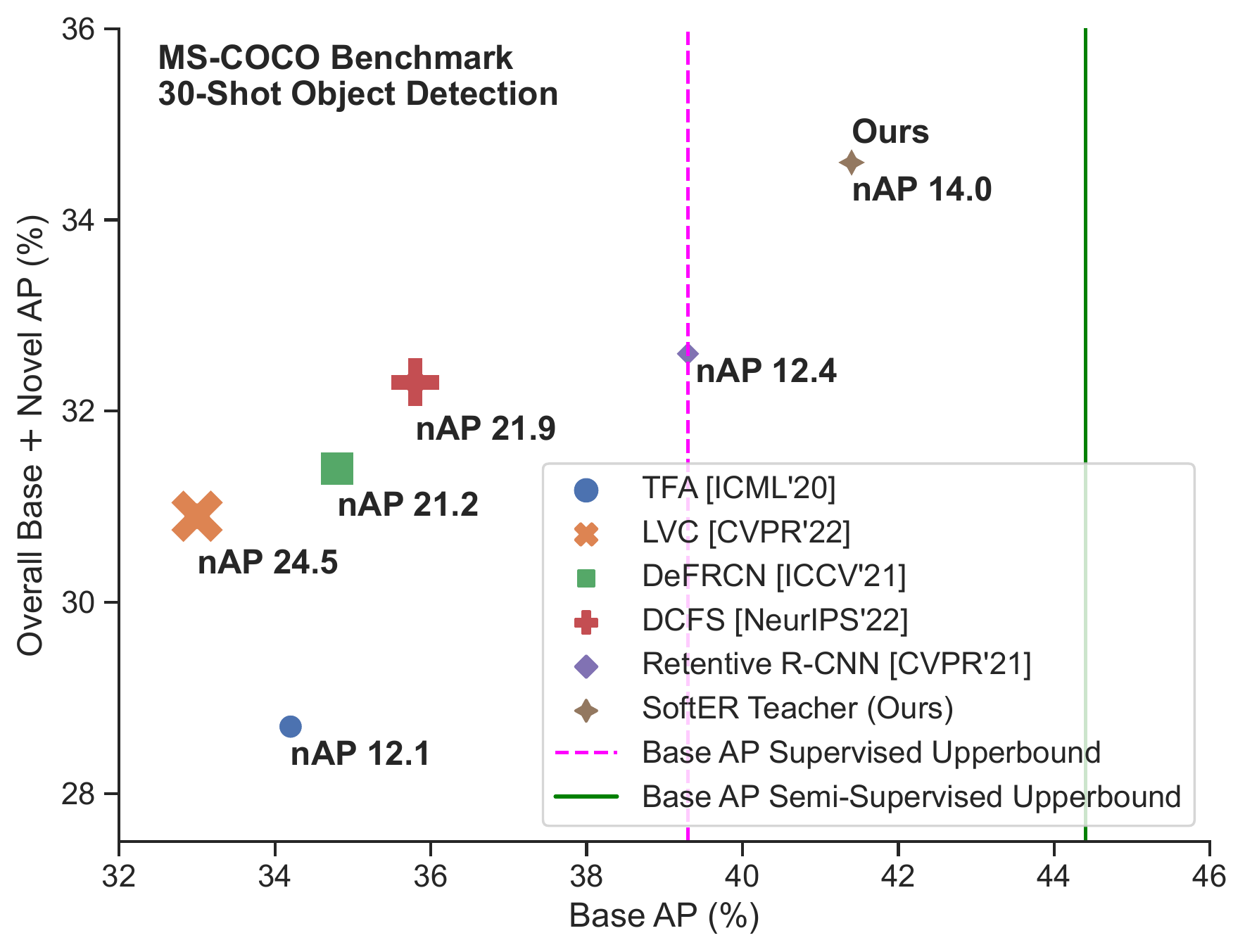}
\captionof{figure}{The evaluation of \emph{generalized} FSOD is characterized by the trade-off between novel performance and base forgetting. We leverage unlabeled data to optimize for semi-supervised FSOD on both base $+$ novel classes (top right). Our approach significantly expands base class AP, $39.3 \rightarrow 44.4$, while incurring less than 9\% in base degradation (\vs~19\% for LVC) and also improving on novel detection (nAP). Our \model is the best model on the \texttt{Overall AP} metric, leading the next best Retentive R-CNN by $+2.0$ AP.}
\label{fig::teaser}
\end{figure}
\begin{figure*}[ht]
\centering
\includegraphics[width=0.82\textwidth]{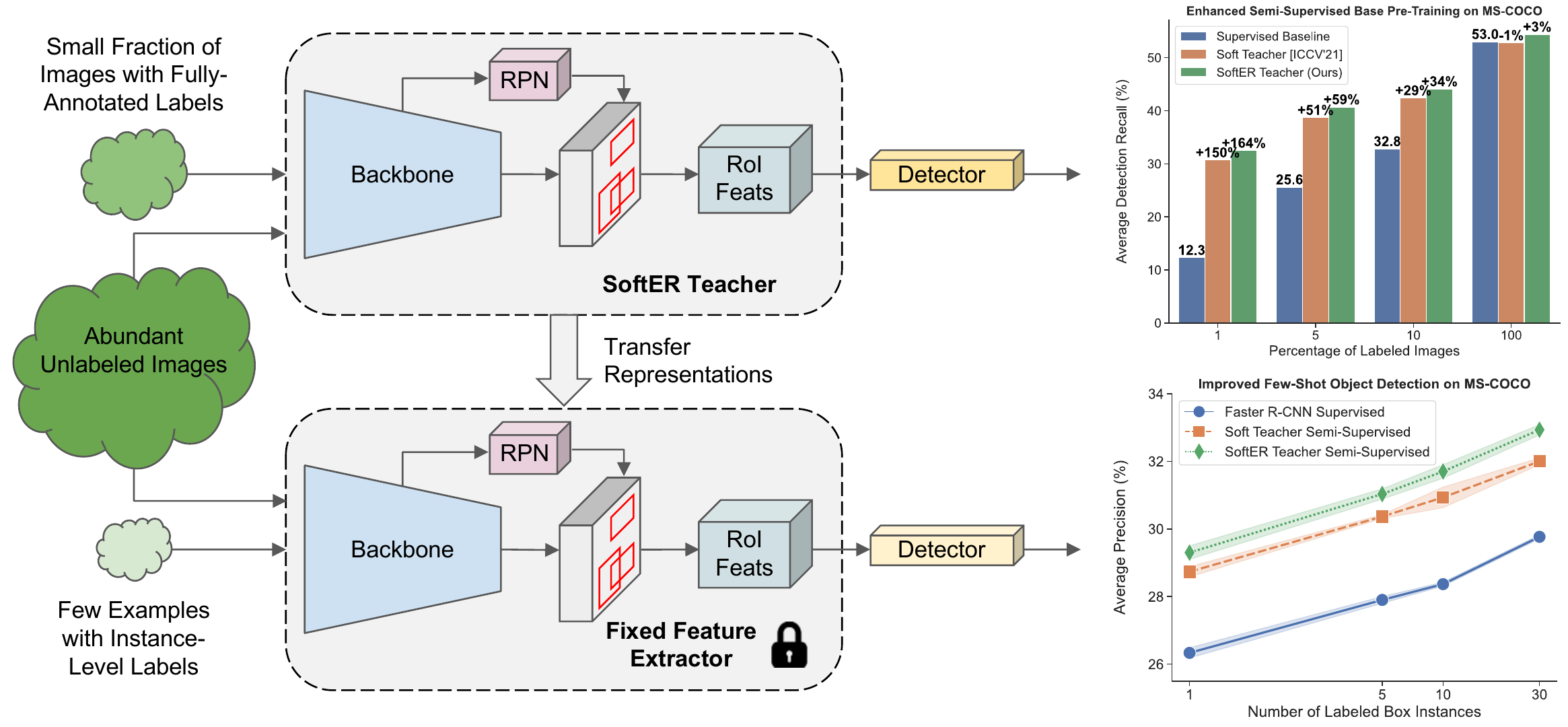}
\caption{We present the Label-Efficient Detection framework to harness supplementary unlabeled data for generalized semi-supervised few-shot detection. At the core of the framework is our proposed \model with Entropy Regression for improved semi-supervised representation learning (\textbf{upper right}). Extensive comparative experiments show that \model is also a more label-efficient few-shot detector (\textbf{lower right}).}
\label{fig::schematic}
\end{figure*}

Modern object detection systems have enjoyed tremendous progress in recent years, with many successful applications across diverse industries. Their success can be mainly attributed to the availability of large-scale, well-annotated datasets such as the MS-COCO benchmark~\citep{coco}. However, the demand for more powerful detection models requires considerable investments in the hand-labeling of massive amounts of data, which are costly to scale. Thus, there is a growing trend in the community to shift toward a more \emph{label-efficient} paradigm, one that can enable detection models to do more with less hand-labeled data. Such emerging research directions include semi-supervised detection (SSOD) and few-shot detection (FSOD), which have shown great promise in alleviating the dependency on large amounts of instance-level annotations.

This paper focuses on the intersection of SSOD and FSOD, which are essentially two sides of the same coin in the context of label-efficient detection. On one side, SSOD investigates the detection problem with a \emph{small fraction of images} containing ground truth labels. On the other side, FSOD addresses the objective of adapting a \emph{base} detector to learn \emph{novel} concepts from \emph{few instance-level} exemplars. Existing approaches to FSOD all assume abundant base classes to train the base detector. However, such assumption is not ideal in practical scenarios where labels may be limited for both base and novel classes, giving rise to the research question: \emph{\textbf{can we advance FSOD given the available unlabeled data without additional hand-labeling?}}

\smallskip

We answer this question by introducing the unique framework of \emph{semi-supervised few-shot detection}, in which we explore the utility of unlabeled data for improving detection with label scarcity for both base and novel classes. Inspired by recent advances in SSOD~\citep{soft-teacher,ubteacherv2,consistent-teacher} and FSOD~\citep{retentive-rcnn,defrcn,dcfs}, our approach is two-fold: (1) we leverage unlabeled data to improve detection with a small fraction of labeled images; and (2) we generalize the resulting semi-supervised detector into a label-efficient few-shot detector by way of transfer learning. Our chief motivation is to not necessarily depend on an abundance of labels for robust few-shot detection, which increases the versatility of our approach in realistic applications.

\smallskip

Moreover, our approach to semi-supervised FSOD adapts a base detector to learn novel concepts with \emph{reduced performance degradation to base classes}, a desirable result missing in most prior approaches. \Cref{fig::teaser} illustrates that while recent work~\citep{defrcn,lvc,dcfs} achieve impressive detection on novel categories, \emph{they all ignore the importance of preserving base class accuracy}. For \emph{generalized} FSOD~\citep{retentive-rcnn}, the goal is to expand the learned vocabulary of the base detector with novel concepts. As such, base and novel class performances are equally important, since samples at test time may contain instances of both objects. Therefore, the more realistic evaluation metric for FSOD is not only novel AP, but the combined base and novel AP, for which our approach establishes a new state of the art.

\smallskip

We measure the utility of unlabeled data within our integrated semi-supervised few-shot framework, and discover an insightful empirical finding linking the effectiveness of unlabeled data to semi-supervised FSOD by way of region proposals. Without bells and whistles, by simply adding unlabeled data to a supervised detector, we show a marked improvement on both base and novel class performances while also mitigating catastrophic base forgetting~\citep{gem}.

\textbf{Main Contributions} ~ First, we introduce a simple and versatile framework to enable semi-supervised FSOD without depending on abundant labels. At the heart of the framework is our proposed \model to combine the strengths of pseudo-labeling with representation learning on unlabeled images, with the unique benefit of enhancing the quality of region proposals to substantially boost semi-supervised FSOD. Moreover, \Cref{sec::effective-fsod} contributes a new empirical analysis on the relationship between unlabeled data and region proposals, extending earlier results on proposal evaluation beyond supervised detection \citep{effective-proposals}.

Second, our work sheds insight into a potential relationship suggesting that a strong semi-supervised detector is also a label-efficient few-shot detector (\Cref{fig::schematic}), an interesting and non-trivial empirical observation linking the two disparate domains. On the task of semi-supervised FSOD, our \model model exceeds the novel class performance of a strong supervised detector~\citep{retentive-rcnn} using less than 10\% of required base labels, while exhibiting less than 9\% in base forgetting. When trained on 100\% of available labels with supplementary unlabeled data, \model sets a new standard on semi-supervised few-shot performance given varying amounts of bounding box annotations.

Third, we establish the Label-Efficient Detection benchmark to quantify the utility of unlabeled data for generalized semi-supervised FSOD. We hope that our benchmark serves as a strong baseline, and a blueprint, to inspire future research toward this new problem setting in the community.

\section{RELATED WORK}
\paragraph{Semi-Supervised Detection}
The current state of the art on SSOD belongs to a family of pseudo-labeling methods, which trains a pair of detectors on pseudo labels along with human labels in the student-teacher framework~\citep{mean-teacher}. One such method is Soft Teacher~\citep{soft-teacher} which vastly improves upon its counterparts STAC~\citep{stac} and Unbiased Teacher~\citep{ubteacher} by enabling end-to-end pseudo-labeling on unlabeled images. In these methods, the teacher model is an exponential moving average (EMA) of its student counterpart and is used to predict pseudo labels on unlabeled data.

We extend the strong performance of Soft Teacher by incorporating a new module for Entropy Regression to learn additional representations from unlabeled images by way of region proposals. Our model, aptly named \emph{\model}, combines the attractive benefits of pseudo-labeling with supplementary proposal learning to establish a stronger baseline for SSOD.

\paragraph{Few-Shot Detection}
The simple yet effective two-stage transfer learning approach following TFA~\citep{tfa} is currently leading the FSOD literature, which comprises an initial stage of base class pre-training followed by a second stage of novel category fine-tuning. While recent two-stage methods, such as DeFRCN~\citep{defrcn} and DCFS~\citep{dcfs}, extend TFA to achieve impressive performance on novel categories, they suffer from considerable base class forgetting, making them sub-optimal in real-world applications requiring accurate detection on test samples containing instances of both classes. Conversely, the methods of Retentive R-CNN~\citep{retentive-rcnn} and DiGeo~\citep{digeo} have proven successful in preserving base class performance, without forgetting, but they both have much room for improvement with regards to novel class detection.

This work takes an important step toward balancing the base-novel performance trade-off by incorporating unlabeled images into the two-stage learning procedure. Our semi-supervised approach demonstrates that both base and novel performances can be further improved, with reduced base degradation, leading to a new standard on multiple challenging FSOD benchmarks.

\paragraph{Semi-Supervised Few-Shot Detection}
There have been few attempts at leveraging unlabeled data to improve FSOD, but to our knowledge none directly addressed the task of optimizing for semi-supervised FSOD, in which setting both base and novel labels are simultaneously scarce. LVC~\citep{lvc} and MINI~\citep{mini} mine novel targets from the \emph{base training dataset} via pseudo-labeling to boost novel class detection, but come at the cost of base performance. UniT~\citep{unit} obtains impressive results on any-shot detection, but assumes abundant image-level labels for the base and novel targets. And SSFOD~\citep{ssfod} performs semi-supervised FSOD within a complicated episodic meta training and $N$-way $k$-shot evaluation framework~\citep{repmet} while also requiring abundant base classes. Our approach is fundamentally different in that we do not strictly depend on abundant base labels, but make effective utilization of \emph{external unlabeled data}, for robust semi-supervised FSOD with reduced base degradation.

\begin{figure*}[t]
  \centering
  \setlength\tabcolsep{1.0pt}
  \begin{tabular}{ccc}
    \makecell{\includegraphics[width=0.47\textwidth]{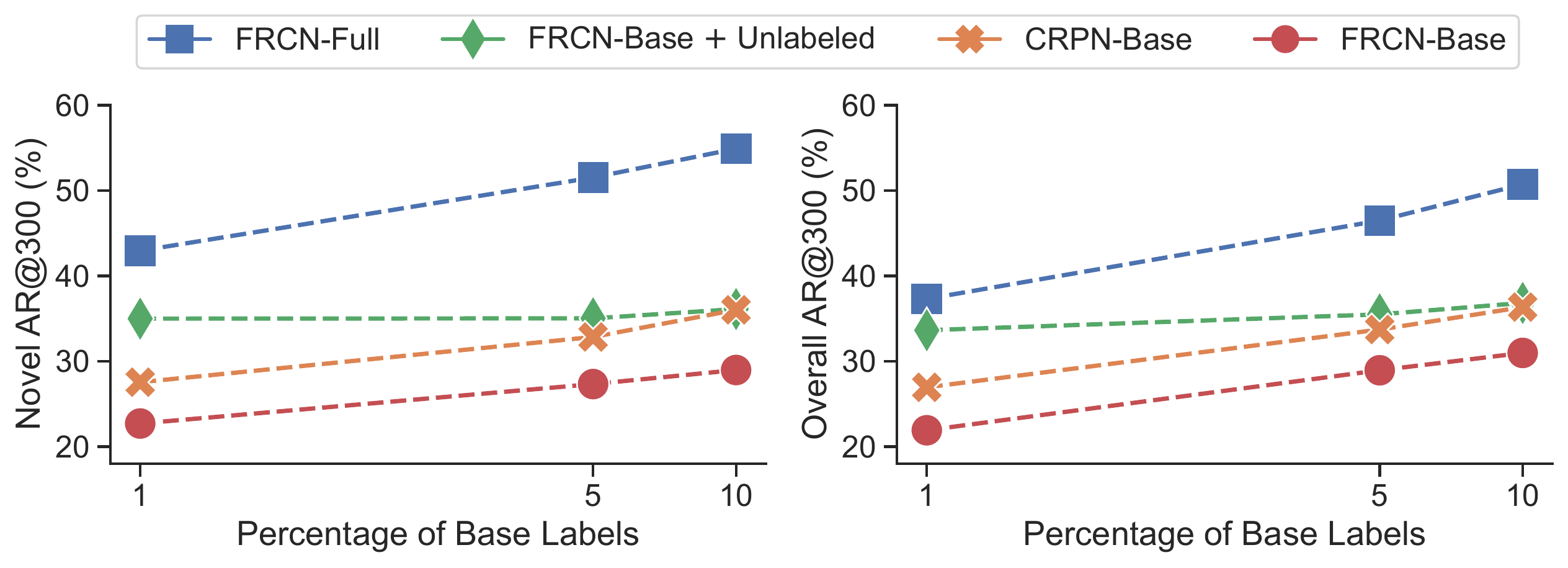}}&
    \makecell{\includegraphics[width=0.24\textwidth]{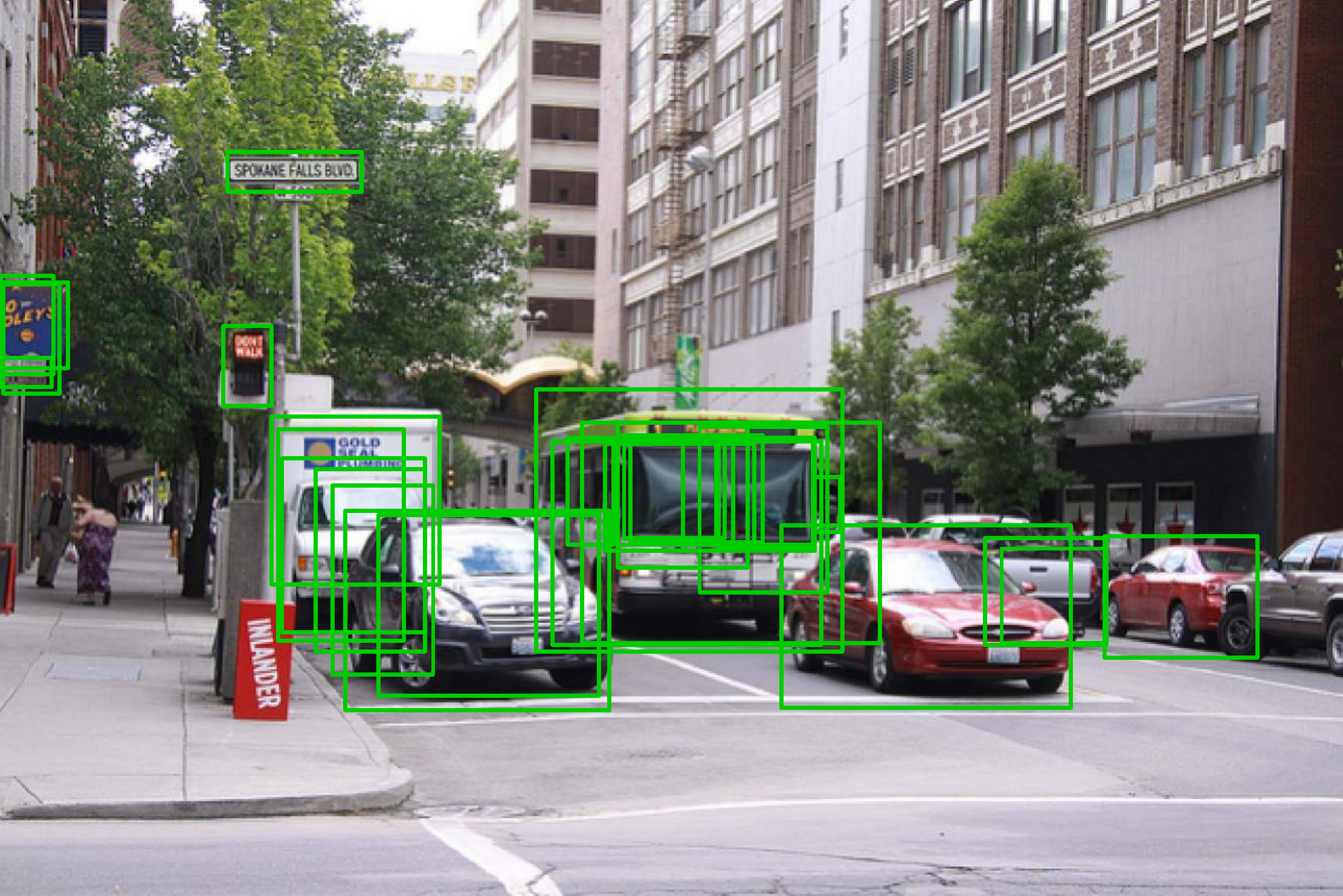}}&
    \makecell{\includegraphics[width=0.231\textwidth]{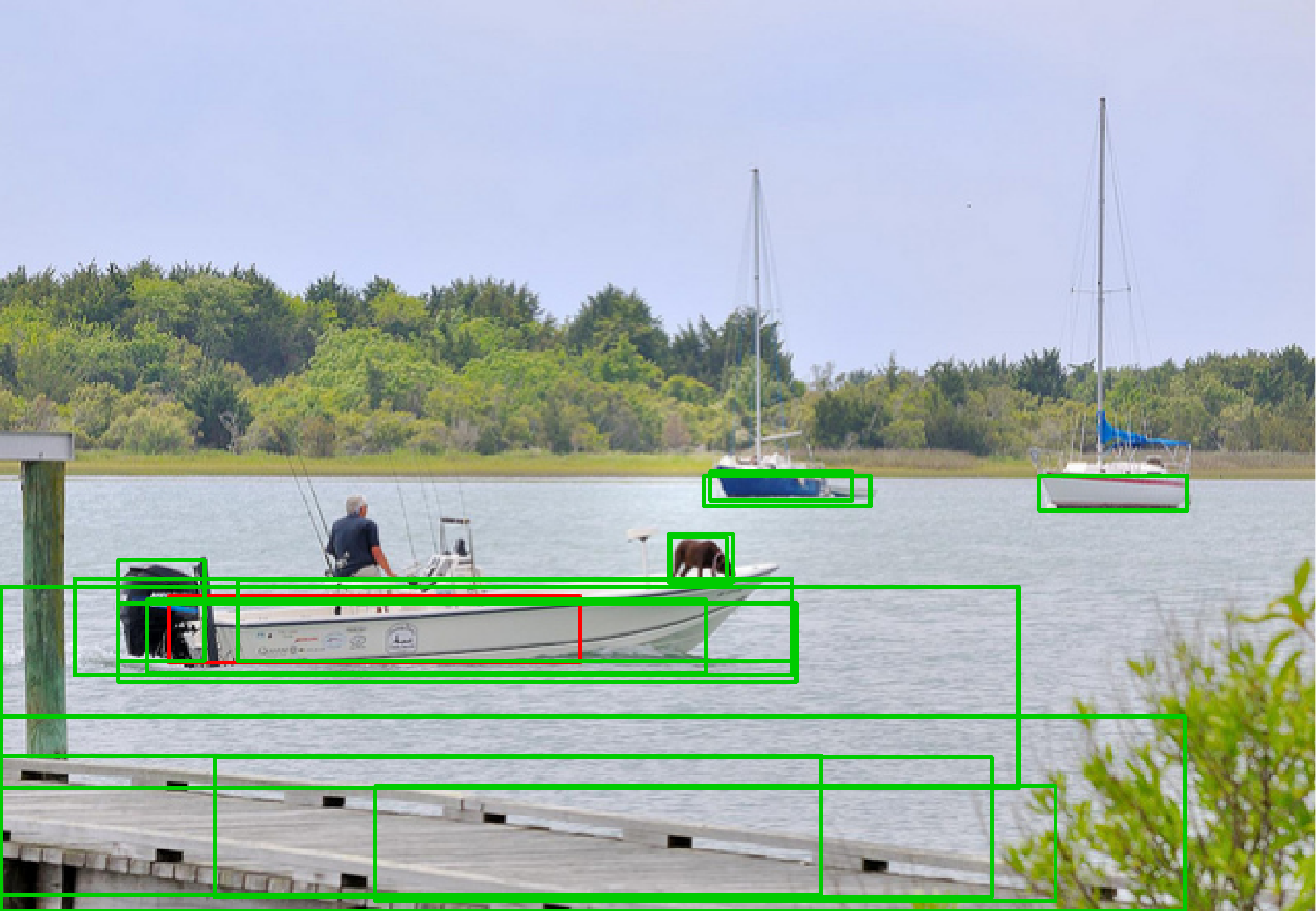}}\\
    \small \textbf{(a)} The impact of unlabeled data on proposal quality. & \small \textbf{(b)} RPN with 1\% labels. & \small \textbf{(c)} RPN with 10\% labels.
  \end{tabular}
  \caption{We analyze the effectiveness of the RPN as a function of base labels. \textbf{(a)} Unlabeled data provide a convincing boost in proposal quality, closing the gap between the \texttt{Base} and \texttt{Full} detectors, which should lead to better discovery of novel categories during fine-tuning. \textbf{(b--c)} In low-label regimes, unlabeled data can help produce diverse proposals (green boxes) on novel unseen objects \{\texttt{boat}, \texttt{bus}, \texttt{car}, \texttt{dog}\}, whereas the vanilla supervised \texttt{FRCN-Base} fails to capture comparable foreground objects with only one red box. Best viewed digitally.}
  \label{fig::rpn-analysis}
\end{figure*}

\section{APPROACH}\label{sec::approach}
We propose to combine the available (limited) labeled examples with supplementary unlabeled images to boost semi-supervised FSOD. We begin with the fully supervised scenario in which we have access to a set of labeled image-target pairs $(x_l, y_l) \in \mathcal{D}_\text{sup}$. The supervised FSOD setting assumes an abundant base dataset $C_\text{base} \in \mathcal{D}_\text{sup}$ to be fully annotated with bounding boxes covering all instances of interest and a novel dataset $C_\text{novel} \in \mathcal{D}_\text{sup}$ with only a few $k$ (\eg, $k \in \{1, 5, 10\}$) randomly labeled instances, or ``shots'', per category. The goal of FSOD is to expand the base detector by adapting it to learn new target concepts such that the resulting detector is optimized for accuracy on a test set comprising both classes in $C_\text{base} \cup C_\text{novel}$.

To maintain parity with the current state of the art, we adopt the two-stage learning procedure consisting of an initial stage of base class pre-training followed by a second stage of novel category fine-tuning. We consider the modern Faster R-CNN (FRCN)~\citep{faster-rcnn} system as our supervised base detector, which consists of a ResNet~\citep{resnet} \emph{backbone} and a feature pyramid network (FPN)~\citep{fpn} \emph{neck} for feature extraction, a region proposal network (RPN), an RoIAlign~\citep{mask-rcnn} operation for mapping proposals to region-of-interest (RoI) features, and a fully-connected \emph{head} for RoI classification and regression. Let FRCN be represented by $f_\theta$, a stochastic function parametrized by a set of learnable weights $\theta$. Formally, the base pre-training step is subjected to the standard supervised objective, over a mini-batch of labeled examples $b_l$, given by:
\begin{equation}
\label{eq::supervised}
    \mathcal{L}_\text{sup} =
        \frac{1}{|b_l|}\sum_{i \in b_l}
        \mathcal{L}^\text{roi}_\text{cls}(f_\theta(x_i), y_i) +
        \mathcal{L}^\text{roi}_\text{reg}(f_\theta(x_i), y_i).
\end{equation}
Here, $f_\theta(x_i)$ denotes a forward pass on the $i$-th image to produce box classification and localization predictions from class-agnostic proposals, $y_i$ is the $i$-th ground truth annotation containing box labels and coordinates, and $\left(\mathcal{L}_\text{cls}^\text{roi}, \mathcal{L}_\text{reg}^\text{roi}\right)$ are the cross-entropy and $L_1$ losses, respectively, for the RoI head. Henceforth for simplicity, we develop our approach only on the RoI head and omit the presentation on the losses of the RPN, which remain constant without changes, to predict and localize the ``objectness'' of region proposals.

\subsection{What Makes for Effective FSOD?}\label{sec::effective-fsod}
We examine this question from the perspective of maximizing representation learning while minimizing base forgetting. In two-stage detectors (\eg, Faster R-CNN), the quality of region proposals is a strong predictor of supervised detection performance~\citep{effective-proposals}, since they focus the detector head on candidate RoIs. This is especially true for FSOD approaches based on transfer learning, in which the established procedure is to freeze the RPN during few-shot fine-tuning. Intuitively, if we can incorporate methods and/or data to boost representation learning by way of the RPN, then the detector should have a higher chance of discovering novel categories to improve few-shot performance.

We perform a motivating empirical analysis on the COCO dataset to verify our intuition. We split the dataset into disjoint 60 base and 20 novel categories and pre-train three variants of the FRCN detector on the base classes: (i) FRCN-Base, (ii) FRCN-Base + Unlabeled, which is augmented with \texttt{COCO-unlabeled2017} images leveraging the Soft Teacher formulation, and (iii) FRCN-Full using both base and novel classes to represent the upper-bound performance. We also experiment with CRPN-Base, a method specifically designed to improve proposal quality and detection performance using a two-stage Cascade RPN~\citep{crpn}.

\Cref{fig::rpn-analysis}a quantifies the ``class-agnosticism'' of various RPNs, using the standard metric \texttt{AR@300} proposals, given varying fractions of base labels. Surprisingly, unlabeled data have the remarkable ability to boost proposal recall on novel-only categories, even in the extremely low 1\% label limit. Somewhat unsurprising is the ability of \texttt{CRPN-Base} to propose novel objects competitive with \texttt{FRCN-Base $+$ Unlabeled} when more labels are available. Consistent with previous findings~\citep{fsce}, Figures~\ref{fig::rpn-analysis}b--c show the vanilla supervised \texttt{FRCN-Base} has a strong tendency to reject novel objects as background, due to the lack of annotations, resulting in the worst recall on novel classes.

\smallskip

As alluded in \Cref{sec::intro}, the contribution of unlabeled data to FSOD can also help mitigate catastrophic base forgetting. We find analogous effectiveness of \texttt{FRCN-Base $+$ Unlabeled} on the combined \texttt{Overall AR@300} metric, for both base $+$ novel objects, suggesting the RPN, when trained with unlabeled data, has the ability to retain base proposals and help combat base degradation during few-shot fine-tuning.

\paragraph{Discussion}
This paper rethinks a different and more versatile way to improve the RPN for FSOD while avoiding catastrophic forgetting. The previous LVC approach proposed to unfreeze the RPN during fine-tuning to obtain large performance gains on novel classes, but comes at the cost of significant base degradation (up to 19\%). Similarly, FSCE~\citep{fsce} proposed to unfreeze the RPN while also doubling the number of proposals to encourage novel foreground detection during fine-tuning. However, this method increases the detection overhead and remains unclear whether it helps mitigate base forgetting. We illustrate that simply adding unlabeled data to the base detector leads to a compelling boost in proposal quality, without the need for any \emph{ad hoc} modifications to the RPN.

We attribute this unique advantage of our approach to the potential base-novel object interactions found in abundant images. When learning with unlabeled data, the base detector can obtain semantically similar cues of novel objects to inform the RPN on foreground detection. \citet{fsce} showed that visually analogous objects have high cosine similarity scores (\eg, $\text{sim}(\texttt{cow}, \texttt{horse}) = 0.39$). With 1\% of labels, these base-novel interactions are limited, resulting in a recall of $22.7\%$. Given a sizable unlabeled dataset, the base detector improves its representations to yield a large gain of $+12.3$ AR points. Extensive experiments in \Cref{sec::proposal-quality} demonstrate that proposal recall indeed has a strong correlation with FSOD performance.

\subsection{Semi-Supervised Base Pre-Training}\label{sec::base-pretrain}
Encouraged by the promising utility of unlabeled data, we now relax the strict assumption on having abundant base classes for FSOD and introduce a more general setting of having a small fraction of base labels given abundant unlabeled images. We approach the task of semi-supervised base pre-training by formulating an unsupervised loss computed on an unlabeled dataset $\mathcal{D}_\text{unsup}$ to be jointly trained with the supervised loss on $\mathcal{D}_\text{sup}$. We consider the following canonical optimization objective widely adopted as part of the framework for semi-supervised learning~\citep{pea,consistency,dac}:
\begin{equation}
\label{canonical}
    \min_\theta \mathcal{L}_\text{sup} (\mathcal{D}_\text{sup}, \theta) + \lambda\mathcal{L}_\text{unsup}(\mathcal{D}_\text{unsup}, \theta),
\end{equation}
where $\lambda > 0$ is a hyper-parameter controlling the contribution of the unsupervised component. Next, we describe the unsupervised criterion on $\mathcal{D}_\text{unsup}$ to make FRCN into a semi-supervised detector.

\paragraph{Soft Teacher}
We adopt Soft Teacher~\citep{soft-teacher} as the baseline SSOD formulation for its simplicity but strong performance. Soft Teacher trains FRCN in a student-teacher fashion on both labeled and unlabeled data. The student is trained on labeled examples in the standard supervised manner per \cref{eq::supervised}. For unlabeled images, the teacher is treated as a fixed detector to generate thousands of box candidates, most of which are eliminated for redundancy with non-maximum suppression. Additionally, box candidates are thresholded for foreground objects and go through an iterative jittering-refinement procedure to reduce localization variance, resulting in a set of high-quality pseudo boxes to be jointly trained with ground truth annotations.

As is common practice~\citep{fixmatch}, the teacher's parameters $\bar{\theta}$ are updated from the student's via $\bar{\theta} = \text{EMA}(\theta)$ at each training step. Integral to the success of Soft Teacher is a student-teacher data augmentation strategy inspired by STAC~\citep{stac}. The student trains on unlabeled images subjected to complex random perturbations, akin to RandAugment~\citep{randaugment}, including affine transforms. Separately, the teacher receives weakly augmented images with simple random resizing and flipping. This multi-stream augmentation design allows the teacher to generate reliable unsupervised targets on easy images to guide the student's learning trajectory on difficult images for better generalizability.

At the time of box classification and regression in the RoI head, we have a set of unlabeled images along with teacher-generated pseudo labels $(x_u, \hat{y}_u) \in \mathcal{D}_\text{unsup}$. The unsupervised loss for Soft Teacher on a mini-batch of unlabeled images $b_u$ is defined as:
\begin{equation}
\label{eq::unsup_soft}
    \mathcal{L}_\text{soft} =
        \frac{1}{|b_u|}\sum_{i \in b_u}
        \mathcal{L}^\text{roi}_\text{cls}(f_\theta(x_i), \hat{y}_i) +
        \mathcal{L}^\text{roi}_\text{reg}(f_\theta(x_i), \hat{y}_i).
\end{equation}

\paragraph{\model}
The design of Soft Teacher employs class confidence thresholding and box jittering to select high-quality pseudo candidates for unsupervised classification and regression. However, \citet{soft-teacher} showed that it uses an aggressive threshold of 0.9, resulting in a trade-off between low recall and high precision at 33\% and 89\%, respectively. We observe that low recall can result in poor detection performance on small and ambiguous objects~\citep{detnet}, especially in low-label regimes where the teacher has insufficient confidence about its predicted pseudo labels. We aim to extend Soft Teacher and improve its detection recall by learning additional representations from abundant region proposals. We show in \Cref{sec::proposal-quality} that recall is key to an effective few-shot detector.

\begin{figure*}
\centering
\includegraphics[width=0.9\textwidth]{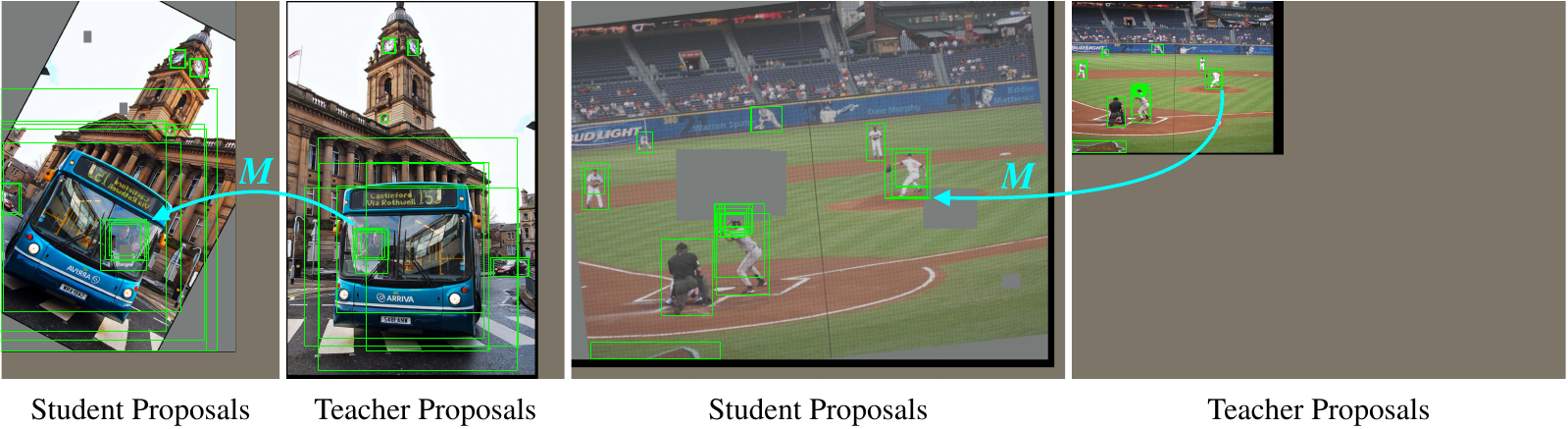}
\caption{Visualizations of student-teacher proposals with confidence scores $\ge 0.99$. As illustrated by the arrow, a pair of student-teacher proposals is related by a transformation matrix $M$, which is used to align proposals between student and teacher images for enforcing box classification similarity and localization consistency.}
\label{fig::multi-views}
\end{figure*}

Given a set of proposals $p$ generated by the student's RPN on a batch of unlabeled images, we apply the student-teacher data augmentation pipeline described above to obtain $(p_s, p_t)$, denoting transformed student and teacher proposals, which are related to each other by a matrix $M$. We then forward pass FRCN twice, as the student $f_\theta$ and teacher $f_{\bar{\theta}}$, to obtain two sets of RoI outputs for predicted box classification logits $(z_s, z_t)$ and localization coordinates $(r_s, r_t)$. Let $g_c$ be the softmax function over the channel dimension $c$. We define an auxiliary unsupervised criterion for proposal box similarity based on a cross-entropy measure $H(z_s, z_t)$:
\begin{equation}\label{eq::cross-entropy-similarity}
\begin{split}
    &\mathcal{L}^\text{ent}_\text{cls} = \frac{1}{\sum_{i}{w_i}}\sum_{i \in p}w_i \cdot H(z_{is}, z_{it}),\\
    \text{where\quad}
    &H(z_{is}, z_{it}) = -\frac{1}{C}\sum_{c \in C}g_c(z_{it}) \log g_c(z_{is}).
\end{split}
\end{equation}
Here, $g_c$ outputs a distribution over $C$ classes and $w_i$ is the Boolean weight for the predicted foreground class: $w_i = 1$ if $\text{argmax}(z_{it}) \neq background$, else $w_i = 0$.

Analogously for proposal box regression, we constrain the predicted box coordinates $(r_s, r_t)$ to be close. Since there are complex geometric distortions between the two, we first map teacher proposal coordinates $r_t$ to the student space using the transformation $M$. Then, we align the proposal boxes via the intersection-over-union (IoU) loss criterion~\citep{giou} to compute their differences:
\begin{equation}\label{eq::iou-regression}
    \mathcal{L}^\text{iou}_\text{reg} = 1 - \frac{1}{|p|}\sum_{i \in p}w_i \cdot \text{IoU}(r_{is}, M(r_{it})).
\end{equation}
Recall that we have two different transformation pipelines operating on each proposal, so we have two augmented views of each proposal. \Cref{fig::multi-views} illustrates that by enforcing these randomly augmented views, and their box coordinates, to be \emph{similar}, we enable the student to tap into abundant region proposals to learn diverse feature representations across a spectrum of scale, color, and geometric perturbations. Our formulation draws inspiration from recent research on self-supervised representation learning with multi-augmented views~\citep{byol,simsiam}. Note the cross-entropy similarity between the student and teacher predictions, \cref{eq::cross-entropy-similarity}, can be interpreted as a form of entropy regularization~\citep{entmin}, which has been proven to work well in various semi-supervised classification scenarios~\citep{vat2,ssleval}. Combining \cref{eq::supervised,eq::unsup_soft,eq::cross-entropy-similarity,eq::iou-regression}, the total optimization objective at the RoI head for our \model model is computed as:
\begin{equation}
\label{eq::final_loss_softer}
    \mathcal{L}_\text{softer}^\text{total} = \mathcal{L}_\text{sup} +
                  \alpha\mathcal{L}_\text{soft} +
                  \beta\left(\mathcal{L}_\text{cls}^\text{ent} + \mathcal{L}_\text{reg}^\text{iou}\right),
\end{equation}
where we set $\alpha = \frac{|b_u|}{|b_l|}$ following Soft Teacher and find $\beta = 2\alpha$ works reliably well across all experiments.

\begin{table*}[ht]
\caption{FSOD results evaluated on COCO \texttt{val2017}. We report the mean and 95\% confidence interval over 5 random samples for our models. \model with ResNet-101 is the best model on the combined \texttt{Overall AP} metric, incurring less than 9\% in base forgetting \vs~11\%--DCFS, 17\%--DeFRCN, 18\%--TFA, and 19\%--LVC.}
\centering
\resizebox{\textwidth}{!}{
    \begin{tabular}{lcc|ccc|ccc|ccc|}
    \toprule
    \textbf{COCO \texttt{val2017}} &
    \multirow{2}{*}{Backbone} &
    \multirow{2}{*}{Base AP$_{50:95}$} &
    \multicolumn{3}{c|}{Base AP$_{50:95}$ (60 Classes)} &
    \multicolumn{3}{c|}{Novel AP$_{50:95}$ (20 Classes)} &
    \multicolumn{3}{c|}{Overall AP$_{50:95}$ (80 Classes)}\\
    \cmidrule{4-12}
    \multirow{1}{*}{Method} &
    & &
    \multirow{1}{*}{5-Shot} &
    \multirow{1}{*}{10-Shot} &
    \multirow{1}{*}{30-Shot} &
    \multirow{1}{*}{5-Shot} &
    \multirow{1}{*}{10-Shot} &
    \multirow{1}{*}{30-Shot} &
    \multirow{1}{*}{5-Shot} &
    \multirow{1}{*}{10-Shot} &
    \multirow{1}{*}{30-Shot}\\
    \midrule
        LVC* [CVPR'22]
        & R-50
        & --
        & ~~--
        & $29.7$
        & $33.3$
        & ~~--
        & \bfseries 17.6
        & \bfseries 25.5
        & ~~--
        & $26.7$
        & $31.4$\\
        \cellcolor{Gray}\model (Ours)
        & \cellcolor{Gray}R-50
        & \cellcolor{Gray}\bfseries 42.0
        & \cellcolor{Gray}\bfseries 37.4 $\pm$ 0.2
        & \cellcolor{Gray}\bfseries 37.4 $\pm$ 0.2
        & \cellcolor{Gray}\bfseries 38.7 $\pm$ 0.2
        & \cellcolor{Gray}\bfseries ~~8.2 $\pm$ 0.3
        & \cellcolor{Gray}$10.3 \pm 0.5$
        & \cellcolor{Gray}$12.9 \pm 0.6$
        & \cellcolor{Gray}\bfseries 30.1 $\pm$ 0.1
        & \cellcolor{Gray}\bfseries 30.7 $\pm$ 0.2
        & \cellcolor{Gray}\bfseries 32.3 $\pm$ 0.1\\
    \midrule
        LVC* [CVPR'22]
        & R-101
        & $39.3$
        & ~~--
        & $31.9$
        & $33.0$
        & ~~--
        & $17.8$
        & \bfseries 24.5
        & ~~--
        & $28.4$
        & $30.9$\\
        TFA [ICML'20]
        & R-101
        & $39.3$
        & $32.3 \pm 0.6$
        & $32.4 \pm 0.6$
        & $34.2 \pm 0.4$
        & $~~7.0 \pm 0.7$
        & $~~9.1 \pm 0.5$
        & $12.1 \pm 0.4$
        & $25.9 \pm 0.6$
        & $26.6 \pm 0.5$
        & $28.7 \pm 0.4$\\
        DeFRCN [ICCV'21]
        & R-101
        & $39.3$
        & $32.6 \pm 0.3$
        & $34.0 \pm 0.2$
        & $34.8 \pm 0.1$
        & $13.6 \pm 0.7$
        & $16.8 \pm 0.6$
        & $21.2 \pm 0.4$
        & $27.8 \pm 0.3$
        & $29.7 \pm 0.2$
        & $31.4 \pm 0.1$\\
        DCFS [NeurIPS'22]
        & R-101
        & $39.3$
        & $35.0 \pm 0.2$
        & $35.7 \pm 0.2$
        & $35.8 \pm 0.2$
        & \bfseries 15.7 $\pm$ 0.5
        & \bfseries 18.3 $\pm$ 0.4
        & $21.9 \pm 0.3$
        & $30.2 \pm 0.2$
        & $31.4 \pm 0.2$
        & $32.3 \pm 0.2$\\
        Retentive R-CNN [CVPR'21]
        & R-101
        & $39.3$
        & $39.3 \pm \text{n/a}$
        & $39.2 \pm \text{n/a}$
        & $39.3 \pm \text{n/a}$
        & $~~7.7 \pm \text{n/a}$
        & $~~9.5 \pm \text{n/a}$
        & $12.4 \pm \text{n/a}$
        & $31.4 \pm \text{n/a}$
        & $31.8 \pm \text{n/a}$
        & $32.6 \pm \text{n/a}$\\
        \cellcolor{Gray}\model (Ours)
        & \cellcolor{Gray}R-101
        & \cellcolor{Gray}\bfseries 44.4
        & \cellcolor{Gray}\bfseries 40.3 $\pm$ 0.2
        & \cellcolor{Gray}\bfseries 40.2 $\pm$ 0.3
        & \cellcolor{Gray}\bfseries 41.4 $\pm$ 0.2
        & \cellcolor{Gray}$~~8.7 \pm 0.6$
        & \cellcolor{Gray}$11.0 \pm 0.4$
        & \cellcolor{Gray}$14.0 \pm 0.6$
        & \cellcolor{Gray}\bfseries 32.4 $\pm$ 0.2
        & \cellcolor{Gray}\bfseries 32.9 $\pm$ 0.1
        & \cellcolor{Gray}\bfseries 34.6 $\pm$ 0.1\\
    \bottomrule
    \multicolumn{12}{l}{* indicates results were reported on a single sample run (hence the lack of error bars), which may have been over-estimated due to the high variance of few-shot training samples.}
    \end{tabular}
}
\label{tab::fsod-sota-comparison}
\end{table*}
\begin{table*}[t]
\caption{FSOD results evaluated on VOC07 \texttt{test}. We report the mean and 95\% confidence interval over 10 samples for our models. \model with ResNet-50 surpasses the supervised MPSR~\citep{mpsr}, TFA, and Retentive R-CNN models with ResNet-101 by a large margin on most metrics under study, while being more parameter-efficient with a smaller backbone. Results for the other two splits are given in \Cref{sec::generalized-voc}.}
\begin{minipage}[t]{\textwidth}
\centering
\resizebox{\columnwidth}{!}{
    \begin{tabular}{lccc|ccc|ccc|ccc|}
    \toprule
    \textbf{VOC07 \texttt{test} -- Split 1} &
    \multirow{2}{*}{Backbone} &
    \multirow{1}{*}{\parbox{0.05\columnwidth}{\vspace{1.5mm}Base}} &
    \multirow{1}{*}{\parbox{0.05\columnwidth}{\vspace{1.5mm}Base}} &
    \multicolumn{3}{c|}{Base AP$_{50}$ (15 Classes)} &
    \multicolumn{3}{c|}{Novel AP$_{50}$ (5 Classes)} &
    \multicolumn{3}{c|}{Overall AP$_{50}$ (20 Classes)}\\
    \cmidrule{5-13}
    \multirow{1}{*}{Method} &
    &
    \multirow{1}{*}{\parbox{0.05\columnwidth}{\vspace{-1.5mm}AP$_{50}$}} &
    \multirow{1}{*}{\parbox{0.05\columnwidth}{\vspace{-1.5mm}AR$_{50}$}} &
    \multirow{1}{*}{1-Shot} &
    \multirow{1}{*}{5-Shot} &
    \multirow{1}{*}{10-Shot} &
    \multirow{1}{*}{1-Shot} &
    \multirow{1}{*}{5-Shot} &
    \multirow{1}{*}{10-Shot} &
    \multirow{1}{*}{1-Shot} &
    \multirow{1}{*}{5-Shot} &
    \multirow{1}{*}{10-Shot} \\
    \midrule
        MPSR* [ECCV'20]
        & R-101
        & $80.8$
        & --
        & $61.5$
        & $69.7$
        & $71.6$
        & \bfseries 42.8
        & $55.3$
        & $61.2$
        & $56.8$
        & $66.1$
        & $69.0$\\
        Retentive R-CNN* [CVPR'21]
        & R-101
        & $80.8$
        & --
        & $80.9$
        & $80.8$
        & $80.8$
        & $42.4$
        & $53.7$
        & $56.1$
        & $71.3$
        & $74.0$
        & $74.6$\\
        TFA [ICML'20]
        & R-101
        & $80.8$
        & --
        & $77.6 \pm 0.2$ 
        & $77.4 \pm 0.3$
        & $77.5 \pm 0.2$
        & $25.3 \pm 2.2$
        & $47.9 \pm 1.2$
        & $52.8 \pm 1.0$
        & $64.5 \pm 0.6$
        & $70.1 \pm 0.4$
        & $71.3 \pm 0.3$\\
        \cellcolor{Gray}{\model (Ours)}
        & \cellcolor{Gray}{R-101}
        & \cellcolor{Gray}{\bfseries 85.0}
        & \cellcolor{Gray}{\bfseries 93.1}
        & \cellcolor{Gray}{\bfseries 83.3 $\pm$ 0.4}
        & \cellcolor{Gray}{\bfseries 84.6 $\pm$ 0.2}
        & \cellcolor{Gray}{\bfseries 84.9 $\pm$ 0.2}
        & \cellcolor{Gray}{$35.5 \pm 3.7$}
        & \cellcolor{Gray}{\bfseries 60.3 $\pm$ 2.8}
        & \cellcolor{Gray}{\bfseries 65.9 $\pm$ 1.6}
        & \cellcolor{Gray}{\bfseries 71.3 $\pm$ 1.2}
        & \cellcolor{Gray}{\bfseries 78.5 $\pm$ 0.8}
        & \cellcolor{Gray}{\bfseries 80.2 $\pm$ 0.4}\\
    \midrule
        Faster R-CNN (Our Impl.)
        & R-50
        & $81.7$
        & $88.0$
        & $82.0 \pm 0.2$
        & $82.4 \pm 0.1$
        & $82.3 \pm 0.1$
        & $27.9 \pm 3.2$
        & $52.1 \pm 2.1$
        & $58.2 \pm 1.6$
        & $68.5 \pm	0.8$
        & $74.9	\pm 0.5$
        & $76.2	\pm 0.4$\\
        Soft Teacher (Our Impl.)
        & R-50
        & \bfseries 85.0
        & $90.9$
        & \bfseries 84.6 $\pm$ 0.4
        & \bfseries 85.2 $\pm$ 0.1
        & \bfseries 85.0 $\pm$ 0.1
        & $29.5 \pm 4.4$
        & $53.8 \pm 2.6$
        & $60.8 \pm 1.5$
        & \bfseries 70.8 $\pm$ 1.2
        & $77.4	\pm 0.7$
        & $79.0	\pm 0.3$\\
        \cellcolor{Gray}{\model (Ours)}
        & \cellcolor{Gray}{R-50}
        & \cellcolor{Gray}{$84.7$}
        & \cellcolor{Gray}{\bfseries 92.3}
        & \cellcolor{Gray}{$83.1 \pm 0.2$}
        & \cellcolor{Gray}{$84.4 \pm 0.2$}
        & \cellcolor{Gray}{$84.6 \pm 0.2$}
        & \cellcolor{Gray}{\bfseries 31.3 $\pm$ 4.3} 
        & \cellcolor{Gray}{\bfseries 57.8 $\pm$ 2.6}
        & \cellcolor{Gray}{\bfseries 62.9 $\pm$ 1.7}
        & \cellcolor{Gray}{$70.2 \pm 1.2$}
        & \cellcolor{Gray}{\bfseries 77.8 $\pm$ 0.7}
        & \cellcolor{Gray}{\bfseries 79.2 $\pm$ 0.4}\\
    \bottomrule
    \multicolumn{13}{l}{* indicates results were reported on a single sample run (hence the lack of error bars), which may have been over-estimated due to the high variance of few-shot training samples.}
    \end{tabular}
}
\end{minipage}
\label{tab::fsod-voc}
\end{table*}

\subsection{Semi-Supervised Few-Shot Fine-Tuning}\label{sec::novel-finetune}
We propose a simple two-step approach to harness unlabeled data for semi-supervised few-shot fine-tuning. First, we initialize the few-shot detector, $f^\prime_{\bar{\theta}} \leftarrow f_{\bar{\theta}}$, with parameters copied from the base \emph{teacher} detector pre-trained with unlabeled data per \cref{eq::final_loss_softer}. And second, we further train the RoI head of $f^\prime_{\bar{\theta}}$ on novel classes using the available few-shot and unlabeled examples while freezing the base backbone, FPN, and RPN components. Then, we fine-tune the few-shot detector on a balanced training set of $k$ shots per class containing both base and novel instances. We update only the RoI box classifier while freezing all other components, including the box regressor, since it is the main source of error~\citep{fsce}. Rigorous experiments show that our approach provides a compelling boost to novel performance while enjoying substantial gains on base accuracy with reduced base forgetting. We present detailed ablation studies in \Cref{sec::ablation,sec::freeze-box-regressor} to validate our approach and design choices.

\section{EXPERIMENTS}

\begin{figure*}[t]
\centering
\includegraphics[width=0.92\textwidth]{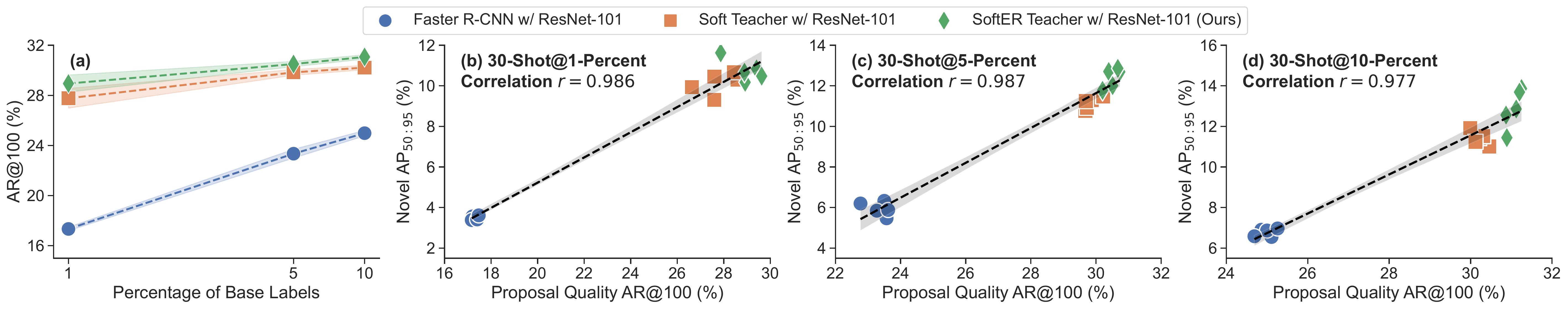}
\caption{Proposal quality is highly correlated with semi-supervised FSOD. \model produces the best proposals among the comparisons \textbf{(a)}, which yield the strongest 30-shot performances \textbf{(b--d)}. Shaded regions denote standard deviation over 5 samples. \Cref{sec::generalized-proposal-quality} gives similar trends for 5-shot and 10-shot results.}
\label{fig::ar-vs-ap}
\end{figure*}
\begin{table*}[t]
\caption{We introduce the Label-Efficient Detection benchmark for generalized semi-supervised FSOD on COCO \texttt{val2017}. All models are equipped with ResNet-101. We report the mean and standard deviation over 5 samples. Using only 10\% of base labels \textbf{(bottom row)}, \model surpasses the supervised novel performance of Retentive R-CNN trained with 100\% of base labels \textbf{(top row)} while incurring less than 9\% in base degradation.}
\centering
\resizebox{\textwidth}{!}{
    \begin{tabular}{lrcc|rrr|rrr|rrr|}
    \toprule
    \multirow{2}{*}{Method} &
    \multirow{2}{*}{\% Labeled} &
    \multirow{2}{*}{Base AP$_{50:95}$} &
    \multirow{2}{*}{Base AR$_{50:95}$} &
    \multicolumn{3}{c|}{Base AP$_{50:95}$ (60 Classes)} &
    \multicolumn{3}{c|}{Novel AP$_{50:95}$ (20 Classes)} &
    \multicolumn{3}{c|}{Overall AP$_{50:95}$ (80 Classes)} \\
    \cmidrule{5-13}
    & & & &
    \multicolumn{1}{c}{\multirow{1}{*}{5-Shot}} &
    \multicolumn{1}{c}{\multirow{1}{*}{10-Shot}} &
    \multicolumn{1}{c|}{\multirow{1}{*}{30-Shot}} &
    \multicolumn{1}{c}{\multirow{1}{*}{5-Shot}} &
    \multicolumn{1}{c}{\multirow{1}{*}{10-Shot}} &
    \multicolumn{1}{c|}{\multirow{1}{*}{30-Shot}} &
    \multicolumn{1}{c}{\multirow{1}{*}{5-Shot}} &
    \multicolumn{1}{c}{\multirow{1}{*}{10-Shot}} &
    \multicolumn{1}{c|}{\multirow{1}{*}{30-Shot}} \\
    \midrule
    Retentive R-CNN
        & \multirow{2}{*}{100}
        & $39.3$
        & ~~--
        & \multicolumn{1}{c}{$39.3$}
        & \multicolumn{1}{c}{$39.2$}
        & \multicolumn{1}{c|}{$39.3$}
        & \multicolumn{1}{c}{$7.7$}
        & \multicolumn{1}{c}{~~$9.5$}
        & \multicolumn{1}{c|}{$12.4$}
        & \multicolumn{1}{c}{$31.4$}
        & \multicolumn{1}{c}{$31.8$}
        & \multicolumn{1}{c|}{$32.6$}\\
    \model
        &
        & $44.4$
        & $56.1$
        & $40.3 \pm 0.2$
        & $40.2 \pm 0.3$
        & $41.4 \pm 0.2$
        & $8.7 \pm 0.6$
        & $11.0 \pm 0.4$
        & $14.0 \pm 0.6$
        & $32.4 \pm 0.2$
        & $32.9 \pm 0.1$
        & $34.6 \pm 0.1$\\
    \midrule
    Faster R-CNN
        & \multirow{3}{*}{1}
        & ~~$8.7 \pm 0.3$
        & $12.3 \pm 0.5$
        & $9.8 \pm 0.3$
        & $10.0 \pm 0.4$
        & $10.8 \pm 0.3$
        & $1.9 \pm 0.3$
        & $2.7 \pm 0.1$
        & $3.5 \pm 0.1$
        & $7.8 \pm 0.2$
        & $8.2 \pm 0.3$
        & $9.0 \pm 0.2$\\
    Soft Teacher
        &
        & $19.9 \pm 1.0$
        & $30.7 \pm 1.1$
        & $19.4 \pm 0.7$
        & $19.9 \pm 0.8$
        & $21.2 \pm 0.7$
        & $5.9 \pm 0.8$
        & $7.9 \pm 0.7$
        & $10.1 \pm 0.5$
        & $16.0 \pm 0.6$
        & $16.9 \pm 0.7$
        & $18.4 \pm 0.6$\\
    \cellcolor{Gray}{\model}
        & \cellcolor{Gray}
        & \cellcolor{Gray}$19.8 \pm 0.9$
        & \cellcolor{Gray}$32.5 \pm 1.0$
        & \cellcolor{Gray}$19.2 \pm 0.6$
        & \cellcolor{Gray}$19.8 \pm 0.5$
        & \cellcolor{Gray}$21.1 \pm 0.5$
        & \cellcolor{Gray}$6.7 \pm 0.3$
        & \cellcolor{Gray}$8.8 \pm 0.2$
        & \cellcolor{Gray}$10.8 \pm 0.5$
        & \cellcolor{Gray}$16.1 \pm 0.5$
        & \cellcolor{Gray}$17.1 \pm 0.4$
        & \cellcolor{Gray}$18.5 \pm 0.5$\\
    \midrule
    Faster R-CNN
        & \multirow{3}{*}{5}
        & $19.1 \pm 0.3$
        & $25.6 \pm 0.4$
        & $18.5 \pm 0.5$
        & $18.9 \pm 0.3$
        & $20.0 \pm 0.5$
        & $3.5 \pm 0.2$
        & $4.6 \pm 0.2$
        & $5.9 \pm 0.3$
        & $14.8 \pm 0.4$
        & $15.3 \pm 0.2$
        & $16.5 \pm 0.4$\\
    Soft Teacher
        &
        & $29.6 \pm 0.3$
        & $38.7 \pm 0.3$
        & $27.5 \pm 0.4$
        & $27.8 \pm 0.5$
        & $29.2 \pm 0.5$
        & $6.7 \pm 0.7$
        & $8.9 \pm 0.4$
        & $11.1 \pm 0.3$
        & $22.3 \pm 0.4$
        & $23.1 \pm 0.3$
        & $24.7 \pm 0.4$\\
    \cellcolor{Gray}{\model}
        & \cellcolor{Gray}
        & \cellcolor{Gray}$30.2 \pm 0.2$
        & \cellcolor{Gray}$40.7 \pm 0.3$
        & \cellcolor{Gray}$27.5 \pm 0.4$
        & \cellcolor{Gray}$27.9 \pm 0.4$
        & \cellcolor{Gray}$29.3 \pm 0.2$
        & \cellcolor{Gray}$7.9 \pm 0.4$
        & \cellcolor{Gray}$10.1 \pm 0.5$
        & \cellcolor{Gray}$12.4 \pm 0.5$
        & \cellcolor{Gray}$22.6 \pm 0.3$
        & \cellcolor{Gray}$23.4 \pm 0.3$
        & \cellcolor{Gray}$25.1 \pm 0.2$\\
    \midrule
    Faster R-CNN
        & \multirow{3}{*}{10}
        & $24.7 \pm 0.2$
        & $32.8 \pm 0.3$
        & $22.6 \pm 0.4$
        & $22.8 \pm 0.1$
        & $24.2 \pm 0.2$
        & $3.8 \pm 0.5$
        & $5.3 \pm 0.2$
        & $6.8 \pm 0.2$
        & $17.9 \pm 0.3$
        & $18.4 \pm 0.1$
        & $19.9 \pm 0.2$\\
    Soft Teacher
        &
        & $33.3 \pm 0.2$
        & $42.4 \pm 0.2$
        & $30.5 \pm 0.5$
        & $30.7 \pm 0.4$
        & $32.1 \pm 0.3$
        & $6.8 \pm 0.3$
        & $9.0 \pm 0.6$
        & $11.4 \pm 0.3$
        & $24.6 \pm 0.4$
        & $25.3 \pm 0.4$
        & $26.9 \pm 0.3$\\
    \cellcolor{Gray}{\model}
        & \cellcolor{Gray}
        & \cellcolor{Gray}$33.4 \pm 0.4$
        & \cellcolor{Gray}$44.1 \pm 0.2$
        & \cellcolor{Gray}$30.3 \pm 0.5$
        & \cellcolor{Gray}$30.6 \pm 0.5$
        & \cellcolor{Gray}$32.0 \pm 0.4$
        & \cellcolor{Gray}$7.9 \pm 1.3$
        & \cellcolor{Gray}$10.4 \pm 1.1$
        & \cellcolor{Gray}$12.9 \pm 1.0$
        & \cellcolor{Gray}$24.6 \pm 0.1$
        & \cellcolor{Gray}$25.6 \pm 0.3$
        & \cellcolor{Gray}$27.2 \pm 0.3$\\
    \bottomrule
    \end{tabular}
}
\label{tab::ledetection-benchmark}
\end{table*}
\begin{figure*}[t]
\begin{minipage}[b]{0.48\textwidth}
\captionof{table}{Ablation experiments quantifying the effectiveness of each component in our semi-supervised approach using 1\% of COCO labels. The first row corresponds to the Soft Teacher baseline and the last row is our \model configuration.}
\centering
\captionsetup{type=table}
\resizebox{\columnwidth}{!}{
    \begin{tabular}[t]{lcrr}
	\toprule
	Proposal Similarity Measure & 
	  Proposal IoU Regression &
	AP$_{50:95}$ &
	AR$_{50:95}$\\
        \midrule
        None &
        \xmark &
        22.4 &
        30.8\\
        KL-Divergence &
        \xmark &
        22.8 &
        31.5\\
        Cross-Entropy (\cref{eq::cross-entropy-similarity}) &
        \xmark &
        22.7 &
        31.6\\
        None &
        \cmark &
        22.3 &
        30.8\\
        KL-Divergence &
        \cmark &
        22.9 &
        31.8\\
        \cellcolor{Gray}{Cross-Entropy (\cref{eq::cross-entropy-similarity})} &
        \cellcolor{Gray}{\cmark} &
        \cellcolor{Gray}{\bfseries 23.0} &
        \cellcolor{Gray}{\bfseries 32.0}\\
    \bottomrule
    \end{tabular}
}
\label{tab::ablation-design}
\end{minipage} \hfill
\begin{minipage}[b]{0.48\textwidth}
    \centering
    \includegraphics[width=0.6\textwidth]{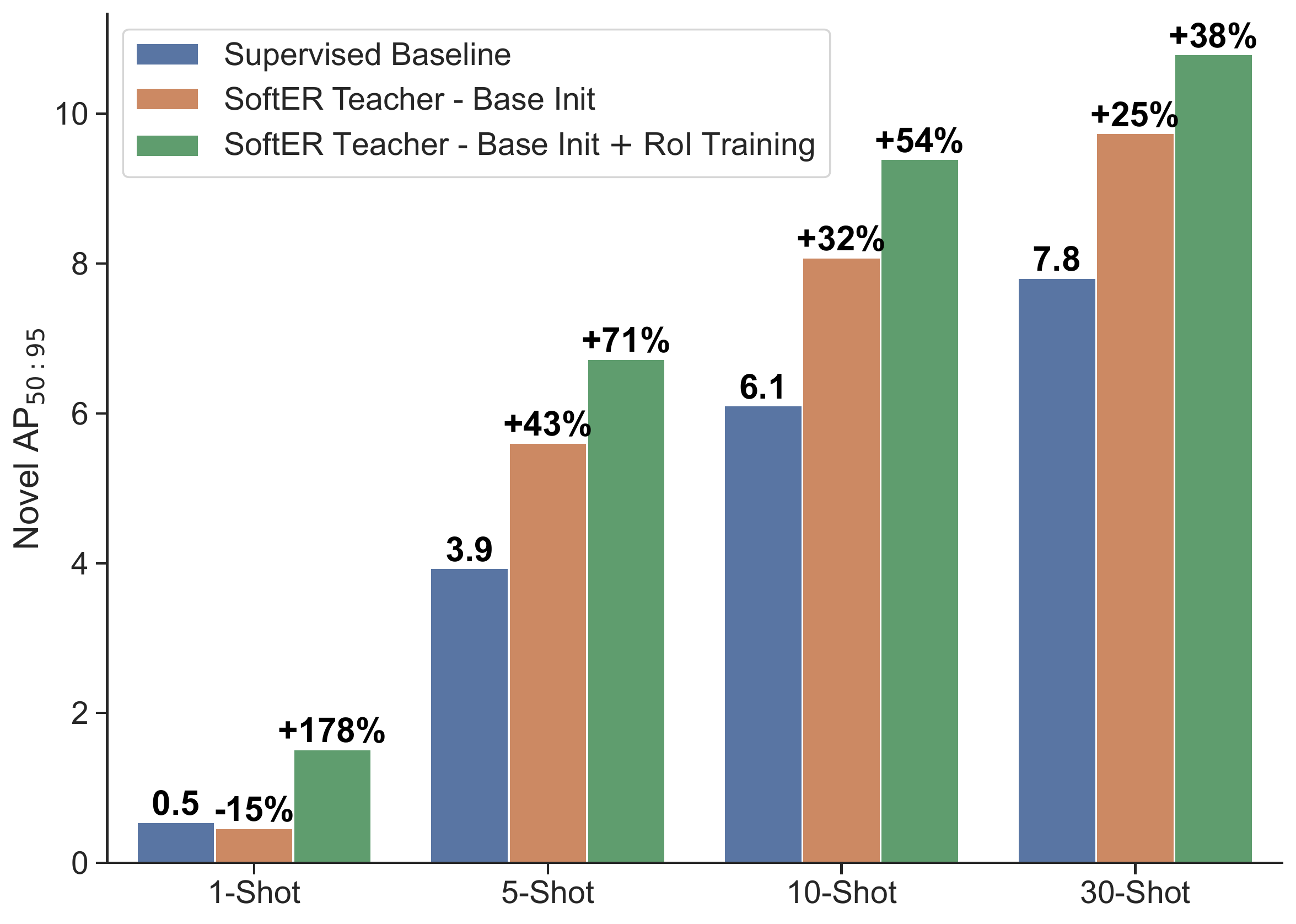}
    \captionof{figure}{Unlabeled data improve semi-supervised few-shot fine-tuning on COCO \texttt{val2017}.}
    \label{fig::ablation-novel-fine-tuning}
\end{minipage}
\end{figure*}

\paragraph{Datasets}
We evaluate our approach on the challenging PASCAL VOC~\citep{voc} and MS-COCO 2017~\citep{coco} detection benchmarks. For VOC, we use the combined VOC07$+$12 \verb|trainval| splits as the labeled training set and evaluate on the VOC2007 \verb|test| set. For COCO, we utilize the \verb|train2017| split as labeled data and test on \verb|val2017|.

\paragraph{Implementation Details}
We conduct our few-shot experiments following the TFA benchmark~\citep{tfa}. The VOC dataset is randomly partitioned into 15 base and 5 novel classes, in which there are $k \in \{1,5,10\}$ labeled boxes per category sampled from the joint VOC07$+$12 \verb|trainval| splits. And the COCO dataset is divided into 60 base and 20 novel classes having the same VOC category names with $k \in \{5,10,30\}$ shots. We leverage \verb|COCO-train2017| and \verb|COCO-unlabeled2017| as external unlabeled sources to augment base pre-training and novel fine-tuning on VOC and COCO, respectively. We adopt the ResNet-101 backbone pre-trained on ImageNet-1K~\citep{imagenet} for a direct comparison with existing work. We assess detection performance using average precision (AP) and recall (AR) metrics, following established protocols. Complete details are given in \Cref{sec::implementation-details}.

\subsection{\model is a Label-Efficient Few-Shot Detector}\label{sec::few-shot-results}
\Cref{tab::fsod-sota-comparison} compares the effectiveness of \model against other two-stage learning methods representing the state of the art on COCO, for the evaluation of both base and novel performances. We report the ideal supervised base AP of $39.3$~\citep{retentive-rcnn} along with our substantially improved semi-supervised base AP of $44.4$ to measure the extent of base forgetting. Recall that the more realistic evaluation metric for \emph{generalized} FSOD is not only novel AP, but the combination of base and novel AP. We summarize the following key takeaways: \textbf{(a)} \model with ResNet-101 trained with supplementary unlabeled data is the best model on the combined \texttt{Overall AP} metric for 80 classes, leading the next best Retentive R-CNN by up to $+2.0$ AP; \textbf{(b)} \model with ResNet-50 surpasses Retentive R-CNN on novel performance while being more parameter-efficient; and \textbf{(c) \model achieves the state of the art while being more efficient with respect to parameters and labels}.

\Cref{tab::fsod-voc} shows comparative VOC results. We report the supervised base AP of $80.8$~\citep{retentive-rcnn} along with our vastly expanded base AP of $85.0$ using unlabeled data. \model with ResNet-50 incurs negligible base forgetting of less than 2\% while exceeding the competition with ResNet-101 on most metrics. When equipped with ResNet-101, \model further improves on both base and novel classes by a notable margin. To our knowledge, MPSR, Retentive R-CNN, and LVC did not perform repeated runs over multiple random seeds per established protocol, the results of which are not comparable to ours. In general, the previous works of TFA, DeFRCN, DCFS, and ours all report marked reduction in novel performances with repeated sample runs when compared to a single trial.

\vspace{1.25pt}

\textbf{Discussion} ~ \Cref{tab::fsod-sota-comparison,tab::fsod-voc} corroborate our observation on the trade-off between novel performance and base forgetting, for which our approach aims to jointly optimize. We accomplish this goal with the help of our Entropy Regression module to learn with external unlabeled data ``in the wild'', without assuming the presence of abundant novel instances in such sources. Our VOC experiments yield state-of-the-art results by harnessing uncurated images from the \texttt{COCO-train2017} dataset, which exhibits strong domain mismatch and contains many classes outside of VOC. By stark contrast, LVC mines novel targets as auxiliary samples from the \emph{base training set}, thereby making an unrealistic assumption that abundant novel classes necessarily be present at training time. While such assumption enables impressive novel performance, LVC suffers significant base forgetting, making it sub-optimal in practical settings.

Similarly, DeFRCN and DCFS extend the base FRCN architecture with complex auxiliary modules specially designed to promote strong novel performance, but also come at the cost of considerable base forgetting. We view these methods as complementary to ours, which can be further extended to take advantage of unlabeled data. Future work would explore how our \model formulation could be integrated with Retentive R-CNN, DeFRCN, and DCFS to push the performance envelope of FSOD without forgetting.

\subsection{How Does Proposal Quality Impact Semi-Supervised Few-Shot Detection?}\label{sec::proposal-quality}
In this section, we continue our discussion from \Cref{sec::effective-fsod} by analyzing semi-supervised FSOD as a function of proposal quality in~\Cref{fig::ar-vs-ap}. We measure proposal quality using the metric AR@$p$, for $p \in \{100,300,1000\}$ proposals, averaged over thresholds between 0.5 and 0.95. We arrive at the following conclusions: \textbf{(a)} \model produces better proposals than the comparisons across varying fractions of base labels; and \textbf{(b--d)} proposal quality is strongly correlated with semi-supervised FSOD, an insightful empirical finding that extends existing results beyond supervised detection~\citep{effective-proposals}.

\smallskip

Although the strong Soft Teacher baseline is also effective at harnessing unlabeled data for semi-supervised FSOD, \emph{our approach demonstrates superior learning by addressing a key shortcoming of Soft Teacher}. \model boosts object recall via our proposed Entropy Regression module, improving on Soft Teacher by $+1.7$ base AR$_{50:95}$, which yields a gain of $+1.5$ novel AP for the \verb|30-Shot@10-Percent| setting. These results further bolster our empirical observation that a stronger semi-supervised detector leads to a more label-efficient few-shot detector. In principle, our versatile semi-supervised FSOD framework can generalize any semi-supervised detector to the few-shot setting. Future work would investigate if our findings can be extended to a more general case with other SSOD formulations, including one-stage detectors.

\subsection{A New Benchmark for Generalized Semi-Supervised Few-Shot Detection}
We present our Label-Efficient Detection benchmark on COCO \verb|val2017| in~\Cref{tab::ledetection-benchmark}. Our protocol for semi-supervised FSOD is as follows. In the first stage, we pre-train the base detector on the disjoint 60 base categories using $\{1,5,10\}$ percent of labels per \cref{eq::final_loss_softer}. In the second stage, we transfer the parameters of the base teacher detector to the few-shot detector and fine-tune its RoI box classifier, keeping other components frozen, on a balanced training set of $k \in \{5,10,30\}$ shots per class containing both base and novel examples. In both stages, we supplement base pre-training and novel fine-tuning with images from \verb|COCO-unlabeled2017|.

We report AP$_{50:95}$ performances on both base and novel classes along with the aggregated overall metric. We also report the ideal AP$_{50:95}$ and AR$_{50:95}$ metrics obtained from the first stage of base pre-training to measure the potential for base forgetting during the few-shot fine-tuning step. We encourage future work to follow suit as we emphasize the importance of optimizing for accuracy on both base and novel classes, a desideratum of generalized few-shot object detection.

\section{Ablation Studies}\label{sec::ablation}
\subsection{\model System Design}
\Cref{tab::ablation-design} shows an ablation study on 1\% of COCO labels to assess the key elements in our \model approach for SSOD. Compared to the Soft Teacher baseline (first row), the addition of the cross-entropy or KL-divergence measure to enforce proposal consistency leads to a boost in both AP and AR, although the performance difference between the two measures is immaterial. Interestingly, the addition of the IoU regression loss alone does not produce a performance improvement over the Soft Teacher baseline. However, when we couple IoU regression with the cross-entropy similarity measure, we obtain the best performing configuration (last row). \model improves on both precision and recall over the strong Soft Teacher baseline via our proposed Entropy Regression module for proposal learning with complex affine transformations.

\subsection{Semi-Supervised Few-Shot Fine-Tuning with Unlabeled Data}
As discussed in \Cref{sec::novel-finetune}, we explore two ways of leveraging unlabeled data to fine-tune the few-shot detector on novel classes: (1) we initialize the few-shot detector with parameters copied from the base \emph{teacher} detector pre-trained with unlabeled data per \cref{eq::final_loss_softer}; and (2) we further train the RoI box classifier and regressor on novel classes using the available few-shot and unlabeled examples while freezing the base backbone, FPN, and RPN components. \Cref{fig::ablation-novel-fine-tuning} illustrates semi-supervised base initialization boosts novel AP by as much as 43\% on COCO \texttt{val2017}, compared to the supervised baseline. In addition to semi-supervised base initialization, training the RoI head on few-shot novel classes with unlabeled images further amplifies the novel AP margin of \model.

\section{CONCLUSION}
This paper presented the Label-Efficient Detection framework to quantify the utility of unlabeled data for generalized semi-supervised FSOD. Central to the framework is our proposed \model model to boost semi-supervised FSOD, by way of consistency learning on diverse region proposals, without relying on an abundance of labels. Our simple and versatile LEDetection framework can inspire exciting future research directions to combine the latest advances in both SSOD and FSOD, thereby helping to unify the two disparate domains.


\acknowledgments{The author is grateful to Victor Palmer for insightful discussions, Brian Keller and Cole Winans for their continued support, and anonymous reviewers for their constructive feedback on this paper.}

\clearpage

\bibliographystyle{author-year}
\bibliography{refs}

\begin{thebibliography}{45}
\providecommand{\natexlab}[1]{#1}
\providecommand{\url}[1]{\texttt{#1}}
\expandafter\ifx\csname urlstyle\endcsname\relax
  \providecommand{\doi}[1]{doi: #1}\else
  \providecommand{\doi}{doi: \begingroup \urlstyle{rm}\Url}\fi

\bibitem[Bachman et~al.(2014)Bachman, Alsharif, and Precup]{pea}
Bachman, P., Alsharif, O., and Precup, D.
\newblock {Learning with Pseudo-Ensembles}.
\newblock In \emph{NeurIPS}, 2014.

\bibitem[Cao et~al.(2022)Cao, Wang, Lin, and Lin]{mini}
Cao, Y., Wang, J., Lin, Y., and Lin, D.
\newblock {MINI: Mining Implicit Novel Instances for Few-Shot Object Detection}.
\newblock https://arxiv.org/abs/2205.03381, 2022.

\bibitem[Chen et~al.(2019)Chen, Wang, Pang, Cao, Xiong, Li, et~al.]{mmdetection}
Chen, K., Wang, J., Pang, J., Cao, Y., Xiong, Y., Li, X., et~al.
\newblock {MMDetection: Open MMLab Detection Toolbox and Benchmark}.
\newblock https://arxiv.org/abs/1906.07155, 2019.

\bibitem[Chen \& He(2021)Chen and He]{simsiam}
Chen, X. and He, K.
\newblock {Exploring Simple Siamese Representation Learning}.
\newblock In \emph{CVPR}, 2021.

\bibitem[Cubuk et~al.(2020)Cubuk, Zoph, Shlens, and Le]{randaugment}
Cubuk, E.~D., Zoph, B., Shlens, J., and Le, Q.~V.
\newblock {RandAugment: Practical Automated Data Augmentation with a Reduced Search Space}.
\newblock In \emph{NeurIPS}, 2020.

\bibitem[Deng et~al.(2009)Deng, Dong, Socher, Li, Li, and Fei{-}Fei]{imagenet}
Deng, J., Dong, W., Socher, R., Li, L., Li, K., and Fei{-}Fei, L.
\newblock {ImageNet: A Large-Scale Hierarchical Image Database}.
\newblock In \emph{CVPR}, pp.\  248--255, 2009.
\newblock \doi{10.1109/CVPR.2009.5206848}.

\bibitem[DeVries \& Taylor(2017)DeVries and Taylor]{cutout}
DeVries, T. and Taylor, G.~W.
\newblock {Improved Regularization of Convolutional Neural Networks with Cutout}.
\newblock https://arxiv.org/abs/1708.04552, 2017.

\bibitem[Everingham et~al.(2010)Everingham, Gool, Williams, Winn, and Zisserman]{voc}
Everingham, M., Gool, L.~V., Williams, C.~K., Winn, J., and Zisserman, A.
\newblock {The PASCAL Visual Object Classes (VOC) Challenge}.
\newblock \emph{IJCV}, 88(2):\penalty0 303--338, 2010.

\bibitem[Fan et~al.(2021)Fan, Ma, Li, and Sun]{retentive-rcnn}
Fan, Z., Ma, Y., Li, Z., and Sun, J.
\newblock {Generalized Few-Shot Object Detection without Forgetting}.
\newblock In \emph{CVPR}, 2021.

\bibitem[Gao et~al.(2022)Gao, Chen, Huang, Nie, Liu, Lai, Jiang, Wang, and Wang]{dcfs}
Gao, B.-B., Chen, X., Huang, Z., Nie, C., Liu, J., Lai, J., Jiang, G., Wang, X., and Wang, C.
\newblock {Decoupling Classifier for Boosting Few-Shot Object Detection and Instance Segmentation}.
\newblock In \emph{NeurIPS}, 2022.

\bibitem[Grandvalet \& Bengio(2004)Grandvalet and Bengio]{entmin}
Grandvalet, Y. and Bengio, Y.
\newblock {Semi-Supervised Learning by Entropy Minimization}.
\newblock In \emph{NeurIPS}, 2004.

\bibitem[Grill et~al.(2020)Grill, Strub, Altch\'{e}, Tallec, Richemond, Buchatskaya, Doersch, Pires, et~al.]{byol}
Grill, J.-B., Strub, F., Altch\'{e}, F., Tallec, C., Richemond, P.~H., Buchatskaya, E., Doersch, C., Pires, B.~A., et~al.
\newblock {Bootstrap Your Own Latent: A New Approach to Self-Supervised Learning}.
\newblock In \emph{NeurIPS}, 2020.

\bibitem[He et~al.(2016)He, Zhang, Ren, and Sun]{resnet}
He, K., Zhang, X., Ren, S., and Sun, J.
\newblock {Deep Residual Learning for Image Recognition}.
\newblock In \emph{CVPR}, 2016.

\bibitem[He et~al.(2017)He, Gkioxari, Doll\'{a}r, and Girshick]{mask-rcnn}
He, K., Gkioxari, G., Doll\'{a}r, P., and Girshick, R.
\newblock {Mask R-CNN}.
\newblock In \emph{ICCV}, 2017.

\bibitem[Hosang et~al.(2016)Hosang, Benenson, Doll\'{a}r, and Schiele]{effective-proposals}
Hosang, J., Benenson, R., Doll\'{a}r, P., and Schiele, B.
\newblock {What Makes for Effective Detection Proposals?}
\newblock \emph{IEEE TPAMI}, 38(4):\penalty0 814--830, 2016.

\bibitem[Karlinsky et~al.(2019)Karlinsky, Shtok, Harary, Schwartz, Aides, Feris, Giryes, and Bronstein]{repmet}
Karlinsky, L., Shtok, J., Harary, S., Schwartz, E., Aides, A., Feris, R., Giryes, R., and Bronstein, A.~M.
\newblock {RepMet: Representative-Based Metric Learning for Classification and Few-Shot Object Detection}.
\newblock In \emph{CVPR}, 2019.

\bibitem[Kaul et~al.(2022)Kaul, Xie, and Zisserman]{lvc}
Kaul, P., Xie, W., and Zisserman, A.
\newblock {Label, Verify, Correct: A Simple Few Shot Object Detection Method}.
\newblock In \emph{CVPR}, 2022.

\bibitem[Khandelwal et~al.(2021)Khandelwal, Goyal, and Sigal]{unit}
Khandelwal, S., Goyal, R., and Sigal, L.
\newblock {UniT: Unified Knowledge Transfer for Any-shot Object Detection and Segmentation}.
\newblock In \emph{CVPR}, 2021.

\bibitem[Li et~al.(2018)Li, Peng, Yu, Zhang, Deng, and Sun]{detnet}
Li, Z., Peng, C., Yu, G., Zhang, X., Deng, Y., and Sun, J.
\newblock {DetNet: A Backbone Network for Object Detection}.
\newblock In \emph{ECCV}, 2018.

\bibitem[Lin et~al.(2014)Lin, Maire, Belongie, Hays, Perona, Ramanan, Doll\'{a}r, and Zitnick]{coco}
Lin, T.-Y., Maire, M., Belongie, S., Hays, J., Perona, P., Ramanan, D., Doll\'{a}r, P., and Zitnick, C.~L.
\newblock {Microsoft COCO: Common Objects in Context}.
\newblock In \emph{ECCV}, 2014.

\bibitem[Lin et~al.(2017)Lin, Doll\'{a}r, Girshick, He, Hariharan, and Belongie]{fpn}
Lin, T.-Y., Doll\'{a}r, P., Girshick, R., He, K., Hariharan, B., and Belongie, S.
\newblock {Feature Pyramid Networks for Object Detection}.
\newblock In \emph{CVPR}, 2017.

\bibitem[Liu et~al.(2021)Liu, Ma, He, Kuo, Chen, Zhang, Wu, Kira, and Vajda]{ubteacher}
Liu, Y.-C., Ma, C.-Y., He, Z., Kuo, C.-W., Chen, K., Zhang, P., Wu, B., Kira, Z., and Vajda, P.
\newblock {Unbiased Teacher for Semi-Supervised Object Detection}.
\newblock In \emph{ICLR}, 2021.

\bibitem[Liu et~al.(2022)Liu, Ma, and Kira]{ubteacherv2}
Liu, Y.-C., Ma, C.-Y., and Kira, Z.
\newblock {Unbiased Teacher v2: Semi-Supervised Object Detection for Anchor-Free and Anchor-Based Detectors}.
\newblock In \emph{CVPR}, 2022.

\bibitem[Lopez-Paz \& Ranzato(2017)Lopez-Paz and Ranzato]{gem}
Lopez-Paz, D. and Ranzato, M.
\newblock {Gradient Episodic Memory for Continual Learning}.
\newblock In \emph{NeurIPS}, 2017.

\bibitem[Ma et~al.(2023)Ma, Niu, Xu, Huang, Han, and Chang]{digeo}
Ma, J., Niu, Y., Xu, J., Huang, S., Han, G., and Chang, S.-F.
\newblock {DiGeo: Discriminative Geometry-Aware Learning for Generalized Few-Shot Object Detection}.
\newblock In \emph{CVPR}, 2023.

\bibitem[Miyato et~al.(2017)Miyato, Maeda, Koyama, and Ishii]{vat2}
Miyato, T., Maeda, S., Koyama, M., and Ishii, S.
\newblock {Virtual Adversarial Training: A Regularization Method for Supervised and Semi-Supervised Learning}.
\newblock \emph{IEEE TPAMI}, 41:\penalty0 1979--1993, 2017.

\bibitem[Oliver et~al.(2018)Oliver, Odena, Raffel, Cubuk, and Goodfellow]{ssleval}
Oliver, A., Odena, A., Raffel, C., Cubuk, E.~D., and Goodfellow, I.~J.
\newblock {Realistic Evaluation of Deep Semi-Supervised Learning Algorithms}.
\newblock In \emph{NeurIPS}, 2018.

\bibitem[Paszke et~al.(2019)Paszke, Gross, Massa, Lerer, Bradbury, et~al.]{pytorch}
Paszke, A., Gross, S., Massa, F., Lerer, A., Bradbury, J., et~al.
\newblock {PyTorch: An Imperative Style, High-Performance Deep Learning Library}.
\newblock In \emph{NeurIPS}, pp.\  8024--8035. Curran Associates, Inc., 2019.

\bibitem[Qiao et~al.(2021)Qiao, Zhao, Li, Qiu, Wu, and Zhang]{defrcn}
Qiao, L., Zhao, Y., Li, Z., Qiu, X., Wu, J., and Zhang, C.
\newblock {DeFRCN: Decoupled Faster R-CNN for Few-Shot Object Detection}.
\newblock In \emph{ICCV}, 2021.

\bibitem[Ren et~al.(2015)Ren, He, Girshick, and Sun]{faster-rcnn}
Ren, S., He, K., Girshick, R., and Sun, J.
\newblock {Faster R-CNN: Towards Real-Time Object Detection with Region Proposal Networks}.
\newblock In \emph{NeurIPS}, 2015.

\bibitem[Rezatofighi et~al.(2019)Rezatofighi, Tsoi, Gwak, Sadeghian, Reid, and Savarese]{giou}
Rezatofighi, H., Tsoi, N., Gwak, J., Sadeghian, A., Reid, I., and Savarese, S.
\newblock {Generalized Intersection over Union: A Metric and A Loss for Bounding Box Regression}.
\newblock In \emph{CVPR}, 2019.

\bibitem[Sajjadi et~al.(2016)Sajjadi, Javanmardi, and Tasdizen]{consistency}
Sajjadi, M., Javanmardi, M., and Tasdizen, T.
\newblock {Regularization with Stochastic Perturbations for Deep Semi-Supervised Learning}.
\newblock In \emph{NeurIPS}, 2016.

\bibitem[Sohn et~al.(2020{\natexlab{a}})Sohn, Berthelot, Li, Zhang, Carlini, Cubuk, Kurakin, Zhang, and Raffel]{fixmatch}
Sohn, K., Berthelot, D., Li, C.-L., Zhang, Z., Carlini, N., Cubuk, E.~D., Kurakin, A., Zhang, H., and Raffel, C.
\newblock {Fixmatch: Simplifying Semi-Supervised Learning with Consistency and Confidence}.
\newblock In \emph{NeurIPS}, 2020{\natexlab{a}}.

\bibitem[Sohn et~al.(2020{\natexlab{b}})Sohn, Zhang, Li, Zhang, Lee, and Pfister]{stac}
Sohn, K., Zhang, Z., Li, C.-L., Zhang, H., Lee, C.-Y., and Pfister, T.
\newblock {A Simple Semi-Supervised Learning Framework for Object Detection}.
\newblock https://arxiv.org/abs/2005.04757, 2020{\natexlab{b}}.

\bibitem[Sun et~al.(2021)Sun, Li, Cai, Yuan, and Zhang]{fsce}
Sun, B., Li, B., Cai, S., Yuan, Y., and Zhang, C.
\newblock {FSCE: Few-Shot Object Detection via Contrastive Proposal Encoding}.
\newblock In \emph{CVPR}, 2021.

\bibitem[Tang et~al.(2021)Tang, Chen, Luo, and Zhang]{humble-teacher}
Tang, Y., Chen, W., Luo, Y., and Zhang, Y.
\newblock {Humble Teachers Teach Better Students for Semi-Supervised Object Detection}.
\newblock In \emph{CVPR}, 2021.

\bibitem[Tarvainen \& Valpola(2017)Tarvainen and Valpola]{mean-teacher}
Tarvainen, A. and Valpola, H.
\newblock {Mean Teachers are Better Role Models: Weight-Averaged Consistency Targets Improve Semi-Supervised Deep Learning Results}.
\newblock In \emph{NeurIPS}, 2017.

\bibitem[Vu et~al.(2019)Vu, Jang, Pham, and Yoo]{crpn}
Vu, T., Jang, H., Pham, T.~X., and Yoo, C.~D.
\newblock {Cascade RPN: Delving into High-Quality Region Proposal Network with Adaptive Convolution}.
\newblock In \emph{NeurIPS}, 2019.

\bibitem[Wang et~al.(2020)Wang, Huang, Darrell, Gonzalez, and Yu]{tfa}
Wang, X., Huang, T.~E., Darrell, T., Gonzalez, J.~E., and Yu, F.
\newblock {Frustratingly Simple Few-Shot Object Detection}.
\newblock In \emph{ICML}, 2020.

\bibitem[Wang et~al.(2023)Wang, Yang, Zhang, Li, Feng, Fang, Lyu, Chen, and Zhang]{consistent-teacher}
Wang, X., Yang, X., Zhang, S., Li, Y., Feng, L., Fang, S., Lyu, C., Chen, K., and Zhang, W.
\newblock {Consistent-Teacher: Towards Reducing Inconsistent Pseudo-Targets in Semi-Supervised Object Detection}.
\newblock In \emph{CVPR}, 2023.

\bibitem[Wu et~al.(2020)Wu, Liu, Huang, and Wang]{mpsr}
Wu, J., Liu, S., Huang, D., and Wang, Y.
\newblock {Multi-Scale Positive Sample Refinement for Few-Shot Object Detection}.
\newblock In \emph{ECCV}, 2020.

\bibitem[Xiong et~al.(2021)Xiong, Cui, and Liu]{ssfod}
Xiong, W., Cui, Y., and Liu, L.
\newblock {Semi-Supervised Few-Shot Object Detection with a Teacher-Student Network}.
\newblock In \emph{BMVC}, 2021.

\bibitem[Xu et~al.(2021)Xu, Zhang, Hu, Wang, Wang, Wei, Bai, and Liu]{soft-teacher}
Xu, M., Zhang, Z., Hu, H., Wang, J., Wang, L., Wei, F., Bai, X., and Liu, Z.
\newblock {End-to-End Semi-Supervised Object Detection with Soft Teacher}.
\newblock In \emph{ICCV}, 2021.

\bibitem[Yang et~al.(2023)Yang, Dong, Ward, Dhillon, Sanghavi, and Lei]{dac}
Yang, S., Dong, Y., Ward, R., Dhillon, I.~S., Sanghavi, S., and Lei, Q.
\newblock {Sample Efficiency of Data Augmentation Consistency Regularization}.
\newblock In \emph{{Proceedings of The 26th International Conference on Artificial Intelligence and Statistics (AISTATS)}}, volume 206, pp.\  3825--3853. PMLR, 2023.

\bibitem[Zhong et~al.(2020)Zhong, Zheng, Kang, Li, and Yang]{random-erase}
Zhong, Z., Zheng, L., Kang, G., Li, S., and Yang, Y.
\newblock {Random Erasing Data Augmentation}.
\newblock In \emph{AAAI}, 2020.

\end{thebibliography}


\clearpage
\appendix
\onecolumn


\begin{table*}[t]
\caption{Generalized FSOD results evaluated on VOC07 \texttt{test} over three random partitions. We compare our \model against its Soft Teacher counterpart and strong supervised baselines. We report the mean and 95\% confidence interval over 10 random samples for our models. \model with ResNet-50 exceeds the supervised models with ResNet-101 by a large margin across most metrics under consideration.}
\begin{minipage}[t]{\textwidth}
\centering
\resizebox{\columnwidth}{!}{
    \begin{tabular}{lccc|ccc|ccc|ccc|}
    \toprule
    \textbf{VOC07 \texttt{test} -- Split 1} &
    \multirow{2}{*}{Backbone} &
    \multirow{1}{*}{\parbox{0.05\columnwidth}{\vspace{1.5mm}Base}} &
    \multirow{1}{*}{\parbox{0.05\columnwidth}{\vspace{1.5mm}Base}} &
    \multicolumn{3}{c|}{Base AP$_{50}$ (15 Classes)} &
    \multicolumn{3}{c|}{Novel AP$_{50}$ (5 Classes)} &
    \multicolumn{3}{c|}{Overall AP$_{50}$ (20 Classes)}\\
    \cmidrule{5-13}
    \multirow{1}{*}{Method} &
    &
    \multirow{1}{*}{\parbox{0.05\columnwidth}{\vspace{-1.5mm}AP$_{50}$}} &
    \multirow{1}{*}{\parbox{0.05\columnwidth}{\vspace{-1.5mm}AR$_{50}$}} &
    \multirow{1}{*}{1-Shot} &
    \multirow{1}{*}{5-Shot} &
    \multirow{1}{*}{10-Shot} &
    \multirow{1}{*}{1-Shot} &
    \multirow{1}{*}{5-Shot} &
    \multirow{1}{*}{10-Shot} &
    \multirow{1}{*}{1-Shot} &
    \multirow{1}{*}{5-Shot} &
    \multirow{1}{*}{10-Shot} \\
    \midrule
        MPSR* [ECCV'20]
        & R-101
        & $80.8$
        & --
        & $61.5$
        & $69.7$
        & $71.6$
        & \bfseries 42.8
        & $55.3$
        & $61.2$
        & $56.8$
        & $66.1$
        & $69.0$\\
        Retentive R-CNN* [CVPR'21]
        & R-101
        & $80.8$
        & --
        & $80.9$
        & $80.8$
        & $80.8$
        & $42.4$
        & $53.7$
        & $56.1$
        & $71.3$
        & $74.0$
        & $74.6$\\
        TFA [ICML'20]
        & R-101
        & $80.8$
        & --
        & $77.6 \pm 0.2$ 
        & $77.4 \pm 0.3$
        & $77.5 \pm 0.2$
        & $25.3 \pm 2.2$
        & $47.9 \pm 1.2$
        & $52.8 \pm 1.0$
        & $64.5 \pm 0.6$
        & $70.1 \pm 0.4$
        & $71.3 \pm 0.3$\\
        \cellcolor{Gray}{\model (Ours)}
        & \cellcolor{Gray}{R-101}
        & \cellcolor{Gray}{\bfseries 85.0}
        & \cellcolor{Gray}{\bfseries 93.1}
        & \cellcolor{Gray}{\bfseries 83.3 $\pm$ 0.4}
        & \cellcolor{Gray}{\bfseries 84.6 $\pm$ 0.2}
        & \cellcolor{Gray}{\bfseries 84.9 $\pm$ 0.2}
        & \cellcolor{Gray}{$35.5 \pm 3.7$}
        & \cellcolor{Gray}{\bfseries 60.3 $\pm$ 2.8}
        & \cellcolor{Gray}{\bfseries 65.9 $\pm$ 1.6}
        & \cellcolor{Gray}{\bfseries 71.3 $\pm$ 1.2}
        & \cellcolor{Gray}{\bfseries 78.5 $\pm$ 0.8}
        & \cellcolor{Gray}{\bfseries 80.2 $\pm$ 0.4}\\
    \midrule
        Faster R-CNN (Our Impl.)
        & R-50
        & $81.7$
        & $88.0$
        & $82.0 \pm 0.2$
        & $82.4 \pm 0.1$
        & $82.3 \pm 0.1$
        & $27.9 \pm 3.2$
        & $52.1 \pm 2.1$
        & $58.2 \pm 1.6$
        & $68.5 \pm	0.8$
        & $74.9	\pm 0.5$
        & $76.2	\pm 0.4$\\
        Soft Teacher (Our Impl.)
        & R-50
        & \bfseries 85.0
        & $90.9$
        & \bfseries 84.6 $\pm$ 0.4
        & \bfseries 85.2 $\pm$ 0.1
        & \bfseries 85.0 $\pm$ 0.1
        & $29.5 \pm 4.4$
        & $53.8 \pm 2.6$
        & $60.8 \pm 1.5$
        & \bfseries 70.8 $\pm$ 1.2
        & $77.4	\pm 0.7$
        & $79.0	\pm 0.3$\\
        \cellcolor{Gray}{\model (Ours)}
        & \cellcolor{Gray}{R-50}
        & \cellcolor{Gray}{$84.7$}
        & \cellcolor{Gray}{\bfseries 92.3}
        & \cellcolor{Gray}{$83.1 \pm 0.2$}
        & \cellcolor{Gray}{$84.4 \pm 0.2$}
        & \cellcolor{Gray}{$84.6 \pm 0.2$}
        & \cellcolor{Gray}{\bfseries 31.3 $\pm$ 4.3} 
        & \cellcolor{Gray}{\bfseries 57.8 $\pm$ 2.6}
        & \cellcolor{Gray}{\bfseries 62.9 $\pm$ 1.7}
        & \cellcolor{Gray}{$70.2 \pm 1.2$}
        & \cellcolor{Gray}{\bfseries 77.8 $\pm$ 0.7}
        & \cellcolor{Gray}{\bfseries 79.2 $\pm$ 0.4}\\
    \end{tabular}
}
\end{minipage}
\begin{minipage}[t]{\textwidth}
\centering
\resizebox{\columnwidth}{!}{
    \begin{tabular}{lccc|ccc|ccc|ccc|}
    \toprule
    \textbf{VOC07 \texttt{test} -- Split 2} &
    \multirow{2}{*}{Backbone} &
    \multirow{1}{*}{\parbox{0.05\columnwidth}{\vspace{1.5mm}Base}} &
    \multirow{1}{*}{\parbox{0.05\columnwidth}{\vspace{1.5mm}Base}} &
    \multicolumn{3}{c|}{Base AP$_{50}$ (15 Classes)} &
    \multicolumn{3}{c|}{Novel AP$_{50}$ (5 Classes)} &
    \multicolumn{3}{c|}{Overall AP$_{50}$ (20 Classes)}\\
    \cmidrule{5-13}
    \multirow{1}{*}{Method} &
    &
    \multirow{1}{*}{\parbox{0.05\columnwidth}{\vspace{-1.5mm}AP$_{50}$}} &
    \multirow{1}{*}{\parbox{0.05\columnwidth}{\vspace{-1.5mm}AR$_{50}$}} &
    \multirow{1}{*}{1-Shot} &
    \multirow{1}{*}{5-Shot} &
    \multirow{1}{*}{10-Shot} &
    \multirow{1}{*}{1-Shot} &
    \multirow{1}{*}{5-Shot} &
    \multirow{1}{*}{10-Shot} &
    \multirow{1}{*}{1-Shot} &
    \multirow{1}{*}{5-Shot} &
    \multirow{1}{*}{10-Shot} \\
    \midrule
        MPSR* [ECCV'20]
        & R-101
        & $81.9$
        & --
        & $60.8$
        & $71.2$
        & $72.7$
        & \bfseries 29.8
        & \bfseries 43.2
        & $47.0$
        & $53.1$
        & $64.2$
        & $66.3$\\
        Retentive R-CNN* [CVPR'21]
        & R-101
        & $81.9$
        & --
        & $81.8$
        & $81.9$
        & $81.9$
        & $21.7$
        & $37.0$
        & $40.3$
        & $66.8$
        & $70.7$
        & $71.5$\\
        TFA [ICML'20]
        & R-101
        & $81.9$
        & --
        & $73.8 \pm 0.8$
        & $76.2 \pm 0.4$
        & $76.9 \pm 0.3$
        & $18.3 \pm 2.4$
        & $34.1 \pm 1.4$
        & $39.5 \pm 1.1$
        & $59.9 \pm 0.8$
        & $65.7 \pm 0.5$
        & $67.6 \pm 0.4$\\
        \cellcolor{Gray}{\model (Ours)}
        & \cellcolor{Gray}{R-101}
        & \cellcolor{Gray}{\bfseries 85.9}
        & \cellcolor{Gray}{\bfseries 93.1}
        & \cellcolor{Gray}{\bfseries 84.9 $\pm$ 0.2}
        & \cellcolor{Gray}{\bfseries 85.6 $\pm$ 0.2}
        & \cellcolor{Gray}{\bfseries 85.9 $\pm$ 0.2}
        & \cellcolor{Gray}{$23.8 \pm 4.7$}
        & \cellcolor{Gray}{$40.9 \pm 2.4$}
        & \cellcolor{Gray}{\bfseries 48.4 $\pm$ 2.1}
        & \cellcolor{Gray}{\bfseries 69.6 $\pm$ 1.1}
        & \cellcolor{Gray}{\bfseries 74.5 $\pm$ 0.7}
        & \cellcolor{Gray}{\bfseries 76.5 $\pm$ 0.5}\\
    \midrule
        Faster R-CNN (Our Impl.)
        & R-50
        & $82.9$
        & $88.7$
        & $83.1 \pm 0.1$
        & $83.5 \pm 0.1$
        & $83.3 \pm 0.1$
        & $18.3 \pm 4.3$
        & $34.9 \pm 1.5$
        & $40.6 \pm 1.7$
        & $66.9	\pm 1.1$
        & $71.4 \pm 0.4$
        & $72.6 \pm 0.4$\\
        Soft Teacher (Our Impl.)
        & R-50
        & $85.3$
        & $91.5$
        & \bfseries 85.1 $\pm$ 0.2
        & \bfseries 85.3 $\pm$ 0.1
        & \bfseries 85.3 $\pm$ 0.1
        & $20.3 \pm 4.7$
        & $37.9 \pm 2.1$
        & $44.0 \pm 1.8$
        & \bfseries 68.9 $\pm$ 1.2
        & $73.5 \pm 0.6$
        & $75.0 \pm 0.5$\\
        \cellcolor{Gray}{\model (Ours)}
        & \cellcolor{Gray}{R-50}
        & \cellcolor{Gray}{\bfseries 85.3}
        & \cellcolor{Gray}{\bfseries 93.1}
        & \cellcolor{Gray}{$84.2 \pm 0.2$}
        & \cellcolor{Gray}{$85.1 \pm 0.2$}
        & \cellcolor{Gray}{$85.2 \pm 0.2$}
        & \cellcolor{Gray}{\bfseries 22.4 $\pm$ 4.3}
        & \cellcolor{Gray}{\bfseries 41.1 $\pm$ 2.3}
        & \cellcolor{Gray}{\bfseries 47.3 $\pm$ 2.1}
        & \cellcolor{Gray}{$68.7 \pm 1.1$}
        & \cellcolor{Gray}{\bfseries 74.1 $\pm$ 0.7}
        & \cellcolor{Gray}{\bfseries 75.8 $\pm$ 0.6}\\
    \end{tabular}
}
\end{minipage}
\begin{minipage}[t]{\textwidth}
\centering
\resizebox{\columnwidth}{!}{
    \begin{tabular}{lccc|ccc|ccc|ccc|}
    \toprule
    \textbf{VOC07 \texttt{test} -- Split 3} &
    \multirow{2}{*}{Backbone} &
    \multirow{1}{*}{\parbox{0.05\columnwidth}{\vspace{1.5mm}Base}} &
    \multirow{1}{*}{\parbox{0.05\columnwidth}{\vspace{1.5mm}Base}} &
    \multicolumn{3}{c|}{Base AP$_{50}$ (15 Classes)} &
    \multicolumn{3}{c|}{Novel AP$_{50}$ (5 Classes)} &
    \multicolumn{3}{c|}{Overall AP$_{50}$ (20 Classes)}\\
    \cmidrule{5-13}
    \multirow{1}{*}{Method} &
    &
    \multirow{1}{*}{\parbox{0.05\columnwidth}{\vspace{-1.5mm}AP$_{50}$}} &
    \multirow{1}{*}{\parbox{0.05\columnwidth}{\vspace{-1.5mm}AR$_{50}$}} &
    \multirow{1}{*}{1-Shot} &
    \multirow{1}{*}{5-Shot} &
    \multirow{1}{*}{10-Shot} &
    \multirow{1}{*}{1-Shot} &
    \multirow{1}{*}{5-Shot} &
    \multirow{1}{*}{10-Shot} &
    \multirow{1}{*}{1-Shot} &
    \multirow{1}{*}{5-Shot} &
    \multirow{1}{*}{10-Shot} \\
    \midrule
        MPSR* [ECCV'20]
        & R-101
        & $82.0$
        & --
        & $61.6$
        & $72.9$
        & $73.2$
        & \bfseries 35.9
        & $48.9$
        & $51.3$
        & $55.2$
        & $66.9$
        & $67.7$\\
        Retentive R-CNN* [CVPR'21]
        & R-101
        & $82.0$
        & --
        & $81.9$
        & $82.0$
        & $82.1$
        & $30.2$
        & \bfseries 49.7
        & $50.1$
        & $69.0$
        & $73.9$
        & $74.1$\\
        TFA [ICML'20]
        & R-101
        & $82.0$
        & --
        & $78.7 \pm 0.2$
        & $78.5 \pm 0.3$
        & $78.6 \pm 0.2$
        & $17.9 \pm 2.0$
        & $40.8 \pm 1.4$
        & $45.6 \pm 1.1$
        & $63.5 \pm 0.6$
        & $69.1 \pm 0.4$
        & $70.3 \pm 0.4$\\
        \cellcolor{Gray}{\model (Ours)}
        & \cellcolor{Gray}{R-101}
        & \cellcolor{Gray}{\bfseries 86.7}
        & \cellcolor{Gray}{\bfseries 93.4}
        & \cellcolor{Gray}{\bfseries 85.4 $\pm$ 0.3}
        & \cellcolor{Gray}{\bfseries 86.4 $\pm$ 0.2}
        & \cellcolor{Gray}{\bfseries 86.5 $\pm$ 0.1}
        & \cellcolor{Gray}{$23.5 \pm 2.8$}
        & \cellcolor{Gray}{$46.6 \pm 2.7$}
        & \cellcolor{Gray}{\bfseries 53.8 $\pm$ 1.6}
        & \cellcolor{Gray}{\bfseries 70.0 $\pm$ 0.8}
        & \cellcolor{Gray}{\bfseries 76.5 $\pm$ 0.7}
        & \cellcolor{Gray}{\bfseries 78.3 $\pm$ 0.4}\\
    \midrule
        Faster R-CNN (Our Impl.)
        & R-50
        & $82.6$
        & $88.0$
        & $83.1 \pm 0.2$
        & $83.6 \pm 0.1$
        & $83.3 \pm 0.1$
        & $19.6 \pm 1.9$
        & $44.1 \pm 1.8$
        & $51.2 \pm 1.3$
        & $67.3 \pm 0.5$
        & $73.7 \pm 0.4$
        & $75.3 \pm 0.3$\\
        Soft Teacher (Our Impl.)
        & R-50
        & $85.1$
        & $91.0$
        & $84.7 \pm 0.2$
        & $85.2 \pm 0.1$
        & $85.0 \pm 0.1$
        & $22.2 \pm 3.0$
        & $49.2 \pm 2.2$
        & $55.6 \pm 1.4$
        & $69.0 \pm 0.9$
        & $76.2 \pm 0.5$
        & $77.7 \pm 0.3$\\
        \cellcolor{Gray}{\model (Ours)}
        & \cellcolor{Gray}{R-50}
        & \cellcolor{Gray}{\bfseries 85.9}
        & \cellcolor{Gray}{\bfseries 92.8}
        & \cellcolor{Gray}{\bfseries 84.7 $\pm$ 0.2}
        & \cellcolor{Gray}{\bfseries 85.4 $\pm$ 0.2}
        & \cellcolor{Gray}{\bfseries 85.4 $\pm$ 0.2}
        & \cellcolor{Gray}{\bfseries 24.4 $\pm$ 2.2}
        & \cellcolor{Gray}{\bfseries 49.5 $\pm$ 1.8}
        & \cellcolor{Gray}{\bfseries 55.9 $\pm$ 1.6}
        & \cellcolor{Gray}{\bfseries 69.6 $\pm$ 0.7}
        & \cellcolor{Gray}{\bfseries 76.4 $\pm$ 0.5}
        & \cellcolor{Gray}{\bfseries 78.0 $\pm$ 0.4}\\
    \bottomrule
    \multicolumn{13}{l}{* indicates results were reported on a single sample run (hence the lack of error bars), which may have been over-estimated due to the high variance of few-shot training samples.}
    \end{tabular}
}
\end{minipage}
\label{tab::fsod-voc-supp}
\end{table*}

\section{Additional Quantitative Results}
\subsection{Generalized Few-Shot Detection on PASCAL VOC}\label{sec::generalized-voc}
We present the generalized FSOD results on VOC in \Cref{tab::fsod-voc-supp}, which comprises three random partition splits. We leverage \texttt{COCO-train2017} as \emph{uncurated} unlabeled data to augment our experiments, which exhibits strong domain mismatch and contains many object classes outside of VOC. We report the ideal supervised base AP from previous work~\citep{tfa,retentive-rcnn} along with our substantially improved semi-supervised base AP to measure the extent of base forgetting. These results further support our observation on the trade-off between novel performance and base forgetting, for which our approach aims to simultaneously optimize. We summarize the following key takeaways.

\paragraph{Base Performance}
Our re-implementation of the supervised Faster R-CNN baseline \emph{does not} degrade base performance compared to the TFA benchmark across all three partitions. Base degradation is negligible with \model at less than $2\%$. We attribute this apparent improvement in base performance to our modified procedure of fine-tuning only the RoI box classifier and to our proposed Entropy Regression module enabling \model to achieve superior learning with unlabeled data.

\paragraph{\model \vs~Supervised Baselines}
\model with ResNet-50 surpasses the supervised MPSR, TFA, and Retentive R-CNN models with ResNet-101 by a large margin on the combined overall base $+$ novel AP metric across most experiments under consideration, while being more parameter-efficient with a smaller backbone. Although MPSR achieves impressive few-shot performance on novel categories, it suffers catastrophic base forgetting by as much as 26\%. Retentive R-CNN does not exhibit base class degradation, but generally falls behind on novel class performance. To our knowledge, MPSR and Retentive R-CNN did not perform repeated experiments over multiple random seeds per established protocol, the results of which may have been over-estimated due to the high variance of few-shot training samples, and thus are not directly comparable to ours. In general, the previous work of TFA and ours observe marked reduction in novel performances with repeated sample runs (\eg, 10 or 30) when compared to a single trial.

\paragraph{\model \vs~Soft Teacher}
Although the strong Soft Teacher baseline is effective at harnessing unlabeled data for semi-supervised FSOD, our \model approach demonstrates superior learning by addressing a key shortcoming of Soft Teacher. \model consistently boosts object recall via our proposed Entropy
Regression module, improving on Soft Teacher by as much as $+1.8$ base AR$_{50}$, which yields a gain of up to $+4.0$ novel AP$_{50}$. These results further bolster our empirical observation that a stronger semi-supervised detector leads to a more label-efficient few-shot detector.

\subsection{The Impact of Proposal Quality on Semi-Supervised Few-Shot Detection}\label{sec::generalized-proposal-quality}
We present expansive results on proposal quality and its relationship with semi-supervised few-shot detection in \Cref{tab::proposal-quality-supp}. Following existing literature~\citep{effective-proposals,crpn}, we measure proposal quality using the metric AR@$p$, for $p \in \{100,300,1000\}$ proposals, averaged over 10 overlap thresholds between 0.5 and 0.95. Proposal quality AR@$p$ is not to be confused with the detection metric AR$_{50:95}$, which is used to evaluate object coverage computed on a per-category basis and averaged over categories.

\begin{table*}[t]
\caption{Proposal quality is highly correlated with semi-supervised few-shot detection. \model produces the best proposal quality AR@$p$, for $p \in \{100,300,1000\}$, among the comparisons, which in turn yields the strongest novel $k$-shot performances given varying fractions of base labels. All models are equipped with the ResNet-101 backbone. We report the mean and standard deviation over 5 random samples.}
\centering
\resizebox{\textwidth}{!}{
    \begin{tabular}{lrccc|rrr|rrr|rrr|}
    \toprule
    \multirow{2}{*}{Method} &
    \multirow{2}{*}{\% Labeled} &
    \multirow{2}{*}{AR@100} &
    \multirow{2}{*}{AR@300} &
    \multirow{2}{*}{AR@1000} &
    \multicolumn{3}{c|}{Base AP$_{50:95}$ (60 Classes)} &
    \multicolumn{3}{c|}{Novel AP$_{50:95}$ (20 Classes)} &
    \multicolumn{3}{c|}{Overall AP$_{50:95}$ (80 Classes)} \\
    \cmidrule{6-14}
    & & & & &
    \multicolumn{1}{c}{\multirow{1}{*}{5-Shot}} &
    \multicolumn{1}{c}{\multirow{1}{*}{10-Shot}} &
    \multicolumn{1}{c|}{\multirow{1}{*}{30-Shot}} &
    \multicolumn{1}{c}{\multirow{1}{*}{5-Shot}} &
    \multicolumn{1}{c}{\multirow{1}{*}{10-Shot}} &
    \multicolumn{1}{c|}{\multirow{1}{*}{30-Shot}} &
    \multicolumn{1}{c}{\multirow{1}{*}{5-Shot}} &
    \multicolumn{1}{c}{\multirow{1}{*}{10-Shot}} &
    \multicolumn{1}{c|}{\multirow{1}{*}{30-Shot}} \\
    \midrule
    Faster R-CNN
        & \multirow{3}{*}{1}
        & $17.3 \pm 0.1$
        & $22.0 \pm 0.2$
        & $27.0 \pm 0.4$
        & $9.8 \pm 0.3$
        & $10.0 \pm 0.4$
        & $10.8 \pm 0.3$
        & $1.9 \pm 0.3$
        & $2.7 \pm 0.1$
        & $3.5 \pm 0.1$
        & $7.8 \pm 0.2$
        & $8.2 \pm 0.3$
        & $9.0 \pm 0.2$\\
    Soft Teacher
        &
        & $27.8 \pm 0.8$
        & $32.4 \pm 0.8$
        & $38.1 \pm 0.9$
        & $19.4 \pm 0.7$
        & $19.9 \pm 0.8$
        & $21.2 \pm 0.7$
        & $5.9 \pm 0.8$
        & $7.9 \pm 0.7$
        & $10.1 \pm 0.5$
        & $16.0 \pm 0.6$
        & $16.9 \pm 0.7$
        & $18.4 \pm 0.6$\\
    \cellcolor{Gray}{\model}
        & \cellcolor{Gray}
        & \cellcolor{Gray}$28.9 \pm 0.7$
        & \cellcolor{Gray}$33.7 \pm 0.6$
        & \cellcolor{Gray}$39.4 \pm 0.6$
        & \cellcolor{Gray}$19.2 \pm 0.6$
        & \cellcolor{Gray}$19.8 \pm 0.5$
        & \cellcolor{Gray}$21.1 \pm 0.5$
        & \cellcolor{Gray}$6.7 \pm 0.3$
        & \cellcolor{Gray}$8.8 \pm 0.2$
        & \cellcolor{Gray}$10.8 \pm 0.5$
        & \cellcolor{Gray}$16.1 \pm 0.5$
        & \cellcolor{Gray}$17.1 \pm 0.4$
        & \cellcolor{Gray}$18.5 \pm 0.5$\\
    \midrule
    Faster R-CNN
        & \multirow{3}{*}{5}
        & $23.3 \pm 0.3$
        & $28.7 \pm 0.4$
        & $34.9 \pm 0.5$
        & $18.5 \pm 0.5$
        & $18.9 \pm 0.3$
        & $20.0 \pm 0.5$
        & $3.5 \pm 0.2$
        & $4.6 \pm 0.2$
        & $5.9 \pm 0.3$
        & $14.8 \pm 0.4$
        & $15.3 \pm 0.2$
        & $16.5 \pm 0.4$\\
    Soft Teacher
        &
        & $29.8 \pm 0.2$
        & $35.2 \pm 0.2$
        & $41.4 \pm 0.3$
        & $27.5 \pm 0.4$
        & $27.8 \pm 0.5$
        & $29.2 \pm 0.5$
        & $6.7 \pm 0.7$
        & $8.9 \pm 0.4$
        & $11.1 \pm 0.3$
        & $22.3 \pm 0.4$
        & $23.1 \pm 0.3$
        & $24.7 \pm 0.4$\\
    \cellcolor{Gray}{\model}
        & \cellcolor{Gray}
        & \cellcolor{Gray}$30.5 \pm 0.2$
        & \cellcolor{Gray}$35.9 \pm 0.2$
        & \cellcolor{Gray}$42.0 \pm 0.2$
        & \cellcolor{Gray}$27.5 \pm 0.4$
        & \cellcolor{Gray}$27.9 \pm 0.4$
        & \cellcolor{Gray}$29.3 \pm 0.2$
        & \cellcolor{Gray}$7.9 \pm 0.4$
        & \cellcolor{Gray}$10.1 \pm 0.5$
        & \cellcolor{Gray}$12.4 \pm 0.5$
        & \cellcolor{Gray}$22.6 \pm 0.3$
        & \cellcolor{Gray}$23.4 \pm 0.3$
        & \cellcolor{Gray}$25.1 \pm 0.2$\\
    \midrule
    Faster R-CNN
        & \multirow{3}{*}{10}
        & $25.0 \pm 0.2$
        & $30.7 \pm 0.3$
        & $37.5 \pm 0.3$
        & $22.6 \pm 0.4$
        & $22.8 \pm 0.1$
        & $24.2 \pm 0.2$
        & $3.8 \pm 0.5$
        & $5.3 \pm 0.2$
        & $6.8 \pm 0.2$
        & $17.9 \pm 0.3$
        & $18.4 \pm 0.1$
        & $19.9 \pm 0.2$\\
    Soft Teacher
        &
        & $30.2 \pm 0.2$
        & $35.9 \pm 0.2$
        & $42.4 \pm 0.2$
        & $30.5 \pm 0.5$
        & $30.7 \pm 0.4$
        & $32.1 \pm 0.3$
        & $6.8 \pm 0.3$
        & $9.0 \pm 0.6$
        & $11.4 \pm 0.3$
        & $24.6 \pm 0.4$
        & $25.3 \pm 0.4$
        & $26.9 \pm 0.3$\\
    \cellcolor{Gray}{\model}
        & \cellcolor{Gray}
        & \cellcolor{Gray}$31.1 \pm 0.2$
        & \cellcolor{Gray}$36.7 \pm 0.2$
        & \cellcolor{Gray}$43.1 \pm 0.3$
        & \cellcolor{Gray}$30.3 \pm 0.5$
        & \cellcolor{Gray}$30.6 \pm 0.5$
        & \cellcolor{Gray}$32.0 \pm 0.4$
        & \cellcolor{Gray}$7.9 \pm 1.3$
        & \cellcolor{Gray}$10.4 \pm 1.1$
        & \cellcolor{Gray}$12.9 \pm 1.0$
        & \cellcolor{Gray}$24.6 \pm 0.1$
        & \cellcolor{Gray}$25.6 \pm 0.3$
        & \cellcolor{Gray}$27.2 \pm 0.3$\\
    \bottomrule
    \end{tabular}
}
\label{tab::proposal-quality-supp}
\end{table*}

\subsection{\model Improves Precision and Recall for Semi-Supervised Detection}\label{sec::ssod}

\begin{table*}[t]
\begin{minipage}[t]{0.48\textwidth}
\captionof{table}{SSOD results on VOC07 \texttt{test}. \texttt{VOC0712} is the combined VOC07$+$12 \texttt{trainval} splits. \texttt{COCO-20} is the subset of \texttt{COCO-train2017} having the same 20 classes as VOC. \model outperforms Humble Teacher and Soft Teacher by a convincing margin.}
\centering
\resizebox{\columnwidth}{!}{
    \begin{tabular}{llccccc}
	\toprule
	Method & 
	\# Labels &
        Unlabeled &
	AP$_{50}$ &
	AP$_{50:95}$ &
	AR$_{50}$ &
	AR$_{50:95}$ \\
        \midrule
        Supervised~\citep{humble-teacher}
        & \multirow{1}{*}{VOC07 (5K)}
	& \multirow{2}{*}{None}
        & 76.30
        & 42.60
        & --
        & --\\
        \cellcolor{Gray}{Supervised (Our Impl.)}
	& \multirow{1}{*}{VOC07 (5K)}
	& 
        & \cellcolor{Gray}{\bfseries 79.34}
        & \cellcolor{Gray}{\bfseries 49.20}
        & \cellcolor{Gray}{\bfseries 85.38}
        & \cellcolor{Gray}{\bfseries 57.50}\\
        \midrule
        Supervised~\citep{humble-teacher}
	& \multirow{1}{*}{VOC0712 (16K)}
	& \multirow{2}{*}{None}
        & 82.17
        & 54.29
        & --
        & --\\
        \cellcolor{Gray}{Supervised (Our Impl.)}
	& \multirow{1}{*}{VOC0712 (16K)}
	& 
        & \cellcolor{Gray}{\bfseries 84.53}
        & \cellcolor{Gray}{\bfseries 57.77}
        & \cellcolor{Gray}{\bfseries 89.73}
        & \cellcolor{Gray}{\bfseries 65.73}\\
        \midrule
        Humble Teacher [CVPR'21]
	& \multirow{3}{*}{VOC07 (5K)}
	& \multirow{3}{*}{VOC12}
        & 80.94
        & \bfseries 53.04
        & --
        & --\\
        Soft Teacher (Our Impl.)
	& 
	& 
        & 82.37
        & 51.10
        & 88.44
        & 59.49\\
        \cellcolor{Gray}{\model (Ours)}
	& 
	& 
        & \cellcolor{Gray}{\bfseries 83.10}
        & \cellcolor{Gray}{51.26}
        & \cellcolor{Gray}{\bfseries 89.74}
        & \cellcolor{Gray}{\bfseries 60.19}\\
        \midrule
        Humble Teacher [CVPR'21]
	& \multirow{3}{*}{VOC07 (5K)}
        & VOC12
        & 81.29
        & 54.41
        & --
        & --\\
        Soft Teacher (Our Impl.)
	& 
	& $+$
        & 82.50
        & 54.47
        & 87.14
        & 62.45\\
        \cellcolor{Gray}{\model (Ours)}
	& 
	& COCO-20
        & \cellcolor{Gray}{\bfseries 84.09}
        & \cellcolor{Gray}{\bfseries 55.34}
        & \cellcolor{Gray}{\bfseries 88.90}
        & \cellcolor{Gray}{\bfseries 63.58}\\
        \bottomrule
    \end{tabular}
}
\label{tab::ssod-voc}
\end{minipage} \hfill
\begin{minipage}[t]{0.48\textwidth}
\captionof{table}{SSOD results on COCO \texttt{val2017}. The $\dag$ setting refers to self-augmented supervised training without unlabeled data, and $\ddag$ refers to the use of extra \texttt{unlabeled2017} images. We report the mean and standard deviation computed over 5 random samples.}
\centering
\resizebox{\columnwidth}{!}{
    \begin{tabular}{lccccc}
    \toprule
    \textbf{COCO \texttt{val2017}}
    & \multicolumn{5}{c}{Average Precision (AP$_{50:95}$)}\\
    \cmidrule{2-6}
    \multirow{1}{*}{Method} &
    \multirow{1}{*}{1\%} &
    \multirow{1}{*}{5\%} &
    \multirow{1}{*}{10\%} &
    \multirow{1}{*}{$^\dag$100\%} &
    \multirow{1}{*}{$^\ddag$100\%} \\
    \midrule
    Supervised (Our Impl.)
        & $10.57 \pm 0.32$
	& $21.33 \pm 0.40$
	& $26.80 \pm 0.26$
	& $41.96$
	& $41.96$\\
    Humble Teacher [CVPR'21]
        & $16.96 \pm 0.38$
	& $27.70 \pm 0.15$
	& $31.61 \pm 0.28$
	& --
	& $42.37$\\
    Soft Teacher (Our Impl.)
        & $21.38 \pm 1.02$
	& $30.65 \pm 0.19$
	& $33.95 \pm 0.13$
	& $43.51$
	& $44.08$\\
    \cellcolor{Gray}{\model (Ours)}
        & \cellcolor{Gray}{\bfseries 21.93 $\pm$ 0.90}
	& \cellcolor{Gray}{\bfseries 31.15 $\pm$ 0.29}
	& \cellcolor{Gray}{\bfseries 34.08 $\pm$ 0.05}
	& \cellcolor{Gray}{\bfseries 44.05}
	& \cellcolor{Gray}{\bfseries 44.22}\\
    \end{tabular}
}
\resizebox{\columnwidth}{!}{
    \begin{tabular}{lccccc}
    \toprule
    \multirow{2}{*}{Method}
    & \multicolumn{5}{c}{Average Recall (AR$_{50:95}$)}\\
    \cmidrule{2-6}
    & 
    \multirow{1}{*}{1\%} &
    \multirow{1}{*}{5\%} &
    \multirow{1}{*}{10\%} &
    \multirow{1}{*}{$^\dag$100\%} &
    \multirow{1}{*}{$^\ddag$100\%} \\
    \midrule
    Supervised (Our Impl.)
        & $15.87 \pm 0.45$
	& $29.07 \pm 0.47$
	& $36.80 \pm 0.46$
	& $55.64$
	& $55.64$\\
    Soft Teacher (Our Impl.)
        & $29.85 \pm 0.89$
	& $38.68 \pm 0.28$
	& $43.48 \pm 0.25$
	& $55.66$
	& $56.18$\\
    \cellcolor{Gray}{\model (Ours)}
        & \cellcolor{Gray}{\bfseries 30.90 $\pm$ 0.88}
	& \cellcolor{Gray}{\bfseries 39.60 $\pm$ 0.41}
	& \cellcolor{Gray}{\bfseries 43.90 $\pm$ 0.55}
	& \cellcolor{Gray}{\bfseries 56.06}
	& \cellcolor{Gray}{\bfseries 56.22}\\
    \bottomrule
    \end{tabular}
}
\label{tab::ssod-coco}
\end{minipage}
\end{table*}

We present SSOD results for VOC and COCO in \Cref{tab::ssod-voc,tab::ssod-coco}, respectively. We leverage \texttt{COCO-20} and \texttt{COCO-unlabeled2017} as unlabeled images on VOC and COCO experiments, respectively, to compare with existing work. On both datasets, we re-implement and re-train the supervised and Soft Teacher models for a direct comparison with \model. As part of our re-implementation, we make a conscientious effort to obtain high-quality supervised and Soft Teacher baselines by maximizing their performance output. This is to ensure that any performance boost demonstrated by \model is directly attributed to our entropy regression module for proposal learning with complex affine transforms.

In \Cref{tab::ssod-voc}, we compare our best-case supervised baselines to those trained by Humble Teacher~\citep{humble-teacher} and show that ours achieve significantly better detection accuracy. Even in the presence of strong supervised and semi-supervised baselines, our \model model continues to improve upon its counterparts across almost all AP and AR metrics. Notably, our approach demonstrates superior learning with unlabeled data by narrowing the gap to less than $0.5$ AP$_{50}$ between the fully supervised model trained on VOC07$+$12 (16K labels) and \model trained on VOC07 (5K labels) augmented with unlabeled images from VOC12$+$COCO-20.

In \Cref{tab::ssod-coco}, our model consistently outperforms its Soft Teacher counterpart over varying fractions of labeled data, although the impact of proposal learning in \model diminishes as the percentage of labeled data increases. We also experiment with 100\% labels, \ie, the entire \verb|train2017| set, in two settings. In the first setting without unlabeled data, referred to as \emph{self-augmented supervised training}, we use the \verb|train2017| set as the source of ``unlabeled data'' to generate pseudo targets. And in the second setting, we supplement supervised training with \verb|unlabeled2017| images. We observe that even without unlabeled data, \model improves on the supervised baseline by $+2.09$ AP, suggesting that more representations can still be learned from \verb|train2017| alone. In the setting with additional unlabeled data, our model further boosts accuracy by another $+0.17$ AP.

\begin{figure*}[t]
  \centering
  \setlength\tabcolsep{1.0pt}
  \begin{tabular}{ccc}
    \rotatebox[origin=c]{90}{\small 1\% of Labels}
    \makecell{\includegraphics[width=0.25\linewidth]{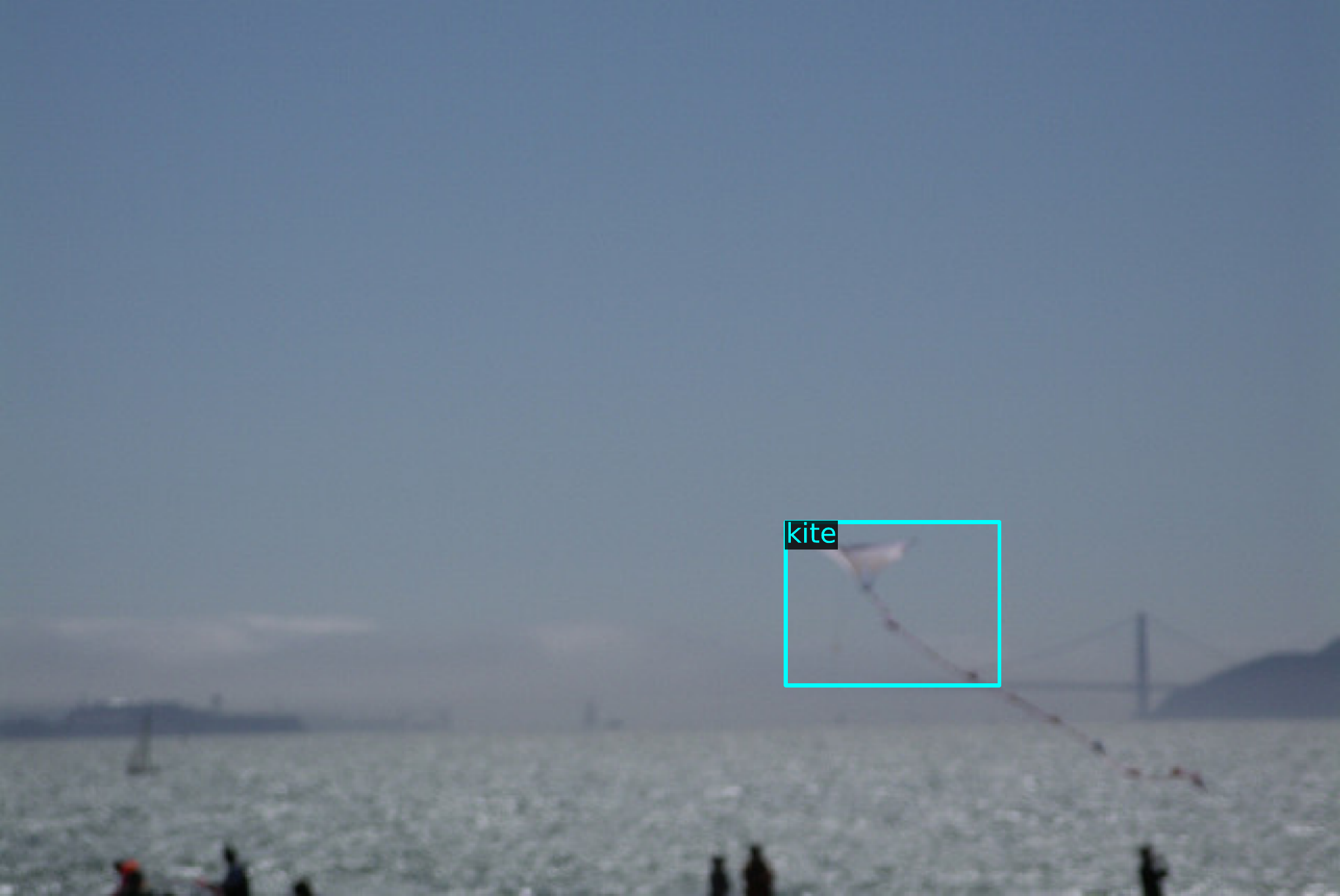}}&
    \makecell{\includegraphics[width=0.25\linewidth]{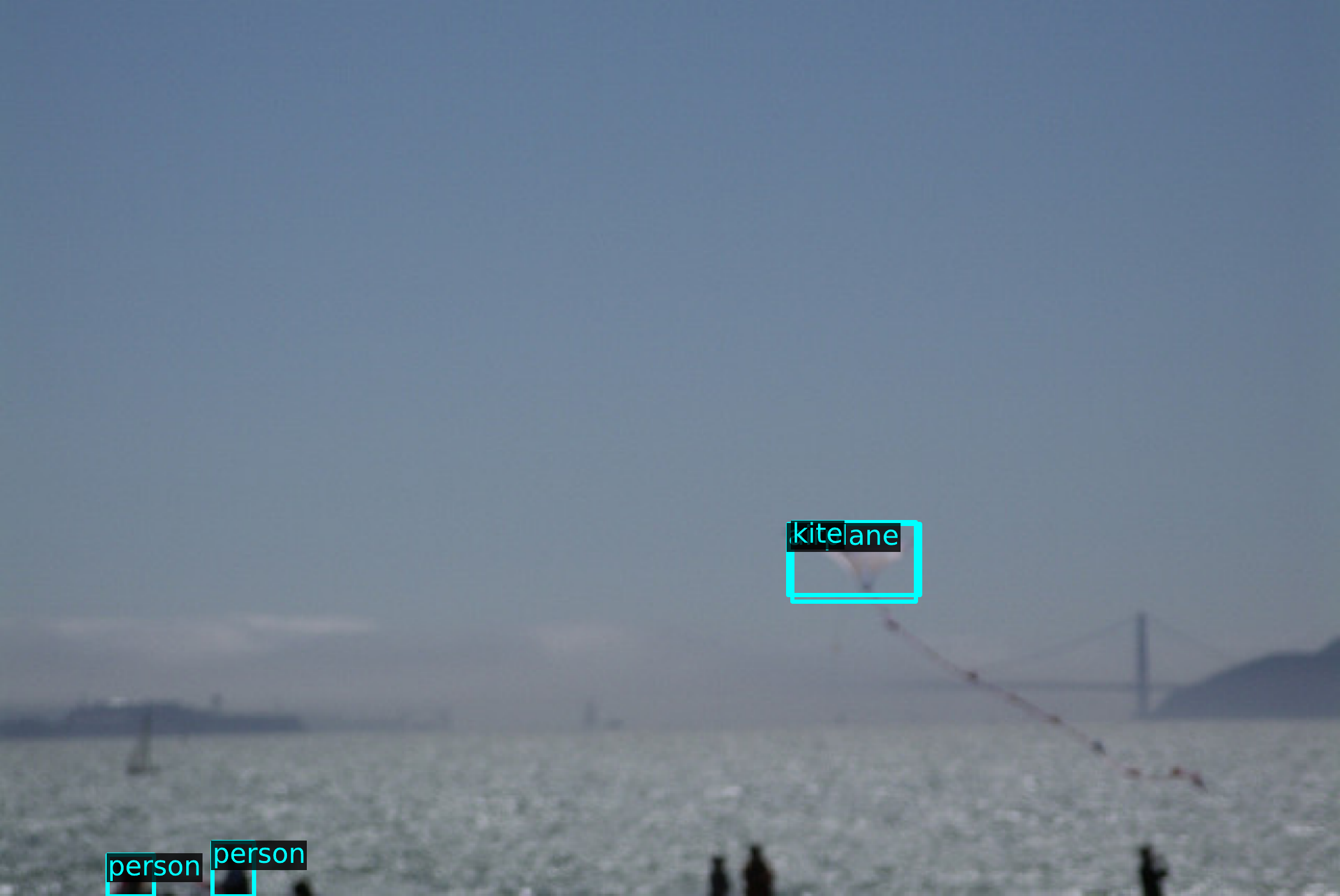}}&
    \makecell{\includegraphics[width=0.25\linewidth]{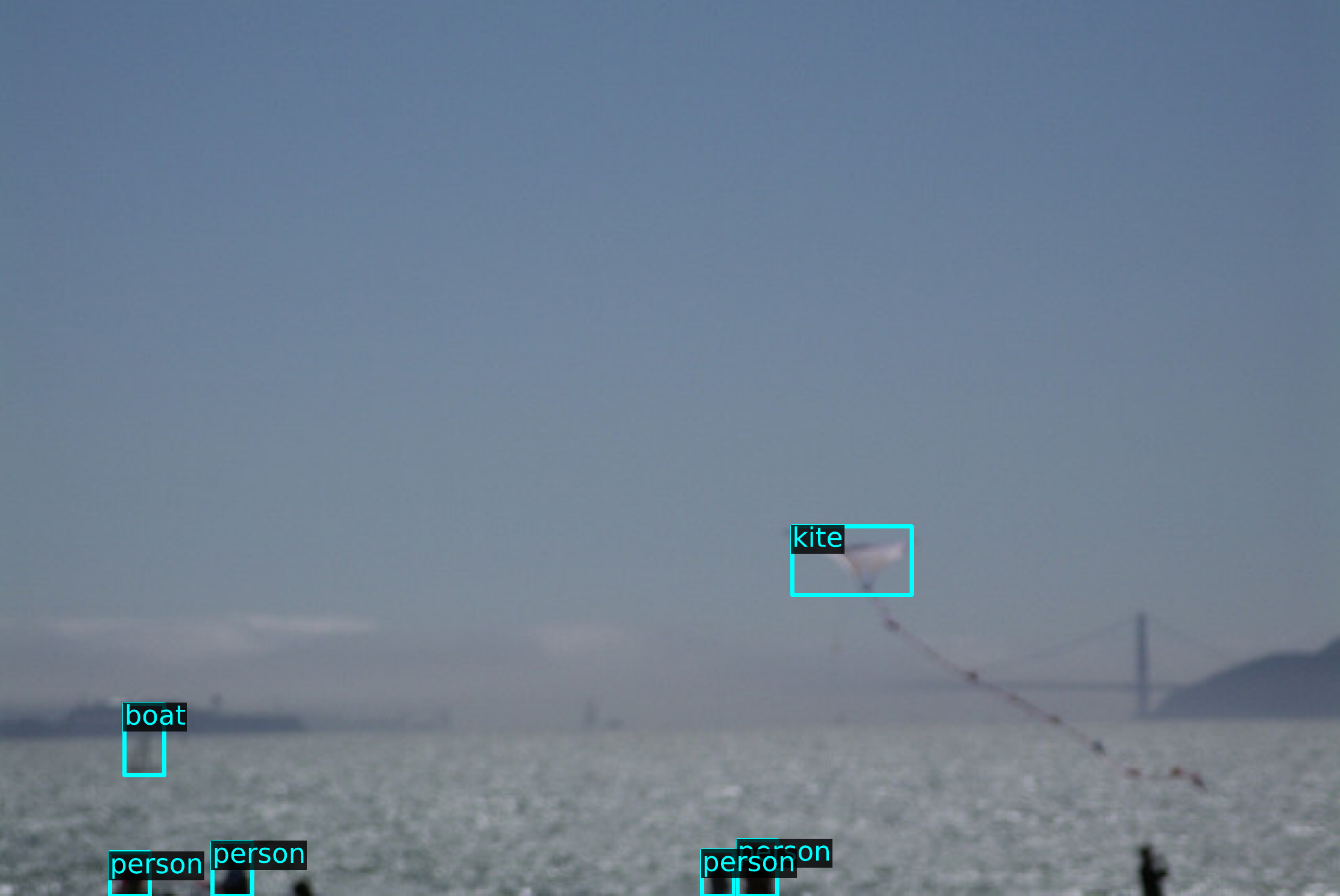}}\\
    [-1.5pt]
    \rotatebox[origin=c]{90}{\small 1\% of Labels}
    \makecell{\includegraphics[width=0.25\linewidth]{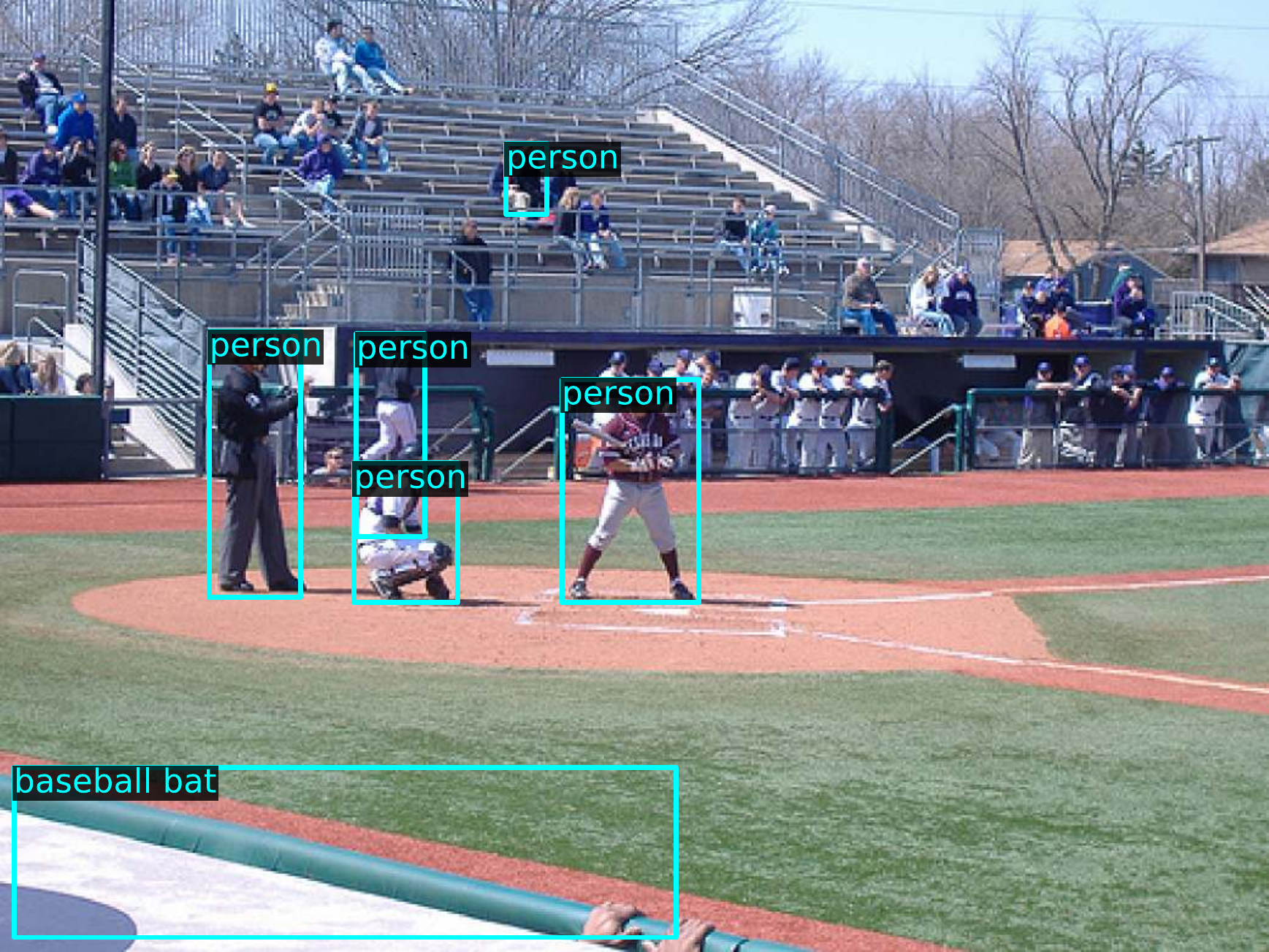}}&
    \makecell{\includegraphics[width=0.25\linewidth]{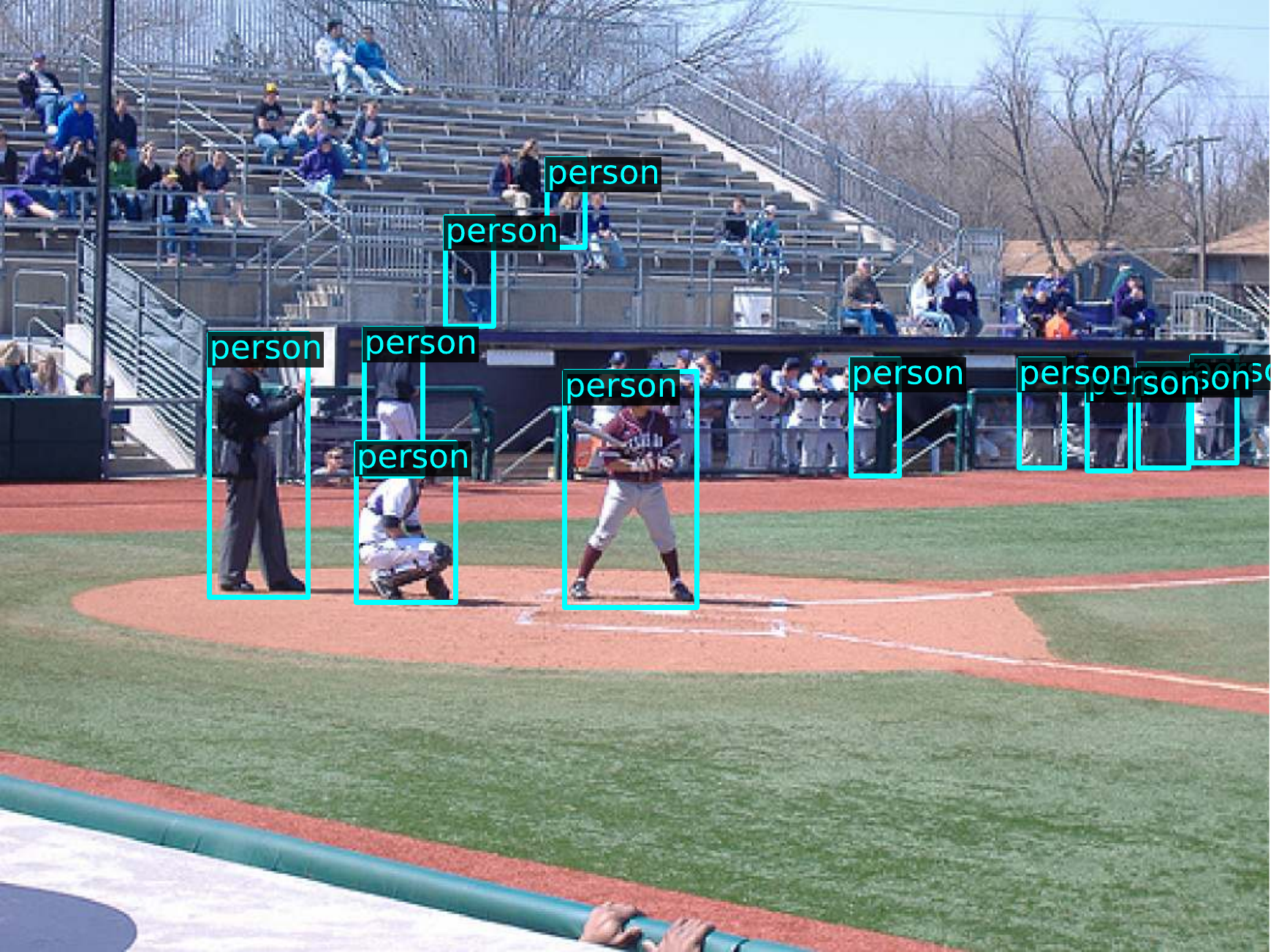}}&
    \makecell{\includegraphics[width=0.25\linewidth]{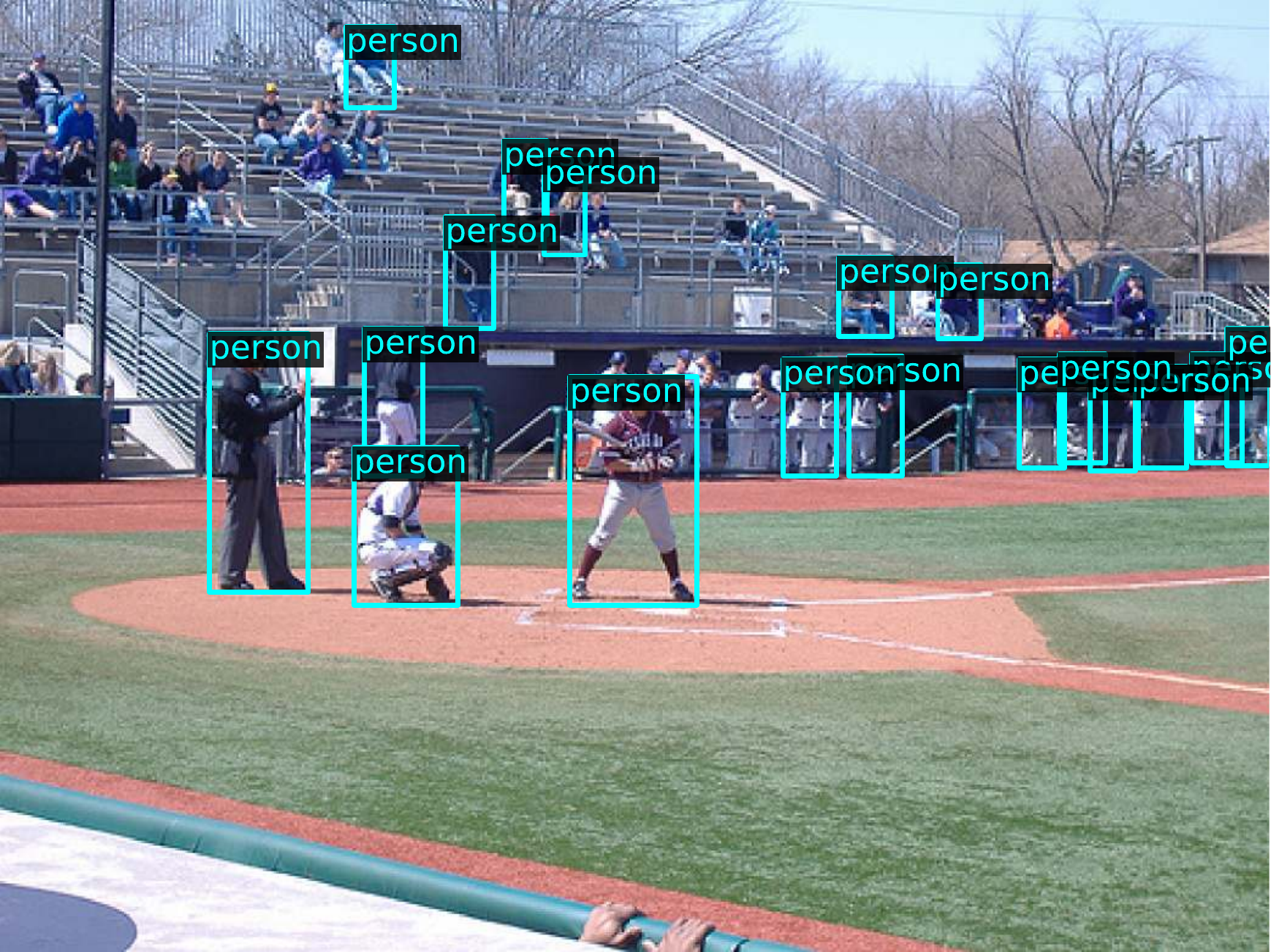}}\\
    \small Supervised & \small Soft Teacher & \small \model (Ours)
  \end{tabular}
  \caption{Qualitative detections on COCO \texttt{val2017} from models trained on 1\% of labels. \model improves on both precision and recall, by recovering more missed objects while making fewer false positive detections, over its corresponding supervised and Soft Teacher counterparts. Best viewed digitally.}
  \label{fig::qualitative-vis}
\end{figure*}

\Cref{fig::qualitative-vis} visualizes exemplar detections from models trained on 1\% of COCO labels, wherein our \model improves on both precision and recall over the comparisons.

\begin{table*}[t]
\caption{FSOD results evaluated on COCO \texttt{val2017}. We report the mean and 95\% confidence interval over 5 random samples for our models. \model with ResNet-50 surpasses TFA with ResNet-101 on both base and novel performances while also uniformly outperforming its Soft Teacher counterpart across all experiments.}
\centering
\resizebox{\textwidth}{!}{
    \begin{tabular}{lccc|cccc|cccc|}
    \toprule
    \textbf{COCO \texttt{val2017}} &
    \multirow{2}{*}{Backbone} &
    \multirow{2}{*}{Base AP$_{50:95}$} &
    \multirow{2}{*}{Base AR$_{50:95}$} &
    \multicolumn{4}{c|}{Base AP$_{50:95}$ (60 Classes)} &
    \multicolumn{4}{c|}{Novel AP$_{50:95}$ (20 Classes)}\\
    \cmidrule{5-12}
    \multirow{1}{*}{Method} &
    & & &
    \multirow{1}{*}{1-Shot} &
    \multirow{1}{*}{5-Shot} &
    \multirow{1}{*}{10-Shot} &
    \multirow{1}{*}{30-Shot} &
    \multirow{1}{*}{1-Shot} &
    \multirow{1}{*}{5-Shot} &
    \multirow{1}{*}{10-Shot} &
    \multirow{1}{*}{30-Shot}\\
    \midrule
        TFA w/cos
        & R-101
        & $39.3$
        & --
        & $31.9 \pm 0.7$
        & $32.3 \pm 0.6$
        & $32.4 \pm 0.6$
        & $34.2 \pm 0.4$
        & $1.9 \pm 0.4$
        & $7.0 \pm 0.7$
        & ~~$9.1 \pm 0.5$
        & $12.1 \pm 0.4$\\
    \midrule
        Faster R-CNN (Our Impl.)
        & R-50
        & $39.3$
        & $53.0$
        & $34.4 \pm 0.6$
        & $33.1 \pm 0.2$
        & $33.2 \pm 0.2$
        & $35.1 \pm 0.3$
        & $1.0 \pm 0.3$
        & $5.1 \pm 0.4$
        & ~~$7.2 \pm 0.4$
        & ~~$9.6 \pm 0.2$\\
        Soft Teacher (Our Impl.)
        & R-50
        & $41.3$
        & $52.8$
        & $37.6 \pm 0.4$
        & $37.0 \pm 0.1$
        & $36.8 \pm 0.3$
        & $38.2 \pm 0.3$
        & $1.7 \pm 0.9$
        & $6.7 \pm 0.4$
        & ~~$8.8 \pm 0.5$
        & $11.2 \pm 0.4$\\
        \cellcolor{Gray}\model (Ours)
        & \cellcolor{Gray}R-50
        & \cellcolor{Gray}42.0
        & \cellcolor{Gray}54.4
        & \cellcolor{Gray}$38.0 \pm 0.4$
        & \cellcolor{Gray}$37.4 \pm 0.2$
        & \cellcolor{Gray}$37.4 \pm 0.2$
        & \cellcolor{Gray}$38.7 \pm 0.2$
        & \cellcolor{Gray}$2.4 \pm 0.6$
        & \cellcolor{Gray}$8.2 \pm 0.3$
        & \cellcolor{Gray}$10.3 \pm 0.5$
        & \cellcolor{Gray}$12.9 \pm 0.6$\\
        \cellcolor{Gray}\model (Ours)
        & \cellcolor{Gray}R-101
        & \cellcolor{Gray}\bfseries 44.4
        & \cellcolor{Gray}\bfseries 56.1
        & \cellcolor{Gray}\bfseries 40.7 $\pm$ 0.3
        & \cellcolor{Gray}\bfseries 40.3 $\pm$ 0.2
        & \cellcolor{Gray}\bfseries 40.2 $\pm$ 0.3
        & \cellcolor{Gray}\bfseries 41.4 $\pm$ 0.2
        & \cellcolor{Gray}\bfseries 2.8 $\pm$ 0.7
        & \cellcolor{Gray}\bfseries 8.7 $\pm$ 0.6
        & \cellcolor{Gray}\bfseries 11.0 $\pm$ 0.4
        & \cellcolor{Gray}\bfseries 14.0 $\pm$ 0.6\\
    \bottomrule
    \end{tabular}
}
\label{tab::fsod-coco}
\end{table*}

\subsection{Generalized Few-Shot Detection on MS-COCO}
We present additional FSOD results on the COCO dataset to include 1-shot detection in \Cref{tab::fsod-coco}, leveraging \texttt{COCO-unlabeled2017} as supplementary unlabeled data. Here, we observe more supporting evidence to strengthen our empirical finding on the potential relationship between SSOD and FSOD to suggest that a stronger semi-supervised detector leads to a more label-efficient few-shot detector. \model uniformly outperforms Soft Teacher across all metrics and experiments under consideration, most notably on novel class detection.

\subsection{\model is Less Prone to Overfitting}
\begin{figure*}[t]
\centering
\includegraphics[width=0.95\textwidth]{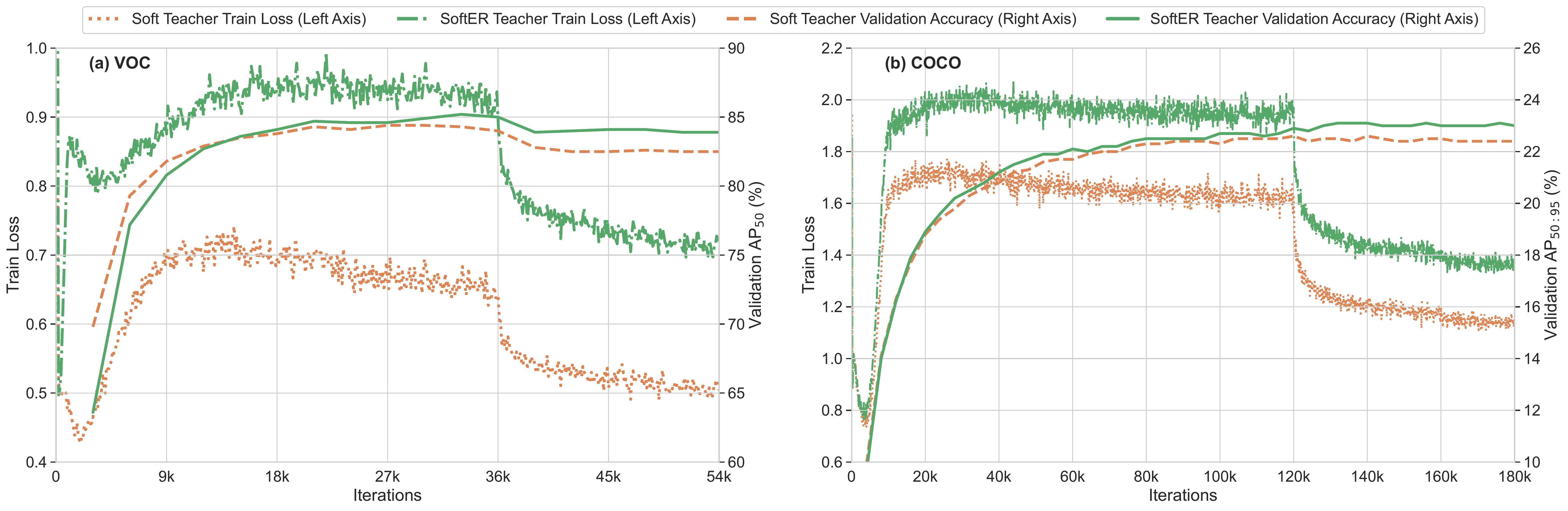}
\caption{Visualization of training and validation behavior of Soft Teacher and \model on \textbf{(a)} VOC07 and \textbf{(b)} 1\% of COCO labels. \textbf{Left:} The validation curve of Soft Teacher tends to overfit more than \model toward the end of training on VOC. \textbf{Right:} \model continues to improve even when Soft Teacher has reached its validation performance plateau at the 120k iterations mark.}
\label{fig::train-loss-val-acc}
\end{figure*}

We analyze the training behavior of Soft Teacher and \model for semi-supervised detection in \Cref{fig::train-loss-val-acc}. For VOC, we train both models on VOC07 \verb|trainval| labels with supplementary unlabeled images from VOC12$+$COCO-20. We observe from the validation curves that Soft Teacher seems to train faster than \model at the beginning, but has the propensity to overfit more than \model toward the end of training. For COCO, we train on 1\% of labels sampled from \verb|train2017| with the remaining 99\% as unlabeled data. Similarly, we see from the validation curves that \model continues to improve even when Soft Teacher has reached its performance plateau. We attribute these characteristics to our entropy regression module for proposal learning, which provides \model a degree of robustness against overfitting.

\begin{table*}[t]
\caption{Ablation experiments evaluated on COCO \texttt{val2017} showing the standard procedure of fine-tuning both box classification and regression heads degrades base performance by as much as 21\%. Our modified protocol of fine-tuning only the box classifier, while keeping the box regressor fixed, helps retain base detection accuracy with a performance drop of less than 11\% for Faster R-CNN and 9\% for \model.}
\centering
\resizebox{\textwidth}{!}{
    \begin{tabular}{lc|rrrr|rrrr|}
    \toprule
    \multirow{2}{*}{Method} &
    \multirow{2}{*}{Base AP$_{50:95}$} &
    \multicolumn{4}{c|}{Base AP$_{50:95}$ (60 Classes)} &
    \multicolumn{4}{c|}{Novel AP$_{50:95}$ (20 Classes)} \\
    \cmidrule{3-10}
    & &
    \multicolumn{1}{c}{\multirow{1}{*}{1-Shot}} &
    \multicolumn{1}{c}{\multirow{1}{*}{5-Shot}} &
    \multicolumn{1}{c}{\multirow{1}{*}{10-Shot}} &
    \multicolumn{1}{c|}{\multirow{1}{*}{30-Shot}} &
    \multicolumn{1}{r}{\multirow{1}{*}{1-Shot}} &
    \multicolumn{1}{r}{\multirow{1}{*}{5-Shot}} &
    \multicolumn{1}{r}{\multirow{1}{*}{10-Shot}} &
    \multicolumn{1}{r|}{\multirow{1}{*}{30-Shot}} \\
    \midrule
    Faster R-CNN (fine-tune \texttt{cls}$+$\texttt{reg})
        & $39.3$
        & $31.2$ ($\downarrow 21\%$)
        & $34.7$ ($\downarrow 12\%$)
        & $34.8$ ($\downarrow 11\%$)
        & $36.7$ ($\downarrow 7\%$)
        & \multicolumn{1}{c}{$0.6$}
        & \multicolumn{1}{c}{$3.9$}
        & \multicolumn{1}{c}{$6.0$}
        & \multicolumn{1}{c|}{$~~7.9$}\\
    \cellcolor{Gray}Faster R-CNN (fine-tune \texttt{cls} only)
        & \cellcolor{Gray}$39.3$
        & \cellcolor{Gray}$34.9$ ($\downarrow 11\%$)
        & \cellcolor{Gray}$35.8$ ($\downarrow ~~9\%$)
        & \cellcolor{Gray}$35.8$ ($\downarrow ~~9\%$)
        & \cellcolor{Gray}$37.1$ ($\downarrow 6\%$)
        & \multicolumn{1}{c}{\cellcolor{Gray}$0.5$}
        & \multicolumn{1}{c}{\cellcolor{Gray}$3.9$}
        & \multicolumn{1}{c}{\cellcolor{Gray}$6.1$}
        & \multicolumn{1}{c|}{\cellcolor{Gray}$~~7.8$}\\
    \model (fine-tune \texttt{cls}$+$\texttt{reg})
        & $42.0$
        & $33.6$ ($\downarrow 20\%$)
        & $37.8$ ($\downarrow 10\%$)
        & $38.1$ ($\downarrow ~~9\%$)
        & $39.9$ ($\downarrow 5\%$)
        & \multicolumn{1}{c}{$1.5$}
        & \multicolumn{1}{c}{$6.7$}
        & \multicolumn{1}{c}{$9.4$}
        & \multicolumn{1}{c|}{$10.8$}\\
    \cellcolor{Gray}{\model (fine-tune \texttt{cls} only)}
        & \cellcolor{Gray}{$42.0$}
        & \cellcolor{Gray}{$38.3$} ($\downarrow ~~9\%$)
        & \cellcolor{Gray}{$39.1$} ($\downarrow ~~7\%$)
        & \cellcolor{Gray}{$39.1$} ($\downarrow ~~7\%$)
        & \cellcolor{Gray}{$40.2$} ($\downarrow 4\%$)
        & \multicolumn{1}{c}{\cellcolor{Gray}{$1.5$}}
        & \multicolumn{1}{c}{\cellcolor{Gray}{$6.7$}}
        & \multicolumn{1}{c}{\cellcolor{Gray}{$9.4$}}
        & \multicolumn{1}{c|}{\cellcolor{Gray}{$10.8$}}\\
    \bottomrule
    \end{tabular}
}
\label{tab::ablation-few-shot-finetune}
\end{table*}

\subsection{To Freeze or Not to Freeze Box Regressor}\label{sec::freeze-box-regressor}
The standard two-stage transfer learning procedure~\citep{tfa} fine-tunes the few-shot detector by updating both the RoI box classifier and regressor while keeping everything else frozen. Intuitively, we expect the RPN to produce accurate object regions during base pre-training, especially in the semi-supervised setting where it is further boosted by supplementary unlabeled images. We postulate that only the box classifier needs to be updated during fine-tuning to adapt base representations to novel concepts, and that fine-tuning the regression head is not necessary and may even hurt base performance. \Cref{tab::ablation-few-shot-finetune} verifies our intuition that fine-tuning both box classification and regression heads degrades base performance by as much as 21\% on COCO \verb|val2017|. By comparison, our modified protocol of fine-tuning only the box classifier helps retain base performance with a drop of less than 11\%. Novel performance is unaffected between the two configurations. Our results are corroborated by existing work confirming that the main source of error with FSOD is indeed associated with the box classifier~\citep{fsce,dcfs}. Recall our goal for FSOD is to maximize novel detection accuracy while minimizing base performance degradation; keeping the box localization parameters fixed during fine-tuning is a simple and straight-forward way to help maintain base class accuracy.

\section{Implementation Details}\label{sec::implementation-details}
\subsection{Few-Shot Datasets and Benchmarks}
In this work, we evaluate our approach on the PASCAL VOC and MS-COCO 2017 detection benchmarks. For COCO experiments, the established evaluation protocol is on COCO 2014 following the TFA benchmark. However, both COCO 2014 and 2017 share the same images. The only difference between the two is the number of validation images (41k images for COCO \texttt{val2014} and 5k for COCO \texttt{val2017}). \citet{tfa} created the TFA benchmark by sampling from COCO 2014 a random subset of 5k images for validation and used the rest in the training split. Thus, both train/val splits from COCO 2014 and 2017 should effectively be the same, with minor variance due to the sampling process. Our preliminary experiments on both COCO 2014 (following the TFA splits) and the official COCO 2017 splits verified that the difference is indeed small with less than $\pm 0.3$ AP. The benefit of using the official COCO 2017 splits is to remove the dependency on the random train/val splits created by the TFA benchmark and to maintain uniformity with our proposed semi-supervised FSOD benchmark in \Cref{tab::ledetection-benchmark}. For VOC experiments, we use the same sample splits taken from the TFA benchmark without changes.

\subsection{Data Augmentation}
We summarize the data augmentation strategy used to train Soft Teacher and \model in~\Cref{tab::data-aug}. There are three pipelines or branches of augmentation. The labeled branch uses random resizing and horizontal flipping along with color transformations. The student detector of the unlabeled branch undergoes the full complement of augmentations including strong affine geometric transformations and cutout~\citep{cutout,random-erase}, akin to RandAugment~\citep{randaugment}, whereas the teacher detector leverages only weak resizing and horizontal flipping. At test time, we resize all instances to the shorter side of $800$ resolution, but otherwise do not perform any test-time augmentation, following standard supervised and semi-supervised protocols.

\begin{table*}[t]
\caption{Summary of the data augmentation pipelines used to train Soft Teacher and \model. \textbf{Left:} transformations applied to the student trained on labeled data. \textbf{Middle:} strong augmentation pipeline, including complex affine transforms and cutout, applied to the student trained on unlabeled data. \textbf{Right:} weak augmentation pipeline applied to the teacher trained on unlabeled data.}
\centering
\resizebox{\textwidth}{!}{
    \begin{tabular}{llll}
    \toprule
    Augmentation &
    Student Labeled Branch &
    Student Unlabeled Branch (Strong) &
    Teacher Unlabeled Branch (Weak)\\
    \midrule
    Resize
    & short edge $\in [400,1200]$
    & short edge $\in [400,1200]$
    & short edge $\in [400,1200]$ \\
    Flip
    & $p=0.5$, horizontal
    & $p=0.5$, horizontal
    & $p=0.5$, horizontal \\
    Identity
    & $p=\nicefrac{1}{9}$
    & $p=\nicefrac{1}{9}$
    & \multicolumn{1}{c}{} \\
    AutoContrast
    & $p=\nicefrac{1}{9}$
    & $p=\nicefrac{1}{9}$
    & \multicolumn{1}{c}{} \\
    Equalize
    & $p=\nicefrac{1}{9}$
    & $p=\nicefrac{1}{9}$
    & \multicolumn{1}{c}{} \\
    Solarize
    & $p=\nicefrac{1}{9}$
    & $p=\nicefrac{1}{9}$
    & \multicolumn{1}{c}{} \\
    Color
    & $p=\nicefrac{1}{9}$
    & $p=\nicefrac{1}{9}$
    & \multicolumn{1}{c}{} \\
    Contrast
    & $p=\nicefrac{1}{9}$
    & $p=\nicefrac{1}{9}$
    & \multicolumn{1}{c}{} \\
    Brightness
    & $p=\nicefrac{1}{9}$
    & $p=\nicefrac{1}{9}$
    & \multicolumn{1}{c}{} \\
    Sharpness
    & $p=\nicefrac{1}{9}$
    & $p=\nicefrac{1}{9}$
    & \multicolumn{1}{c}{} \\
    Posterize
    & $p=\nicefrac{1}{9}$
    & $p=\nicefrac{1}{9}$
    & \multicolumn{1}{c}{} \\
    Translation
    & \multicolumn{1}{l}{}
    & $p=\nicefrac{1}{3}$, $(x,y) \in (-0.1,0.1)$
    & \multicolumn{1}{c}{}\\
    Shearing
    & \multicolumn{1}{l}{}
    & $p=\nicefrac{1}{3}$, $(x,y) \in (-30^{\circ}, 30^{\circ})$
    & \multicolumn{1}{c}{} \\
    Rotation
    & \multicolumn{1}{l}{}
    & $p=\nicefrac{1}{3}$, angle $\in (-30^{\circ}, 30^{\circ})$
    & \multicolumn{1}{c}{} \\
    Cutout
    &
    & $n \in [1,5]$, size $\in [0.0,0.2]$
    & \multicolumn{1}{c}{} \\
    \bottomrule
    \end{tabular}
}
\label{tab::data-aug}
\end{table*}
\begin{table*}[t]
\begin{minipage}[t]{0.48\textwidth}
\captionof{table}{Supervised and semi-supervised training protocols on PASCAL VOC. \texttt{COCO-20} is the subset of \texttt{COCO-train2017} containing objects with the same 20 category names as VOC objects. \texttt{Sample Ratio} denotes the blend of (labeled, unlabeled) examples in a mini-batch. All settings are configured for $8\times$ multi-GPU training.}
\centering
\resizebox{\columnwidth}{!}{
    \begin{tabular}{llcccccr}
	\toprule
	Method & 
	Labeled &
        Unlabeled &
        Batch Size &
        Sample Ratio &
        $lr$ &
        $lr$ Step &
        Iterations\\
        \midrule
        Supervised
        & \multirow{1}{*}{VOC07}
	& \multirow{2}{*}{None}
        & 16
        & (16, 0)
        & 0.02
        & (12k, 16k)
        & 18k\\
        Supervised
	& \multirow{1}{*}{VOC0712}
	& 
        & 16
        & (16, 0)
        & 0.02
        & (36k, 48k)
        & 54k\\
        \midrule
        Soft Teacher
	& \multirow{2}{*}{VOC07}
	& \multirow{2}{*}{VOC12}
        & 64
        & (32, 32)
        & 0.01
        & (12k, 16k)
        & 18k\\
        \model
	& 
	& 
        & 64
        & (32, 32)
        & 0.01
        & (12k, 16k)
        & 18k\\
        \midrule
        Soft Teacher
	& \multirow{2}{*}{VOC07}
        & VOC12$+$
        & 64
        & (32, 32)
        & 0.01
        & (36k, 48k)
        & 54k\\
        \model
	& 
	& COCO-20
        & 64
        & (32, 32)
        & 0.01
        & (36k, 48k)
        & 54k\\
        \midrule
        Soft Teacher
	& \multirow{2}{*}{VOC0712}
        & \multirow{2}{*}{COCO-20}
        & 64
        & (32, 32)
        & 0.01
        & (40k, 52k)
        & 60k\\
        \model
	& 
	& 
        & 64
        & (32, 32)
        & 0.01
        & (40k, 52k)
        & 60k\\
        \bottomrule
    \end{tabular}
}
\label{tab::training-protocol-voc}
\end{minipage} \hfill
\begin{minipage}[t]{0.48\textwidth}
\captionof{table}{Supervised and semi-supervised training protocols on COCO 2017. The $\dag$ setting refers to self-augmented supervised training without unlabeled data, and $\ddag$ corresponds to the use of supplementary \texttt{unlabeled2017} images. \texttt{Sample Ratio} denotes the blend of (labeled, unlabeled) examples in a mini-batch. All settings are configured for $8\times$ multi-GPU training.}
\centering
\resizebox{\columnwidth}{!}{
    \begin{tabular}{clccccr}
    \toprule
    \multirow{1}{*}{\% Labeled} &
    \multirow{1}{*}{Method} &
    \multirow{1}{*}{Batch Size} &
    \multirow{1}{*}{Sample Ratio} &
    \multirow{1}{*}{$lr$} &
    \multirow{1}{*}{$lr$ Step} &
    \multirow{1}{*}{Iterations}\\
    \midrule
    \multirow{3}{*}{1} &
    \multirow{1}{*}{Supervised}
    & 8
    & (8, 0)
    & 0.01
    & (120k, 160k)
    & 180k\\
    &
    \multirow{1}{*}{Soft Teacher}
    & 40
    & (8, 32)
    & 0.01
    & (120k, 160k)
    & 180k\\
    &
    \multirow{1}{*}{\model}
    & 40
    & (8, 32)
    & 0.01
    & (120k, 160k)
    & 180k\\
    \cmidrule{1-7}
    \multirow{3}{*}{5} &
    \multirow{1}{*}{Supervised}
    & 8
    & (8, 0)
    & 0.01
    & (120k, 160k)
    & 180k\\
    &
    \multirow{1}{*}{Soft Teacher}
    & 40
    & (8, 32)
    & 0.01
    & (120k, 160k)
    & 180k\\
    &
    \multirow{1}{*}{\model}
    & 40
    & (8, 32)
    & 0.01
    & (120k, 160k)
    & 180k\\
    \cmidrule{1-7}
    \multirow{3}{*}{10} &
    \multirow{1}{*}{Supervised}
    & 8
    & (8, 0)
    & 0.01
    & (120k, 160k)
    & 180k\\
    &
    \multirow{1}{*}{Soft Teacher}
    & 40
    & (8, 32)
    & 0.01
    & (120k, 160k)
    & 180k\\
    &
    \multirow{1}{*}{\model}
    & 40
    & (8, 32)
    & 0.01
    & (120k, 160k)
    & 180k\\
    \cmidrule{1-7}
    \multirow{3}{*}{$^\dag$100} &
    \multirow{1}{*}{Supervised}
    & 16
    & (16, 0)
    & 0.02
    & (480k, 640k)
    & 720k\\
    &
    \multirow{1}{*}{Soft Teacher}
    & 64
    & (32, 32)
    & 0.01
    & (480k, 640k)
    & 720k\\
    &
    \multirow{1}{*}{\model}
    & 64
    & (32, 32)
    & 0.01
    & (480k, 640k)
    & 720k\\
    \cmidrule{1-7}
    \multirow{3}{*}{$^\ddag$100} &
    \multirow{1}{*}{Supervised}
    & 16
    & (16, 0)
    & 0.02
    & (480k, 640k)
    & 720k\\
    &
    \multirow{1}{*}{Soft Teacher}
    & 64
    & (32, 32)
    & 0.01
    & (480k, 640k)
    & 720k\\
    &
    \multirow{1}{*}{\model}
    & 64
    & (32, 32)
    & 0.01
    & (480k, 640k)
    & 720k\\
    \bottomrule
    \end{tabular}
}
\label{tab::training-protocol-coco}
\end{minipage}
\end{table*}

\subsection{Supervised and Semi-Supervised Training}
\paragraph{High-Quality Baselines}
Following existing literature~\citep{stac,ubteacher,humble-teacher,soft-teacher}, we evaluate our approach for semi-supervised detection on VOC and COCO 2017 datasets. On both datasets, we re-implement and re-train the supervised Faster R-CNN and Soft Teacher\footnote{We leverage the original authors' source code made publicly available at \url{https://github.com/microsoft/SoftTeacher}.} models for a direct comparison with \model. As part of our re-implementation, we make a conscientious effort to obtain the best-case supervised and Soft Teacher baselines by maximizing their performance output. We train the strong supervised baseline by using a longer training schedule (see~\Cref{tab::training-protocol-voc,tab::training-protocol-coco}) and applying diverse color augmentations in addition to random resizing and horizontal flipping (see~\Cref{tab::data-aug}). And we re-train Soft Teacher exactly as is according to the authors' source code. This is to ensure that any performance boost demonstrated by \model is directly attributed to our entropy regression module for learning representations from region proposals, and not to changes in model configuration and training protocol.

\paragraph{VOC Evaluation}
We experiment with two supervised settings: (1) using VOC07 \verb|trainval| split as labeled data, and (2) utilizing the joint VOC07$+$12 labeled set as an upper bound for supervised detection performance. We also have two semi-supervised settings: (1) augmenting supervised training on VOC07 with VOC12 as unlabeled data, and (2) leveraging the combined VOC12$+$COCO-20 as unlabeled data. \texttt{COCO-20} is the subset of \texttt{COCO-train2017} having the same 20 category names as VOC objects. Model performance is evaluated on the VOC07 \verb|test| set. Detailed comparative results are given in \Cref{tab::ssod-voc}.

\paragraph{COCO Evaluation}
There are three experimental settings: (1) \emph{Partially labeled}, where we train on $\{1,5,10\}$ percent of labels randomly sampled from the \verb|train2017| split while treating the remaining images as unlabeled data. (2) \emph{Fully labeled}, where we leverage the extra 123k images from the \verb|unlabeled2017| set to supplement supervised training on the entire \verb|train2017|. And (3) \emph{Self-augmented supervised training}, where we apply the \verb|train2017| set, discarding all label information, as the source of ``unlabeled'' data to generate unsupervised pseudo targets. For each setting, we also train on the labeled portion alone, without using unlabeled data, to establish the lower-bound supervised baseline. Model performance is evaluated on the \verb|val2017| set. See \Cref{tab::ssod-coco} for detailed comparative results.

\paragraph{Top-$N$ Proposals}
To learn representations on region proposals, we extract the top 512 proposals, after non-maximum suppression, from each unlabeled image as generated by the student's RPN. Our motivation for selecting the top 512 proposals is to balance the trade-off among accuracy performance, memory requirements, and training duration. Moreover, our choice of $N=512$ is consistent with $N=640$ proposals empirically found by Humble Teacher~\citep{humble-teacher} to be an optimal number with regards to detection accuracy.

\paragraph{Training Parameters}
We summarize our training protocols on VOC and COCO in~\Cref{tab::training-protocol-voc,tab::training-protocol-coco} for the supervised, Soft Teacher, and \model models. In general, Soft Teacher and our \model share the same configuration to ensure we can directly measure the impact of proposal learning and its contribution to detection accuracy. All hyper-parameters related to Soft Teacher remain the same, including the EMA momentum, which defaults to $0.999$ following common practice in the semi-supervised classification literature~\citep{mean-teacher,fixmatch}. We implement our models in MMDetection~\citep{mmdetection} and PyTorch~\citep{pytorch}, and train them using vanilla SGD optimization with momentum and weight decay set to $0.9$ and $0.0001$, respectively. We train on $8\times$ A6000 GPUs each with 48GB of memory. One experiment takes between 12 hours and 10 days to complete, depending on the scope. At test time, we extract the teacher model from the final check-point for evaluation.

\subsection{Semi-Supervised Few-Shot Training}
In this section, we expound on our protocol for semi-supervised few-shot training on VOC and COCO datasets. We conduct our few-shot experiments following the TFA benchmark~\citep{tfa}. The VOC dataset is randomly partitioned into 15 base and 5 novel classes, where there are $k \in \{1, 5, 10\}$ labeled boxes per category sampled from the combined VOC07$+$12 \verb|trainval| splits. This process is repeated three times to create three partitions. And the COCO dataset is divided into 60 base and 20 novel classes having the same VOC category names with $k \in \{1, 5, 10, 30\}$ shots. We leverage \texttt{COCO-train2017} as the source of external unlabeled data to supplement few-shot training on VOC, and \verb|COCO-unlabeled2017| images to augment experiments on COCO.

\paragraph{Semi-Supervised Base Pre-Training}
In the first stage, we train a base detector on base classes, along with the available unlabeled data, according to the formulation described in \Cref{sec::base-pretrain}. For the supervised base detector, we equip Faster R-CNN with the ResNet-101 backbone. For the semi-supervised base detectors, we experiment with Soft Teacher and our proposed \model using the same ResNet-101 backbone. In some experiments, we also employ ResNet-50 to explore parameter-efficient learning with \model. The incorporation of unlabeled data in the base pre-training step achieves two benefits of practical significance in real-world settings. First, our versatile approach helps make the deployment of few-shot applications easier by not strictly depending on an abundance of labels, which is a fundamental limitation of prior work. Second, unlabeled images have the remarkable ability to advance FSOD by way of region proposals, as demonstrated in \Cref{sec::effective-fsod,sec::proposal-quality}, enabling \model to learn more efficiently with reduced labels for both base and novel classes.

\bigskip

\begin{minipage}[t]{0.48\textwidth}
\paragraph{Semi-Supervised Few-Shot Fine-Tuning}
In the second stage, we combine the parameters of the (semi-supervised) base detector with those of the novel detector into the overall few-shot detector and fine-tune it on a small balanced training set of $k$ shots per class containing both base and novel examples. Before fine-tuning, we obtain the parameters of the novel detector in two ways. For VOC, we initialize the parameters of the novel classifier and regressor with normally distributed random values, analogous to TFA. For the COCO dataset, we reuse the base model pre-trained in the first stage, but further train the detector head from scratch on novel classes. We optimize the novel detector on both few-shot and unlabeled examples according to the semi-supervised protocols. At the fine-tuning step, we update only the RoI box classifier of the few-shot detector while freezing all other components, including the box regressor. We justify our approach and design choices with detailed ablation studies presented in \Cref{sec::ablation}. \Cref{tab::training-protocol-fsod} summarizes our few-shot fine-tuning protocol.
\end{minipage} \hfill
\begin{minipage}[t]{0.48\textwidth}
\vspace{-8pt}
\captionof{table}{Protocol for few-shot fine-tuning on VOC and COCO datasets. All settings are configured for $8\times$ multi-GPU training.}
\centering
\resizebox{\textwidth}{!}{
    \begin{tabular}{clrr}
    \toprule
    \multirow{1}{*}{\# Shot} &
    \multirow{1}{*}{Parameter} &
    VOC07$+$12 &
    COCO 2017\\
    \midrule
    \multirow{5}{*}{1} &
    \multirow{1}{*}{Batch Size} & 16 & 16 \\
    &
    \multirow{1}{*}{$lr$} & 0.001 & 0.001 \\
    &
    \multirow{1}{*}{$lr$ Step} & 9k & 14k \\
    &
    \multirow{1}{*}{Iterations} & 10k & 16k \\
    &
    \multirow{1}{*}{Fine-Tune Layer} & \texttt{cls}$+$\texttt{reg} & \texttt{cls} \\
    \midrule
    \multirow{5}{*}{5} &
    \multirow{1}{*}{Batch Size} & 16 & 16 \\
    &
    \multirow{1}{*}{$lr$} & 0.001 & 0.001 \\
    &
    \multirow{1}{*}{$lr$ Step} & 18k & 72k \\
    &
    \multirow{1}{*}{Iterations} & 20k & 80k \\
    &
    \multirow{1}{*}{Fine-Tune Layer} & \texttt{cls}$+$\texttt{reg} & \texttt{cls} \\
    \midrule
    \multirow{5}{*}{10} &
    \multirow{1}{*}{Batch Size} & 16 & 16 \\
    &
    \multirow{1}{*}{$lr$} & 0.001 & 0.001 \\
    &
    \multirow{1}{*}{$lr$ Step} & 36k & 144k \\
    &
    \multirow{1}{*}{Iterations} & 40k & 160k \\
    &
    \multirow{1}{*}{Fine-Tune Layer} & \texttt{cls}$+$\texttt{reg} & \texttt{cls} \\
    \midrule
    \multirow{5}{*}{30} &
    \multirow{1}{*}{Batch Size} & -- & 16 \\
    &
    \multirow{1}{*}{$lr$} & -- & 0.001 \\
    &
    \multirow{1}{*}{$lr$ Step} & -- & 216k \\
    &
    \multirow{1}{*}{Iterations} & -- & 240k \\
    &
    \multirow{1}{*}{Fine-Tune Layer} & -- & \texttt{cls} \\
    \bottomrule
    \end{tabular}
}
\label{tab::training-protocol-fsod}
\end{minipage}

\section{Additional Qualitative Results}

\begin{figure*}[t]
  \centering
  \setlength\tabcolsep{2.0pt}
  \begin{tabular}{cccc}
    \small Student Proposals & \small Teacher Proposals & \small Student Proposals & \small Teacher Proposals\\
    \makecell{\includegraphics[width=0.24\linewidth]{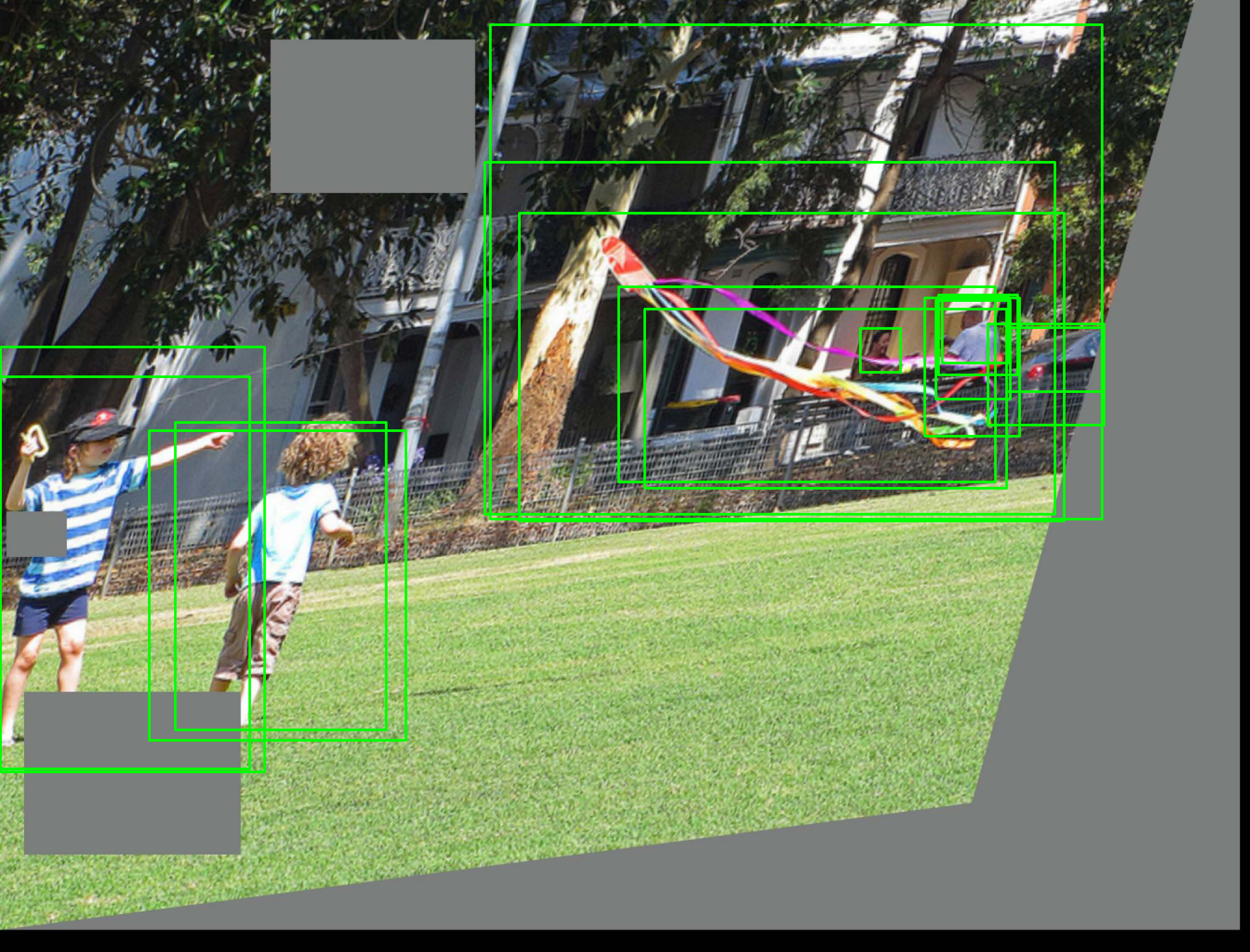}}&
    \makecell{\includegraphics[width=0.24\linewidth]{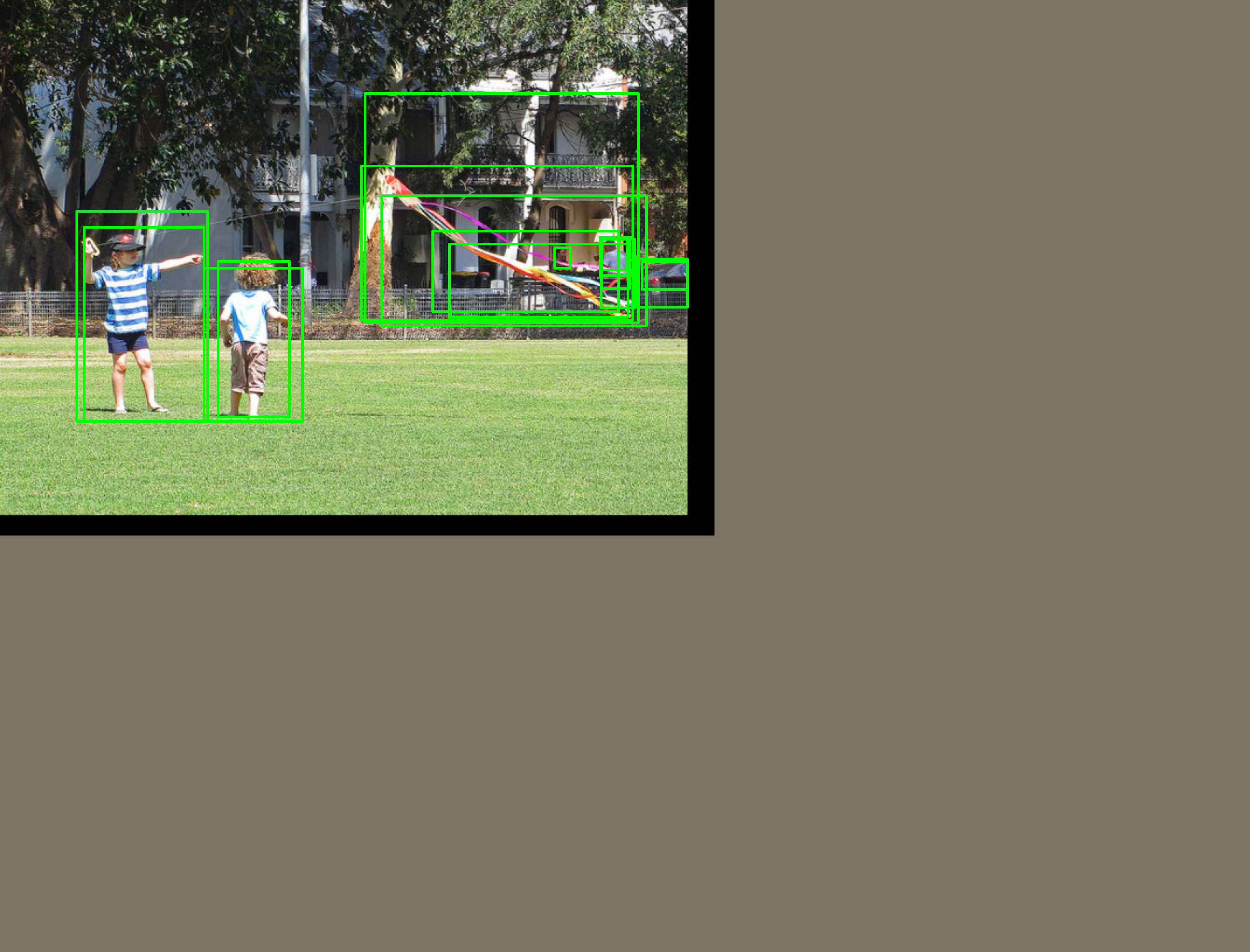}}&
    \makecell{\includegraphics[width=0.24\linewidth]{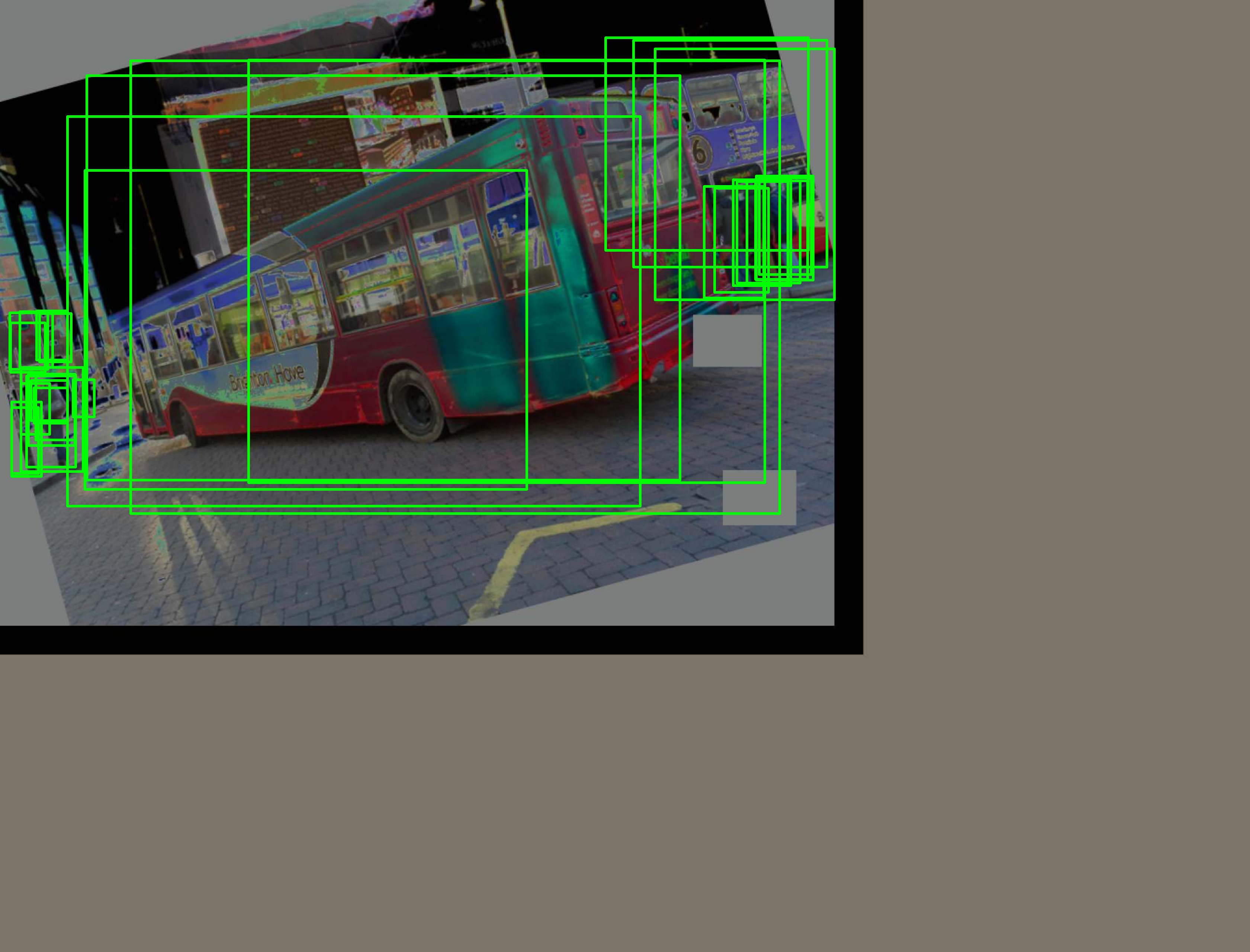}}&
    \makecell{\includegraphics[width=0.24\linewidth]{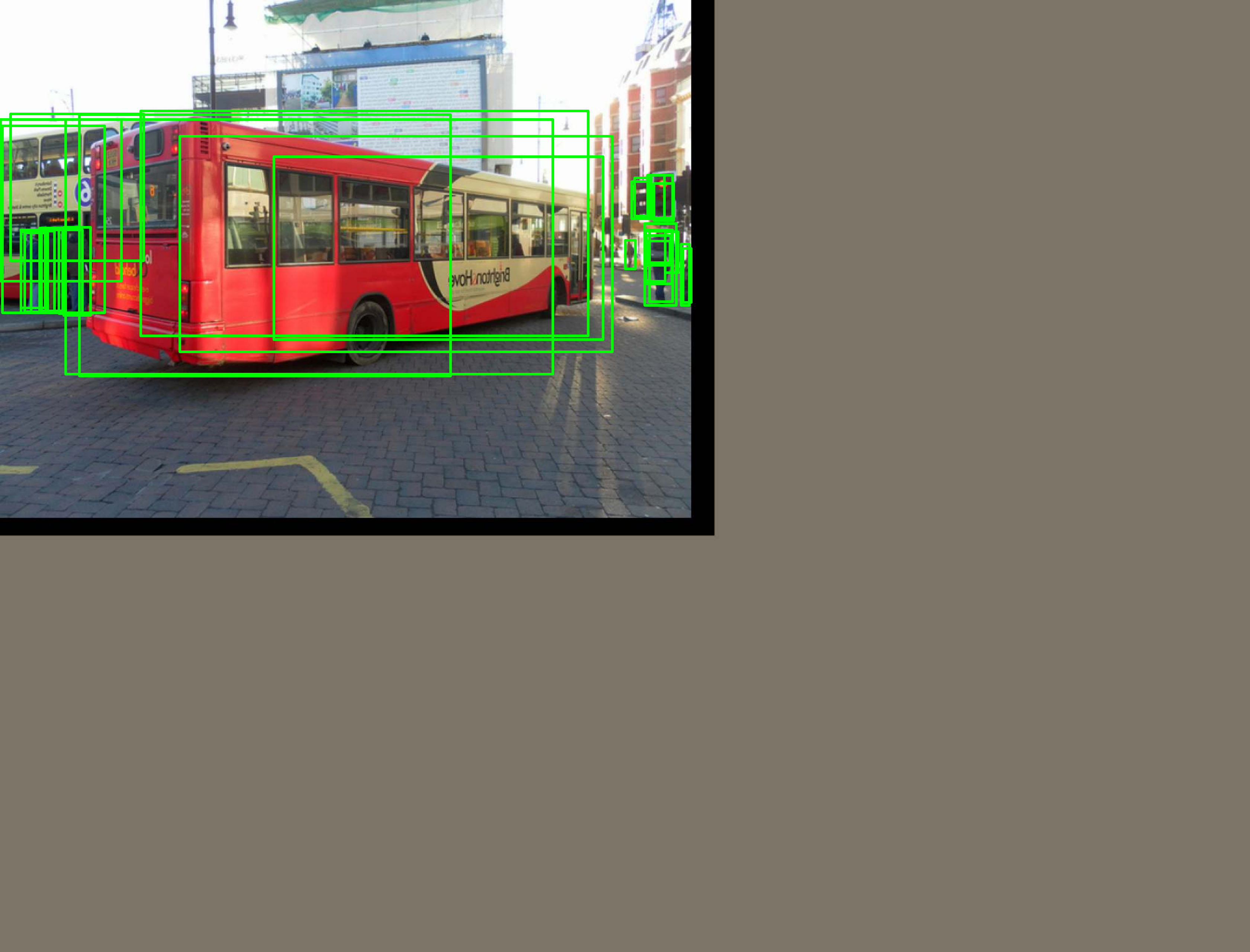}}\\
    \makecell{\includegraphics[width=0.24\linewidth]{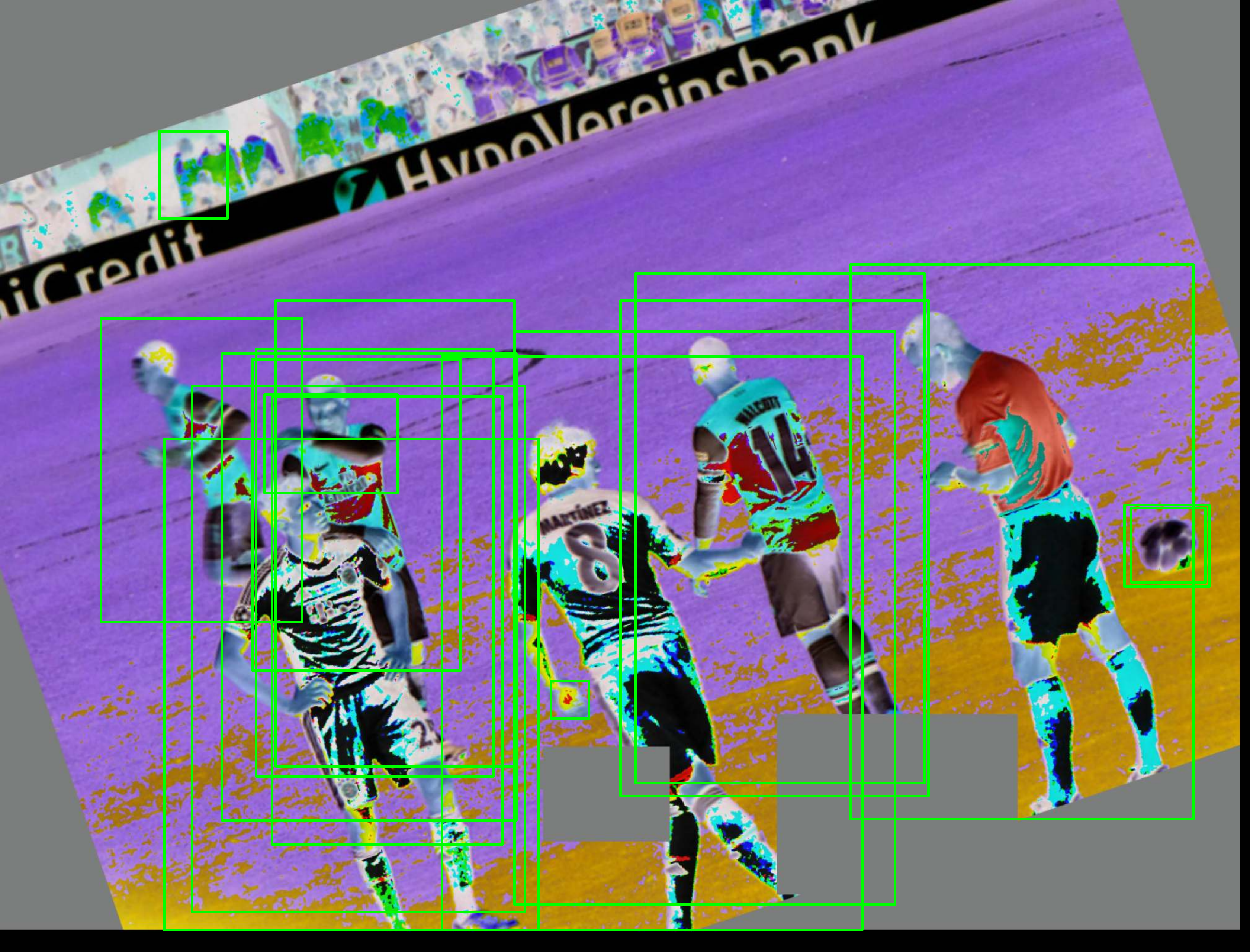}}&
    \makecell{\includegraphics[width=0.24\linewidth]{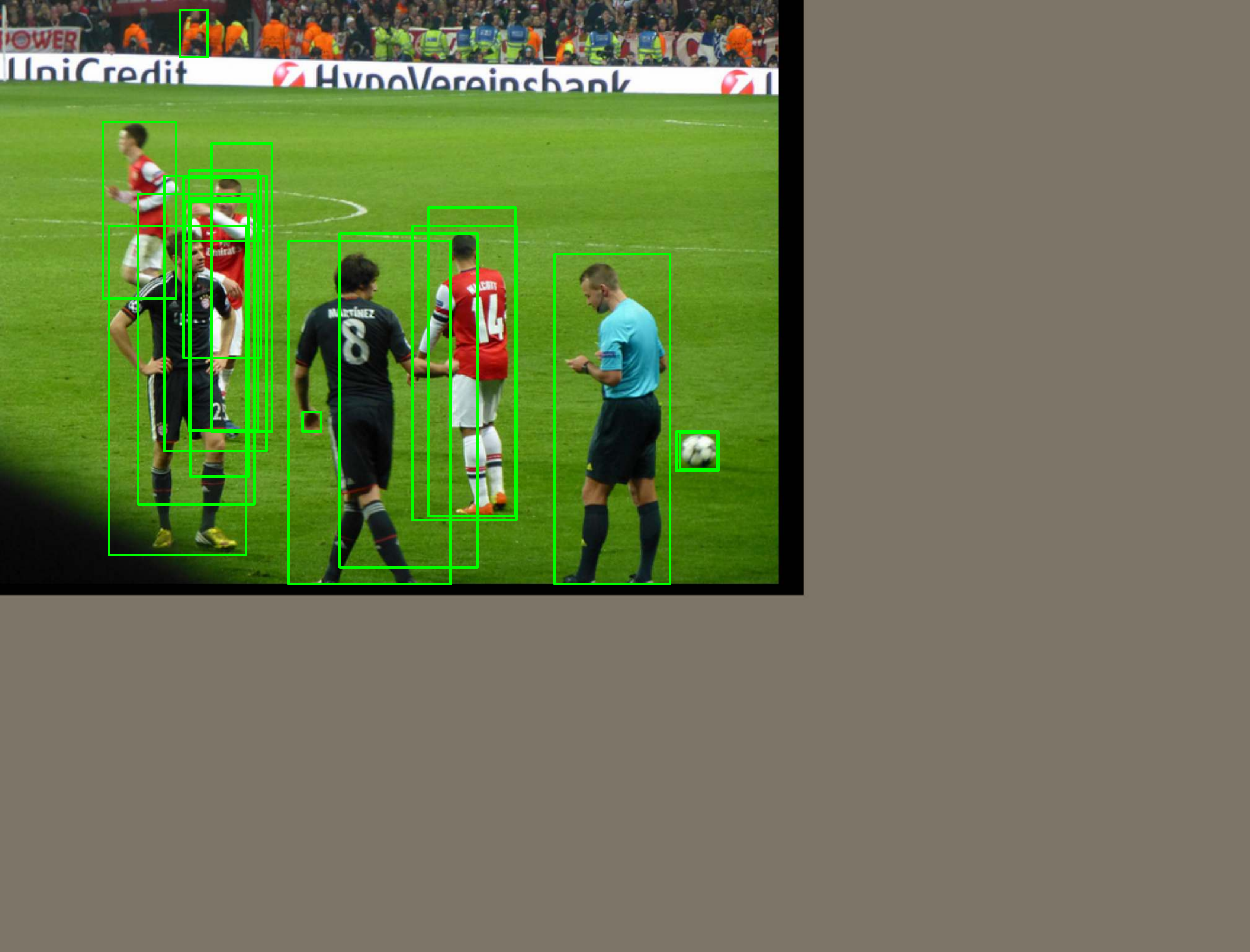}}&
    \makecell{\includegraphics[width=0.24\linewidth]{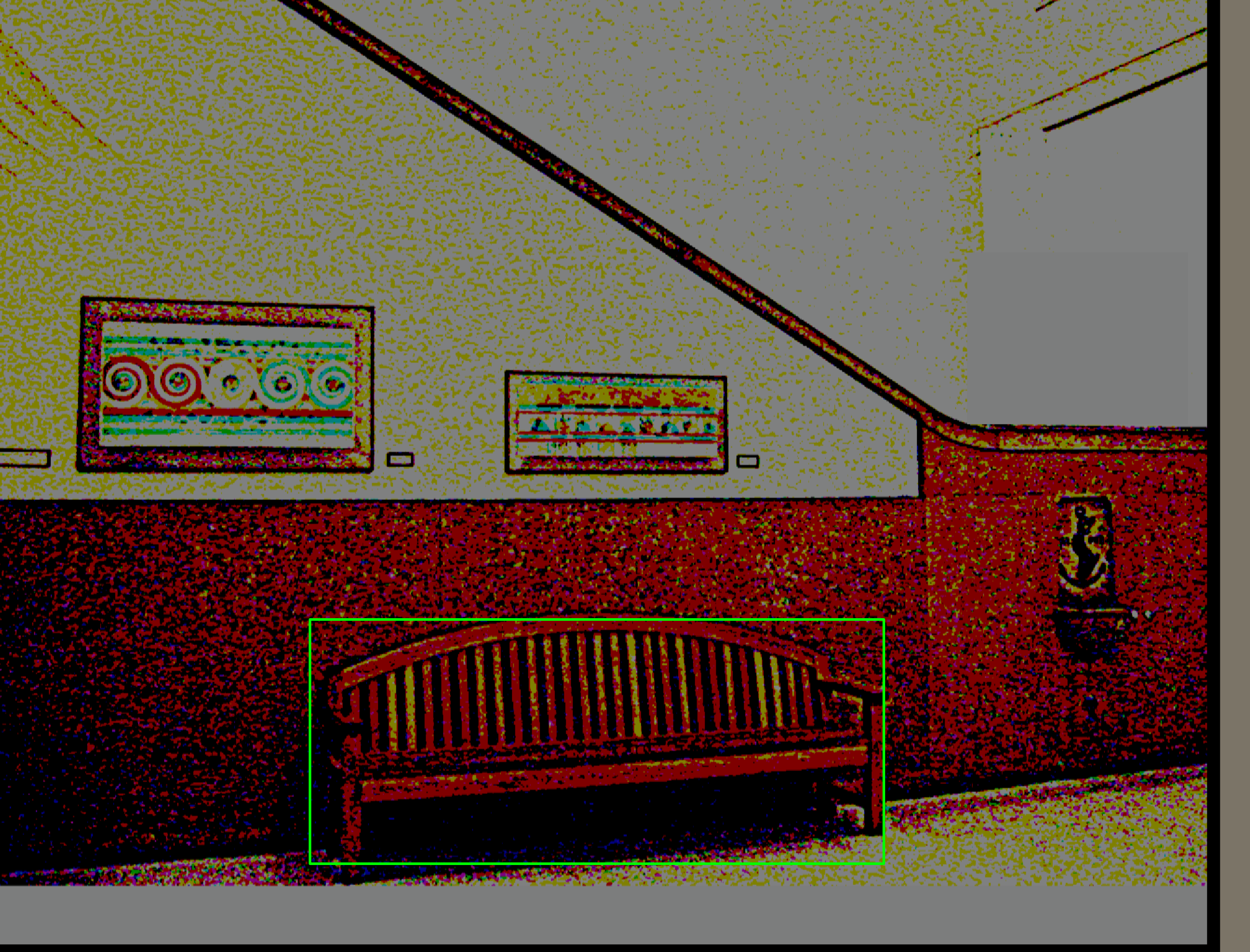}}&
    \makecell{\includegraphics[width=0.24\linewidth]{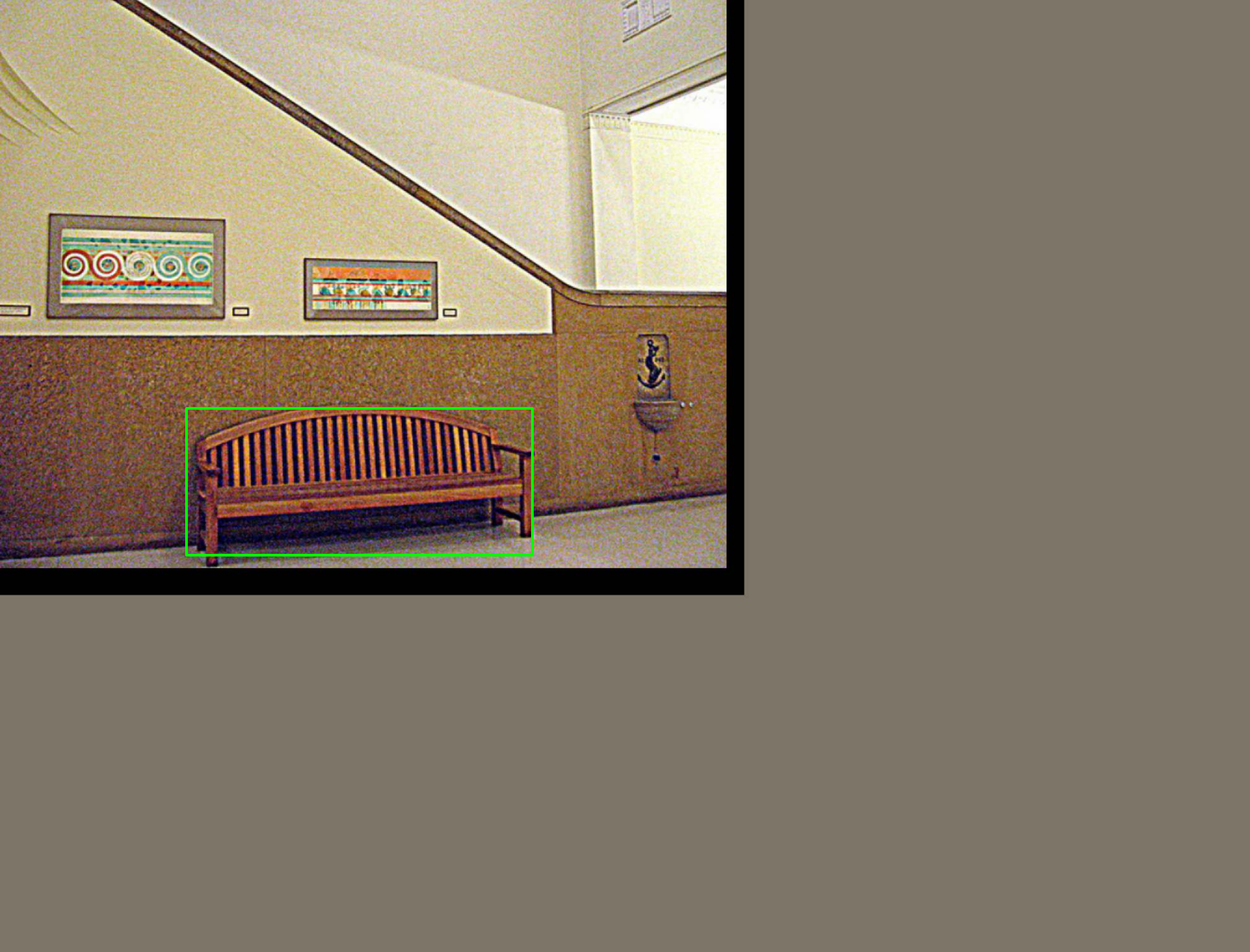}}\\
    \makecell{\includegraphics[width=0.24\linewidth]{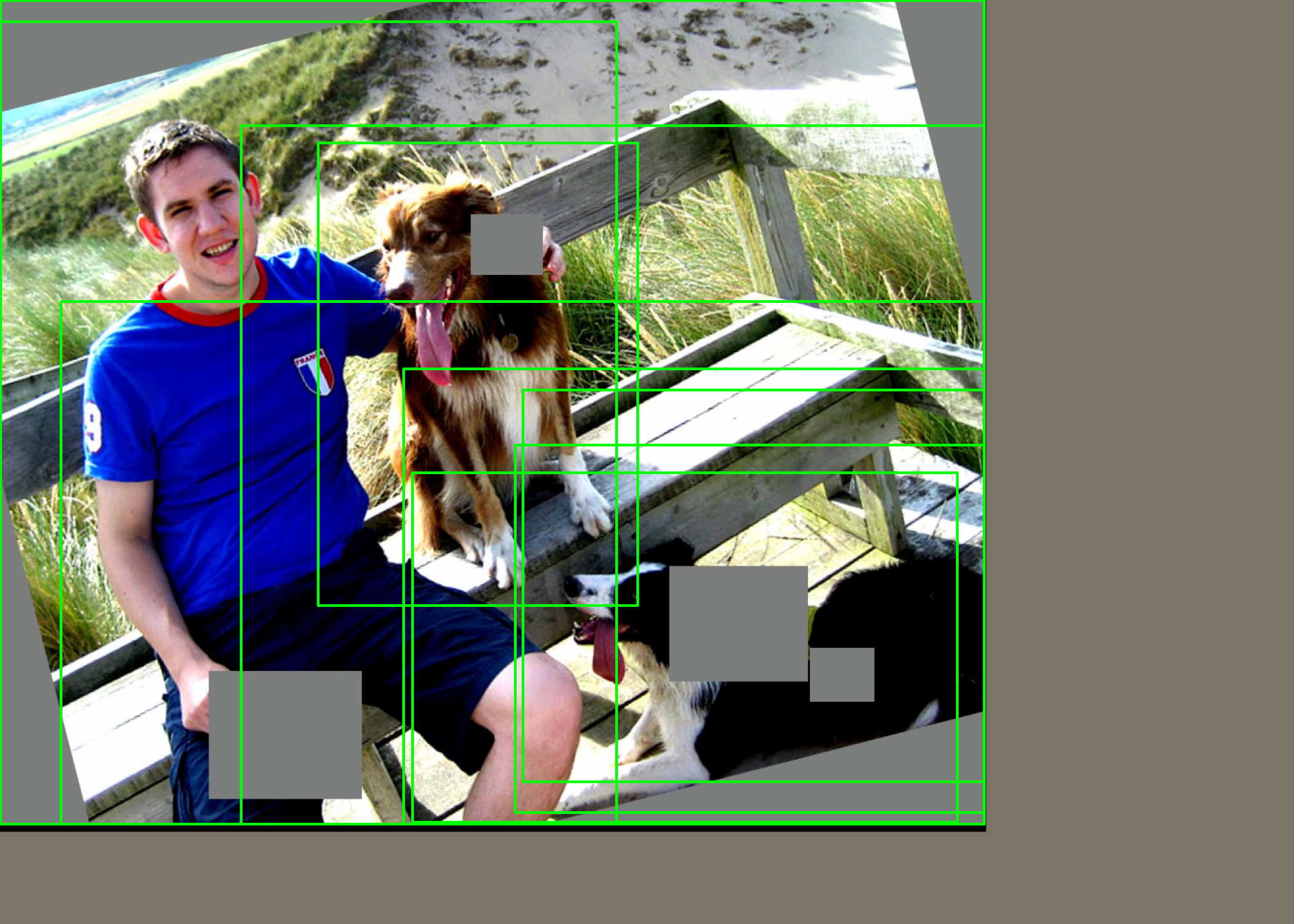}}&
    \makecell{\includegraphics[width=0.24\linewidth]{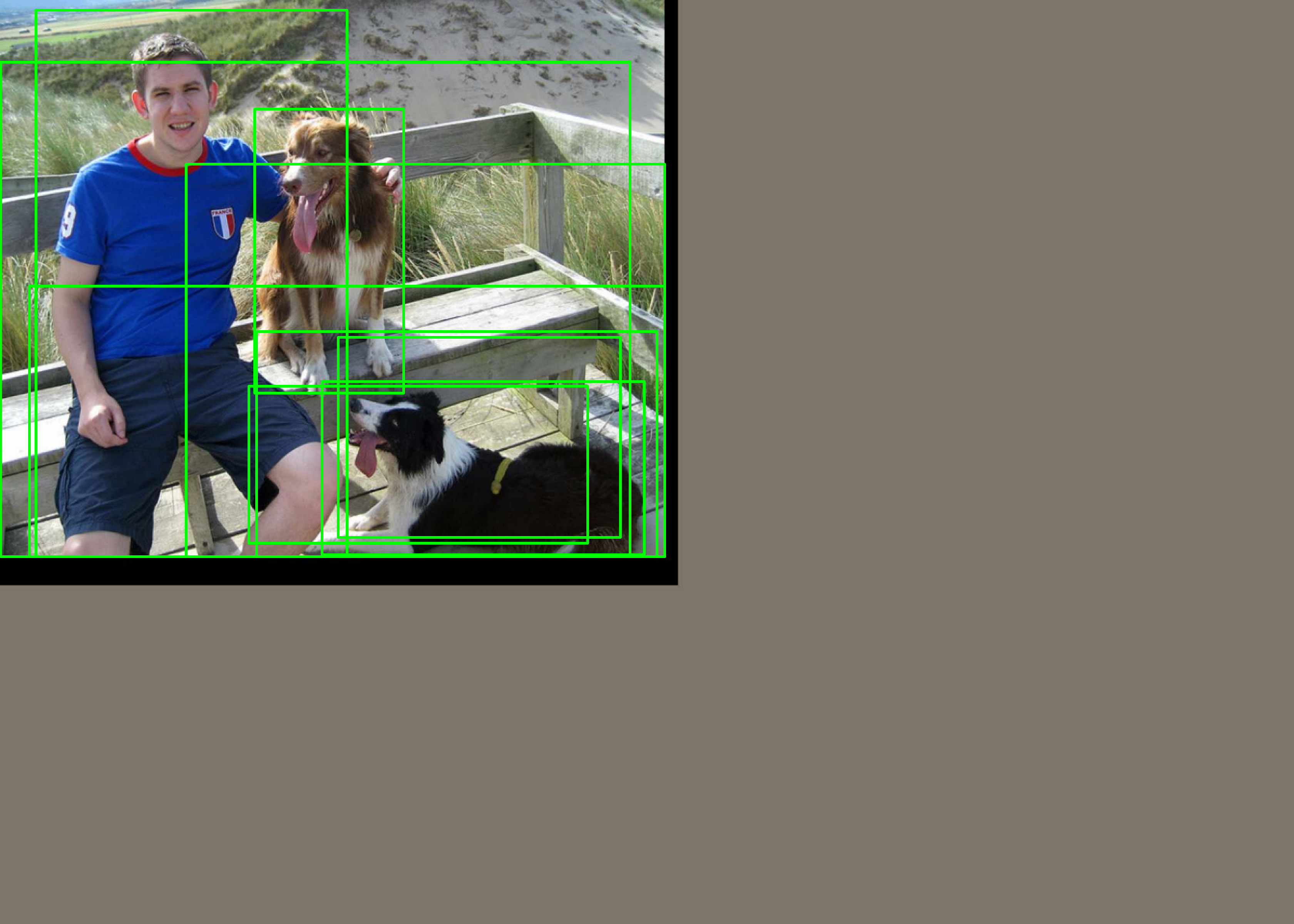}}&
    \makecell{\includegraphics[width=0.24\linewidth]{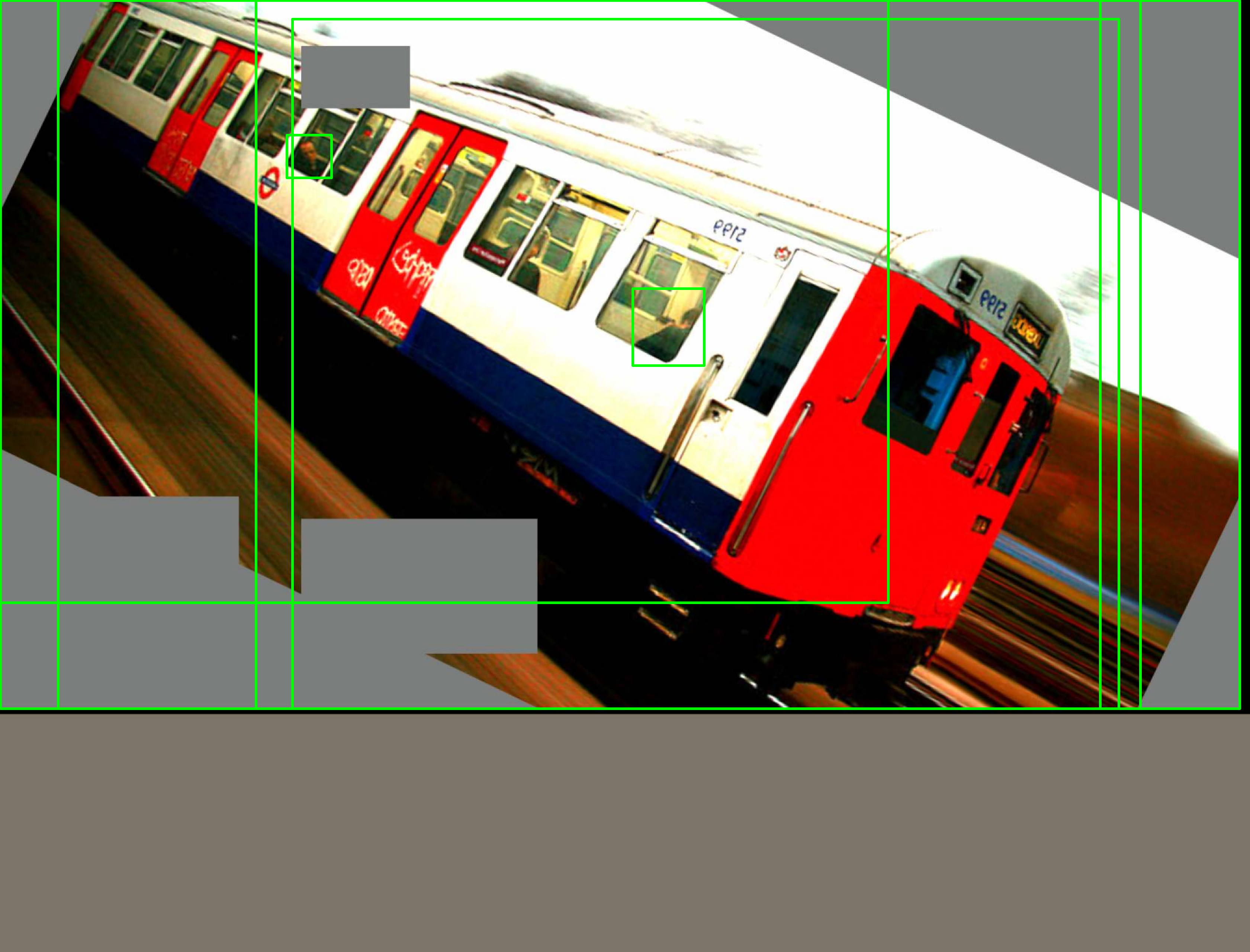}}&
    \makecell{\includegraphics[width=0.24\linewidth]{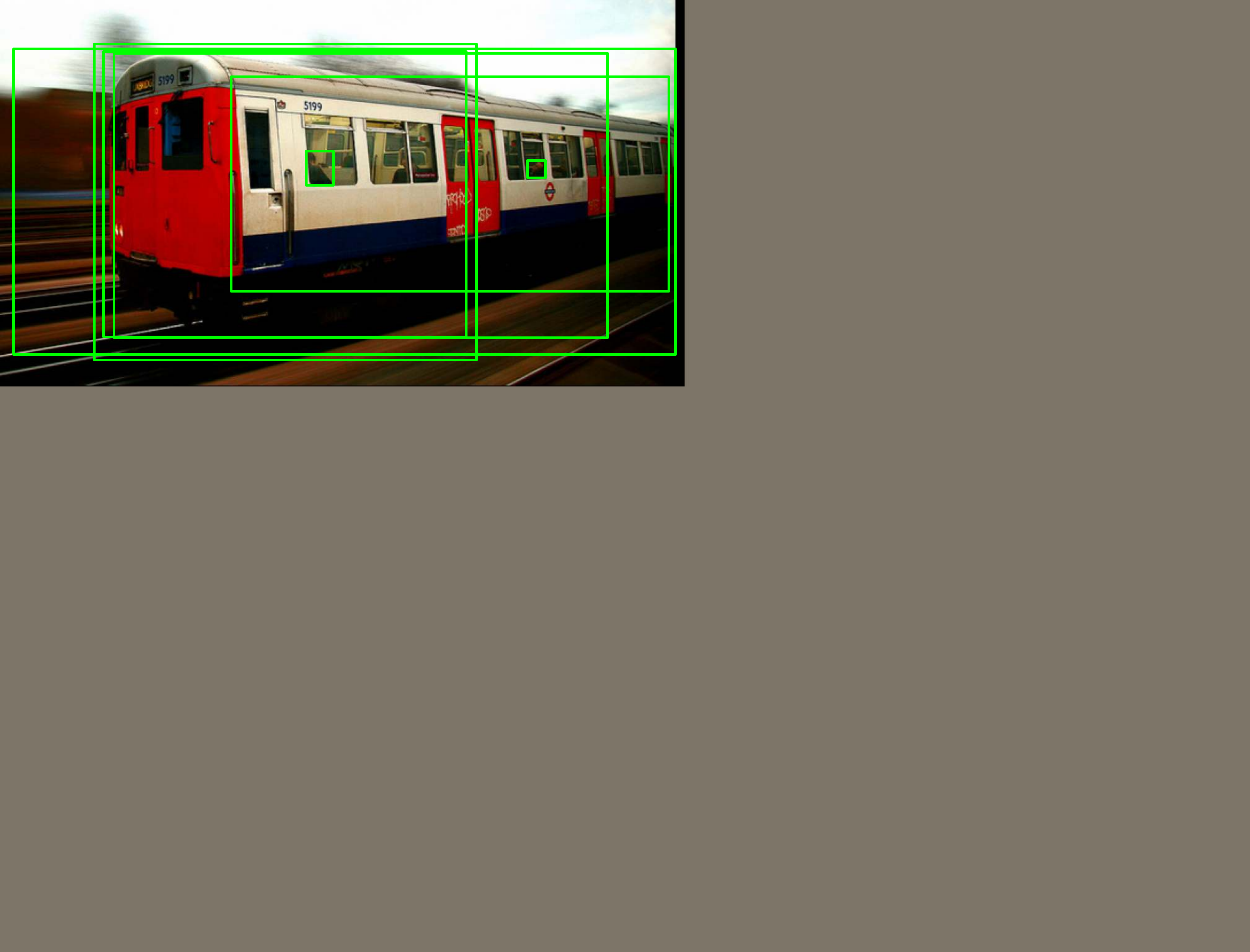}}\\
    \makecell{\includegraphics[width=0.24\linewidth]{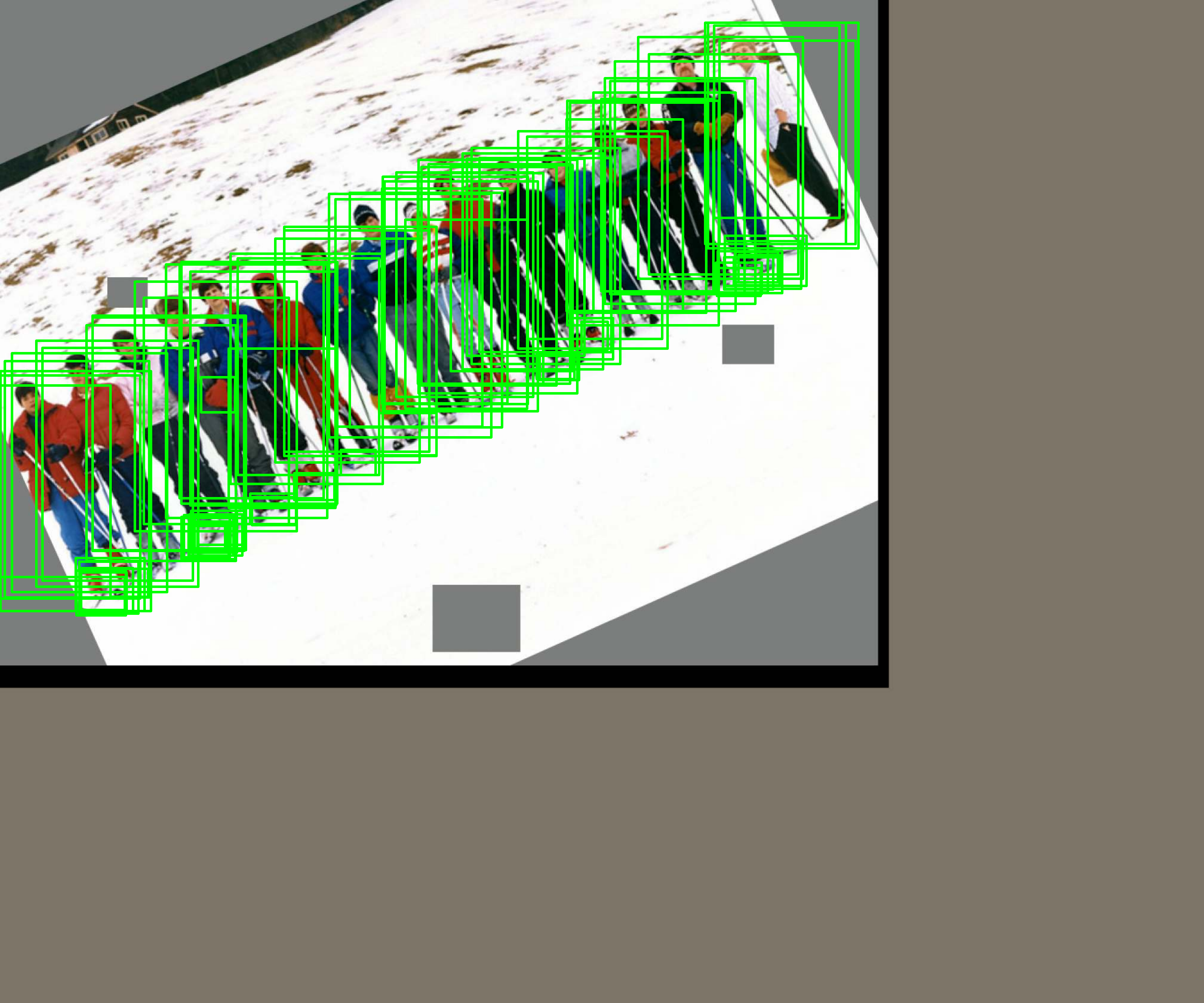}}&
    \makecell{\includegraphics[width=0.24\linewidth]{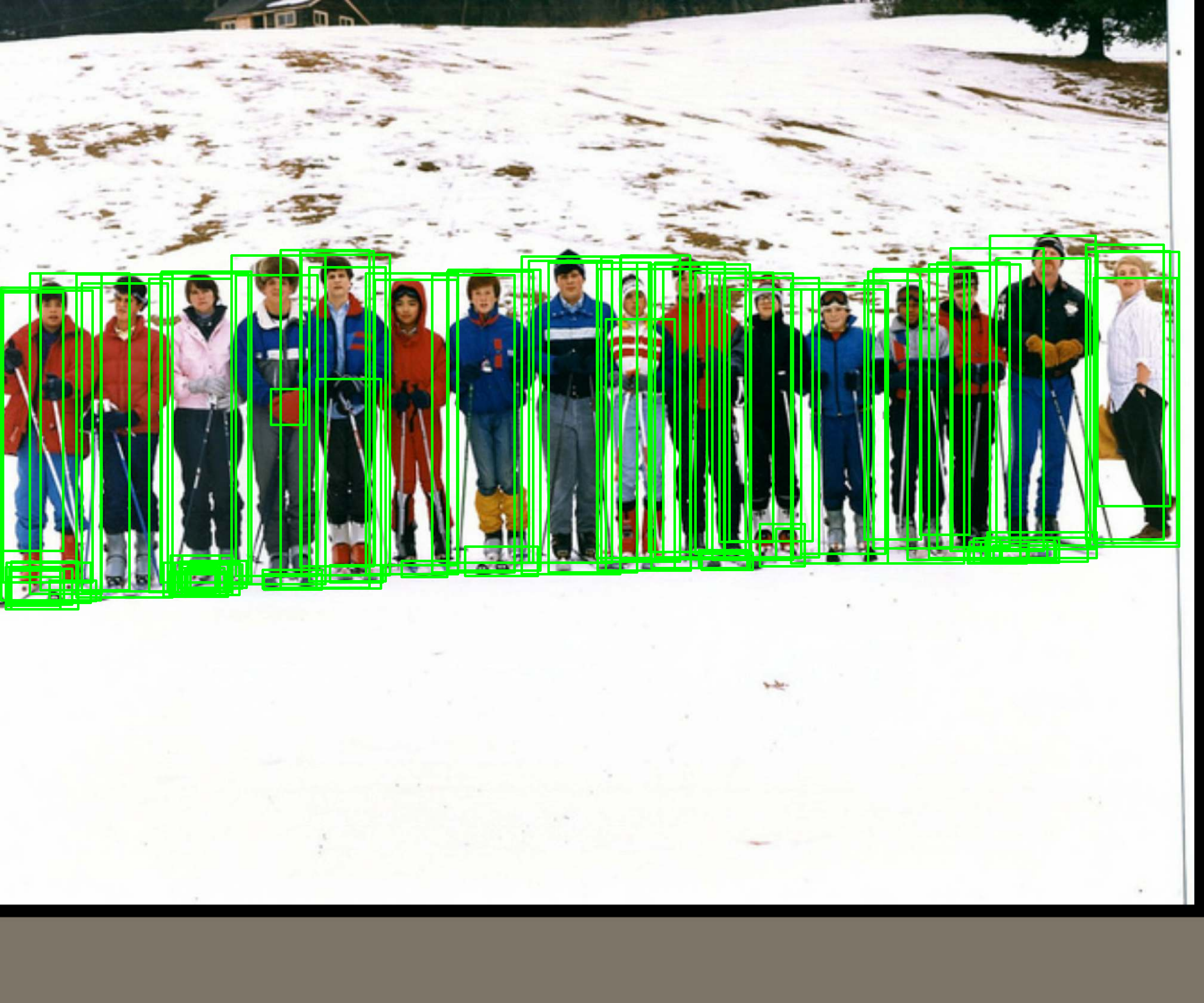}}&
    \makecell{\includegraphics[width=0.24\linewidth]{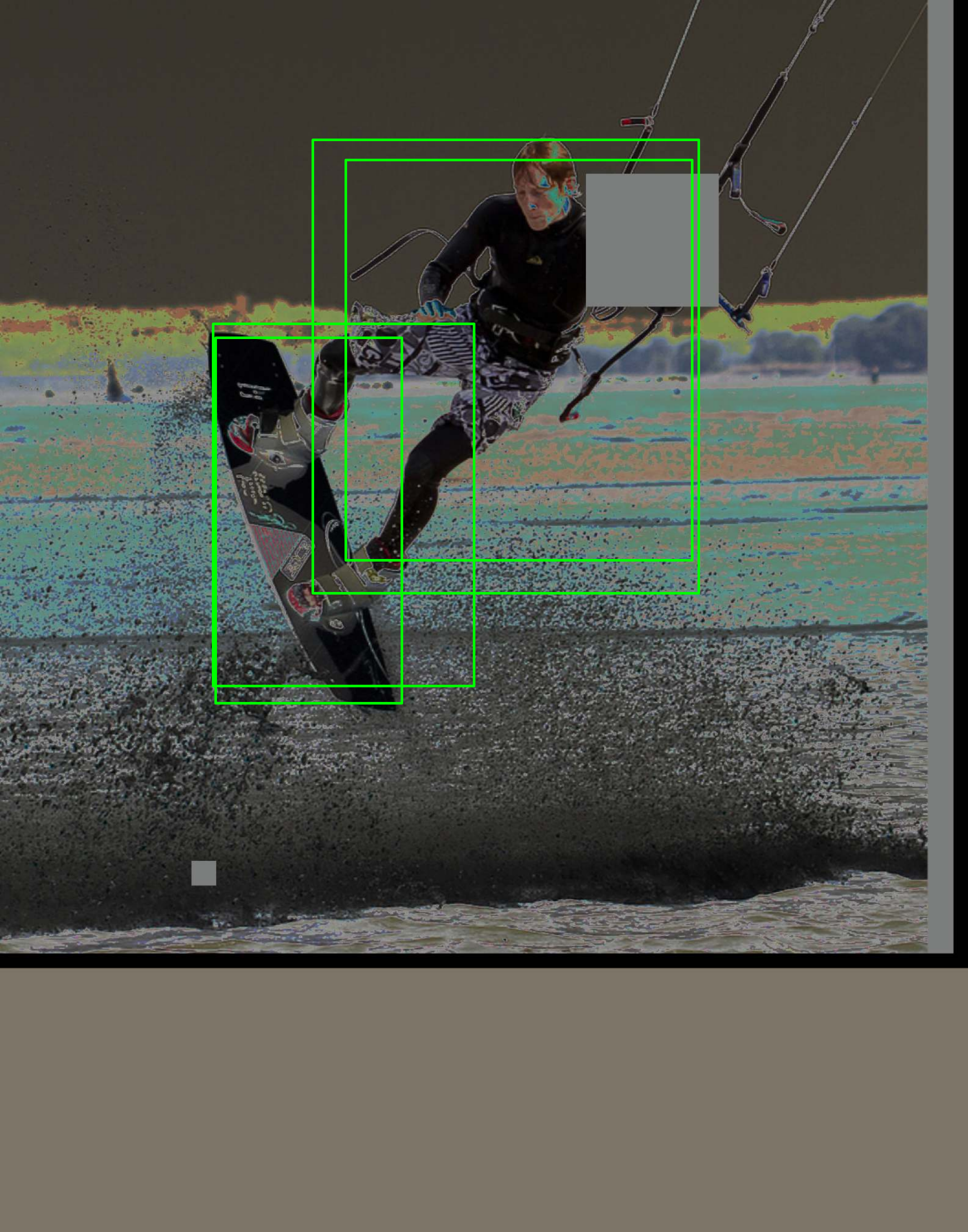}}&
    \makecell{\includegraphics[width=0.24\linewidth]{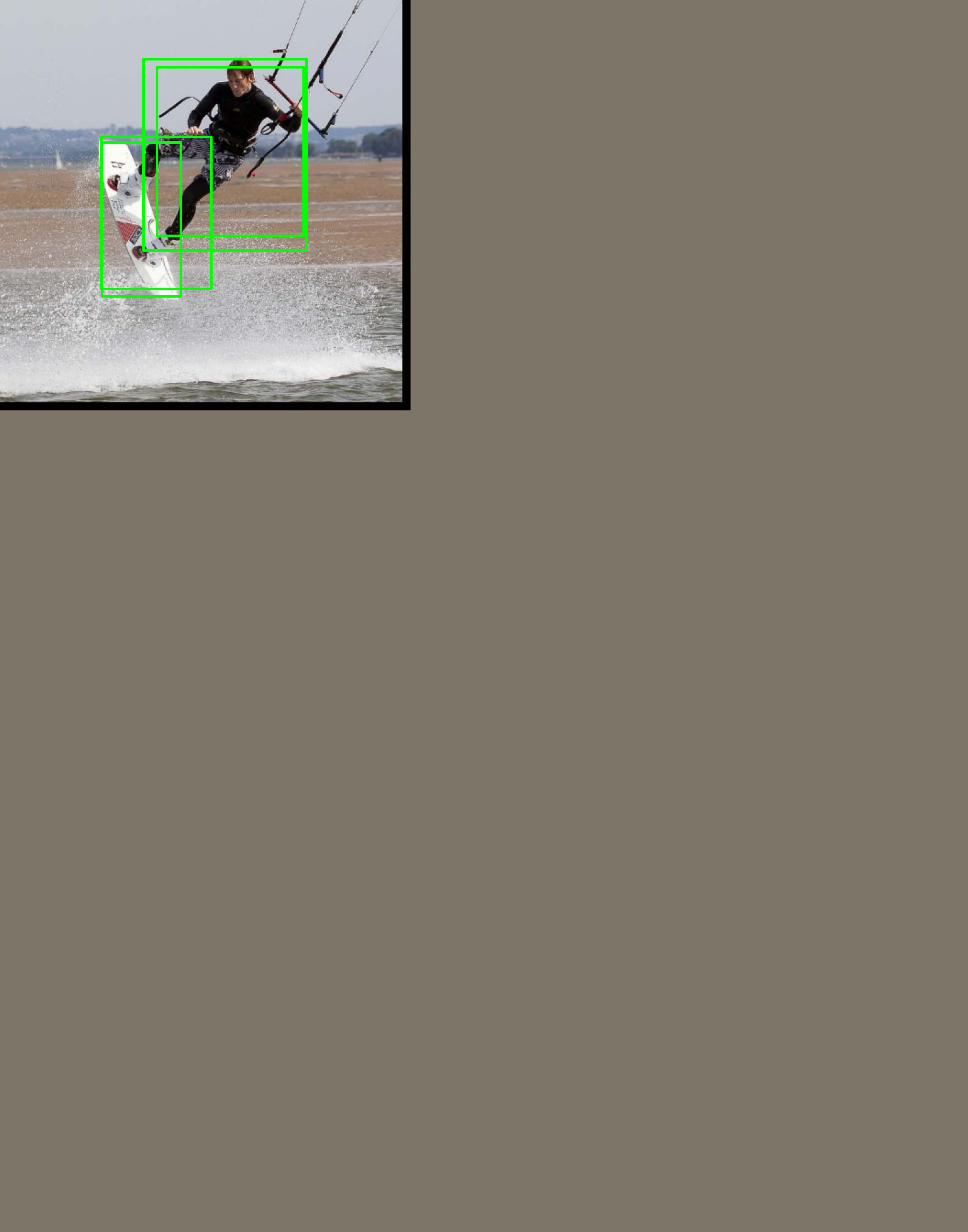}}
  \end{tabular}
  \caption{Visualizations of student-teacher proposals with confidence scores $\ge 0.99$. A pair of student-teacher proposals is aligned between student and teacher images for the purpose of enforcing classification similarity and localization consistency. The student images are subjected to a wide spectrum of complex scale, color, and geometric distortions, whereas the teacher images undergo simple random resizing and horizontal flipping transformations as the basis for generating reliable unsupervised pseudo targets to regularize the student’s learning trajectory. This multi-stream data augmentation strategy enables the student to tap into abundant region proposals to capture diverse feature representations that would otherwise be lost with aggressive confidence thresholding associated with pseudo-labeling. Best viewed digitally.}
  \label{fig::multi-views-supp}
\end{figure*}
\begin{figure*}[t]
  \centering
  \setlength\tabcolsep{2.0pt}
  \begin{tabular}{ccc}
    \small Supervised & \small Soft Teacher & \small \model (Ours)\\
    \rotatebox[origin=c]{90}{\small 1\% of Labels} \makecell{\includegraphics[width=0.255\textwidth]{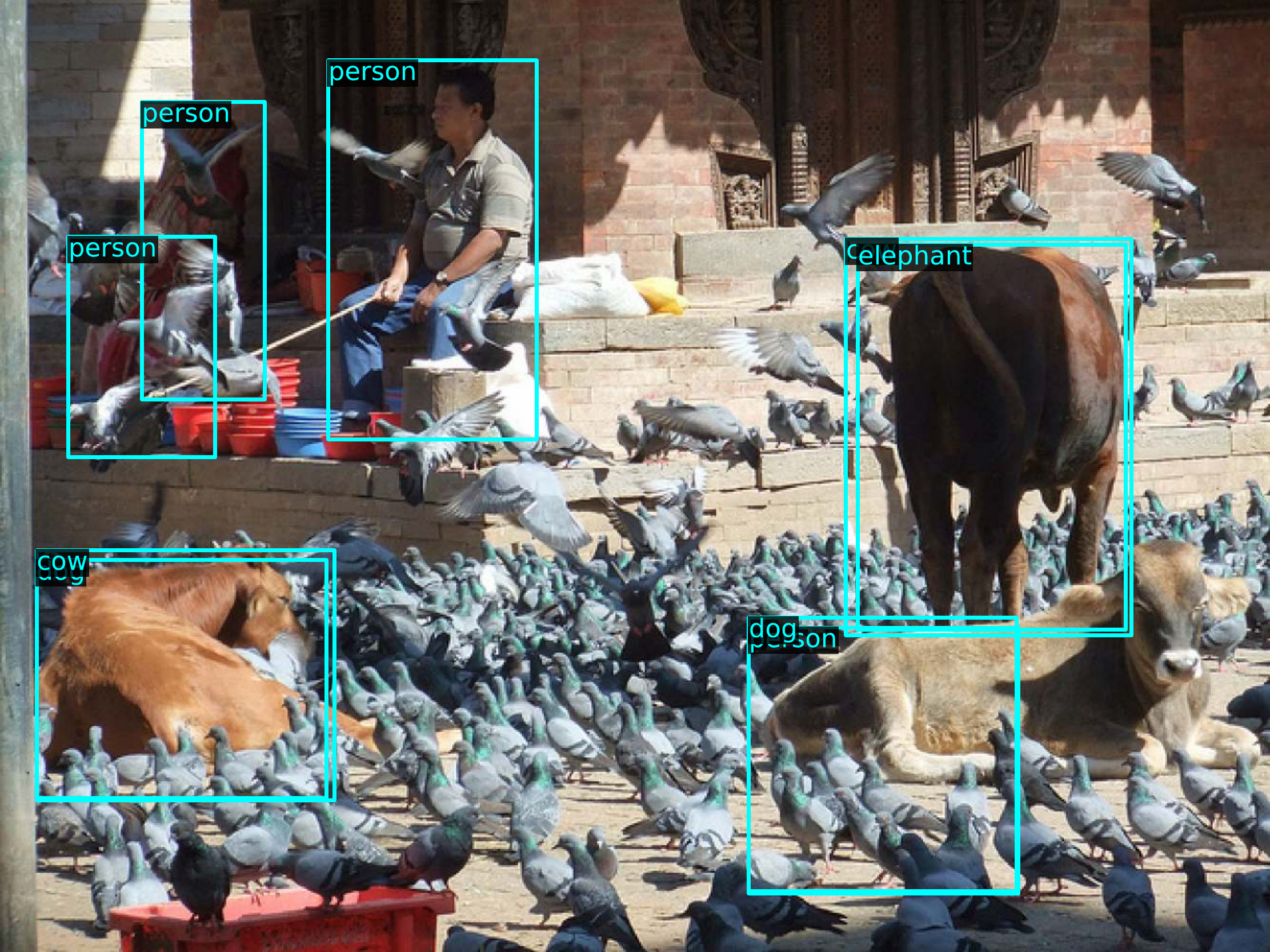}}&
    \makecell{\includegraphics[width=0.255\textwidth]{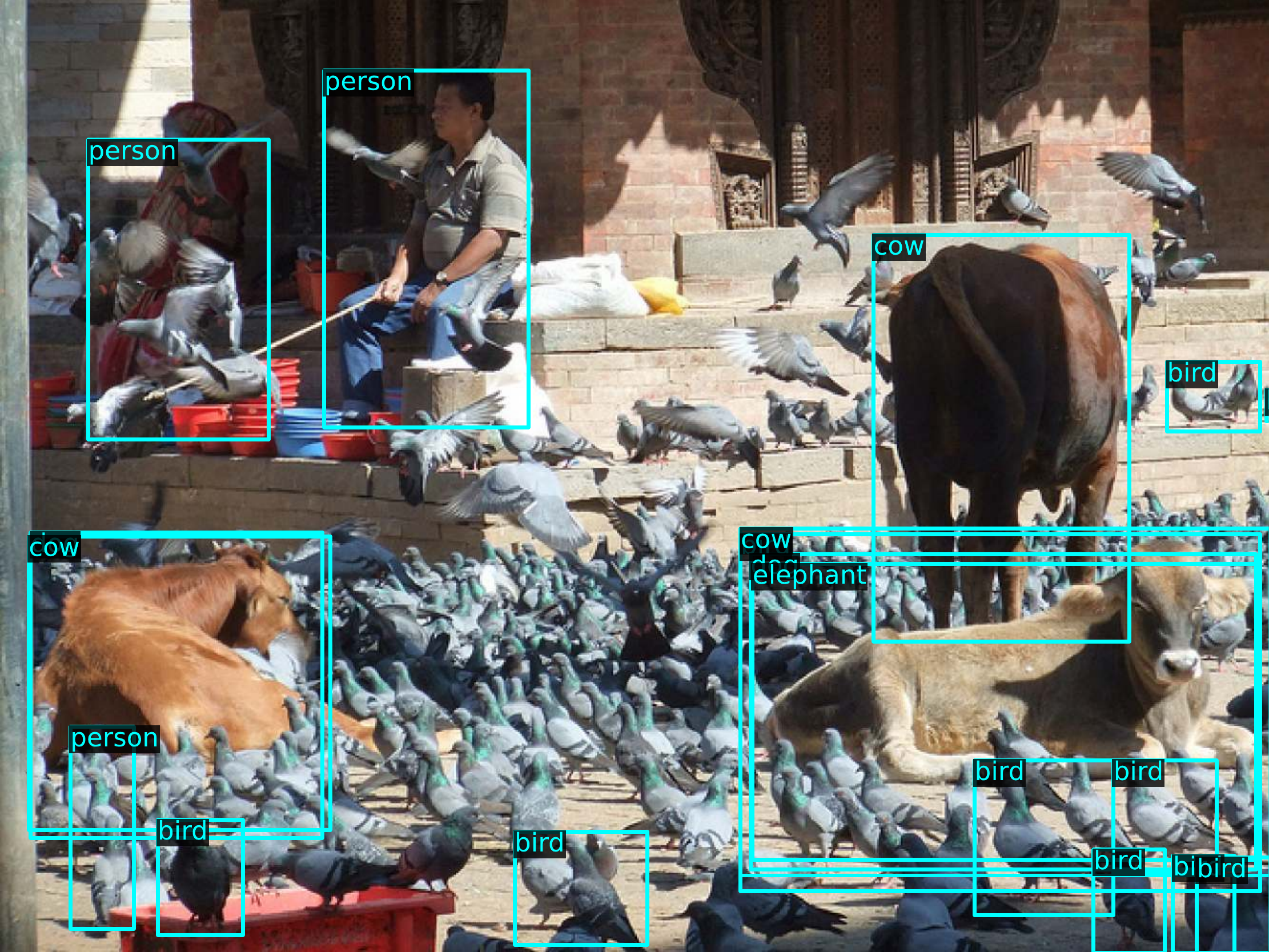}}&
    \makecell{\includegraphics[width=0.255\textwidth]{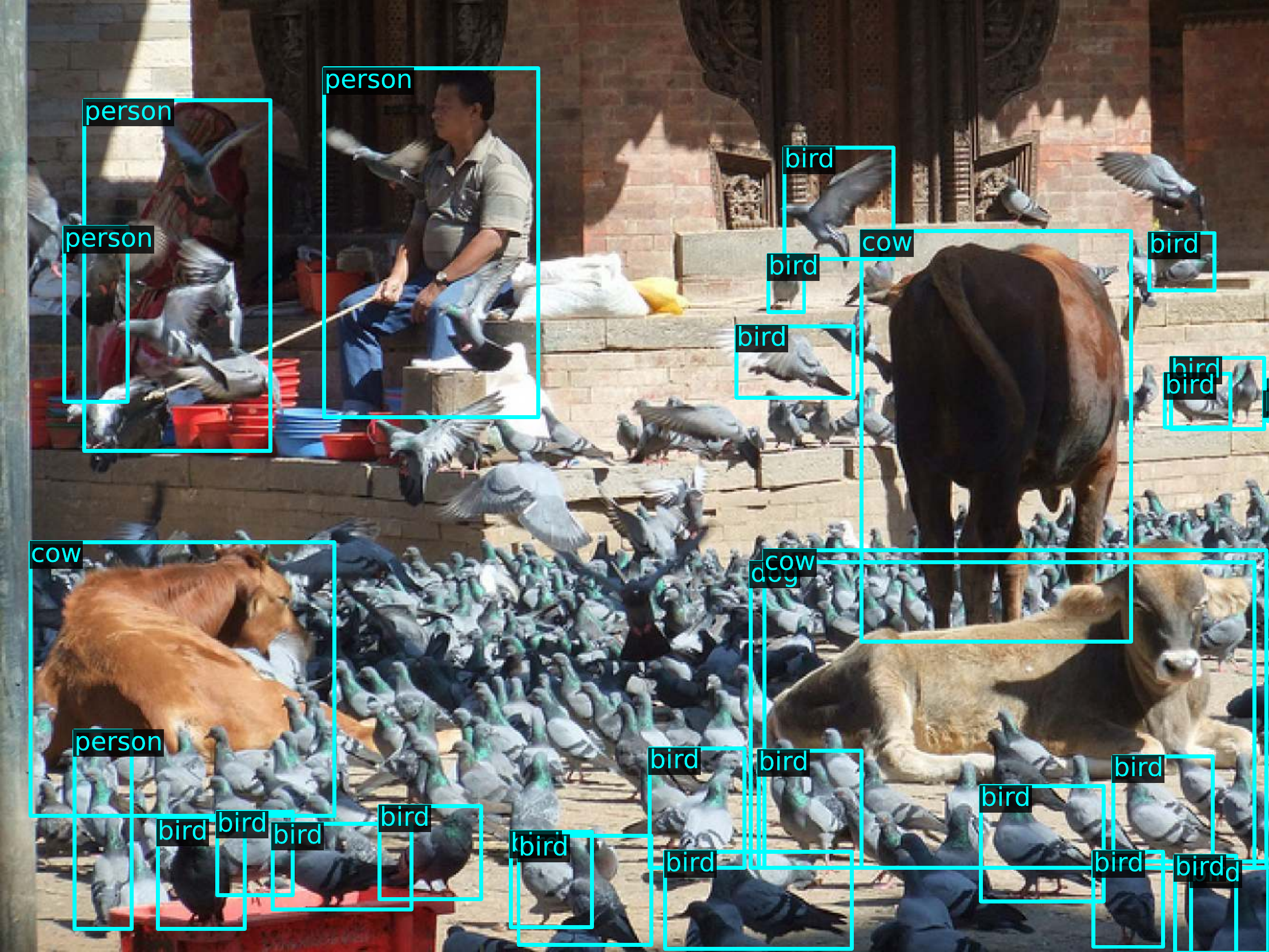}}\\
    \rotatebox[origin=c]{90}{\small 1\% of Labels} \makecell{\includegraphics[width=0.255\textwidth]{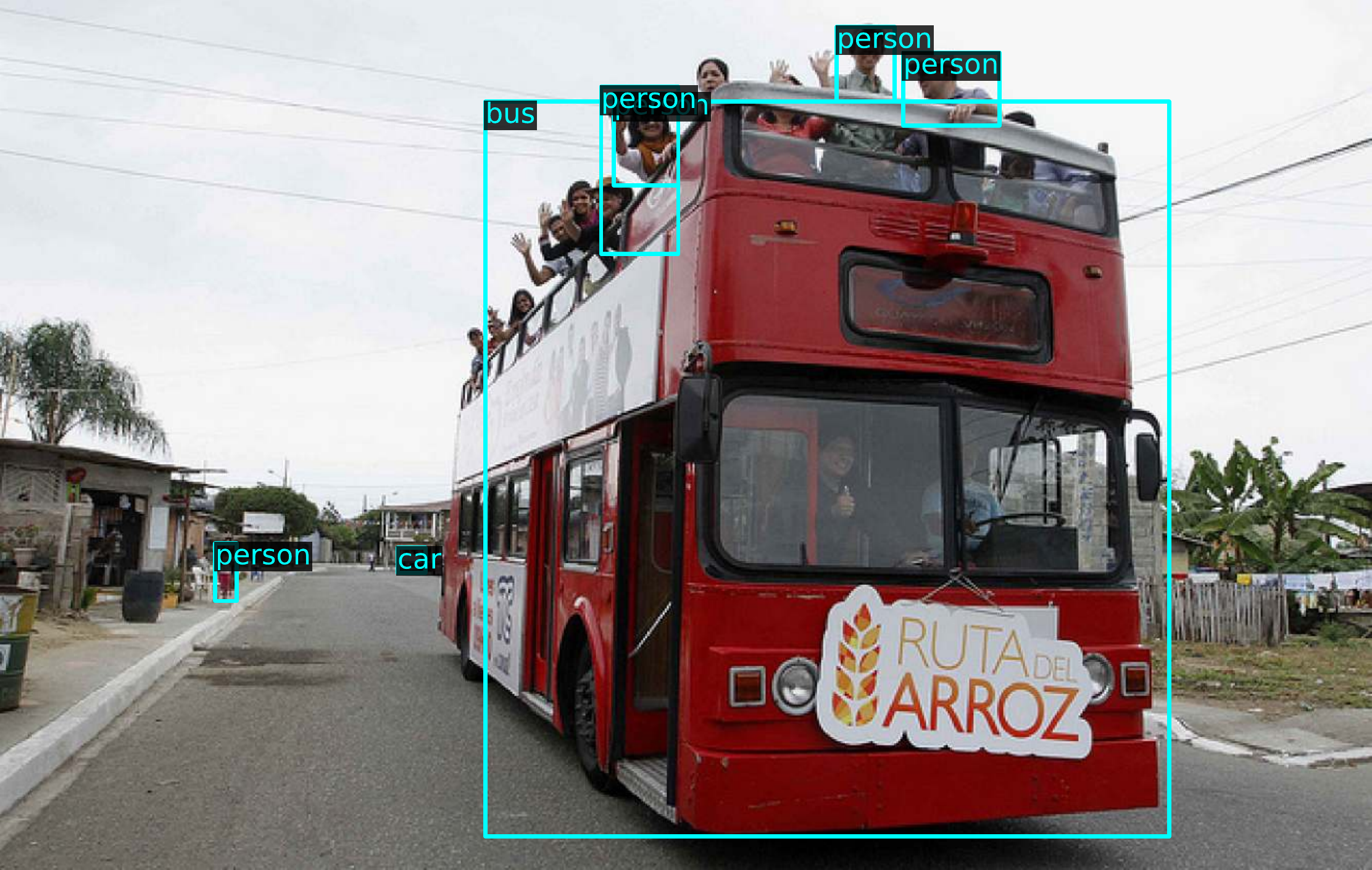}}&
    \makecell{\includegraphics[width=0.255\textwidth]{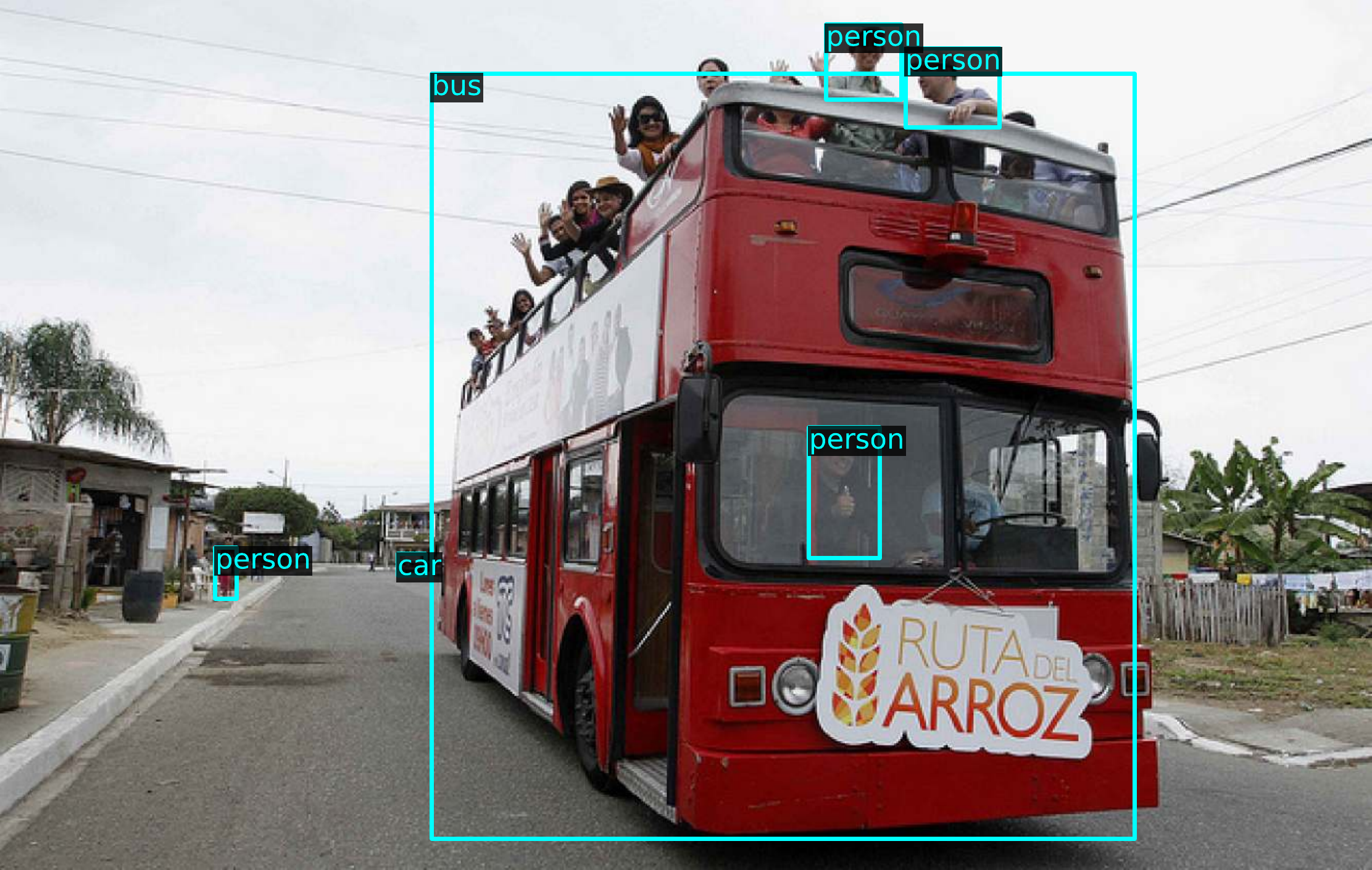}}&
    \makecell{\includegraphics[width=0.255\textwidth]{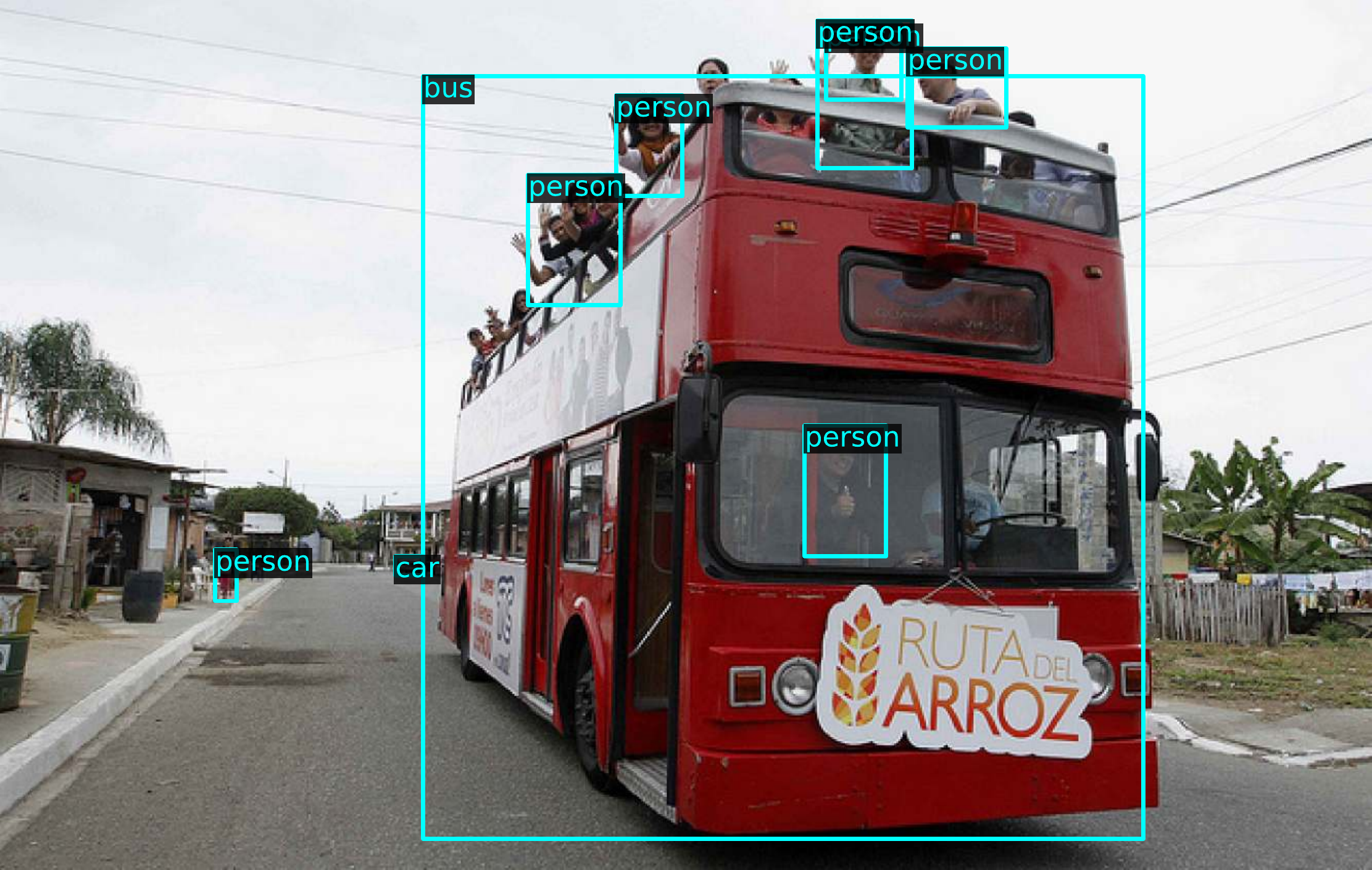}}\\
    \rotatebox[origin=c]{90}{\small 5\% of Labels} \makecell{\includegraphics[width=0.255\textwidth]{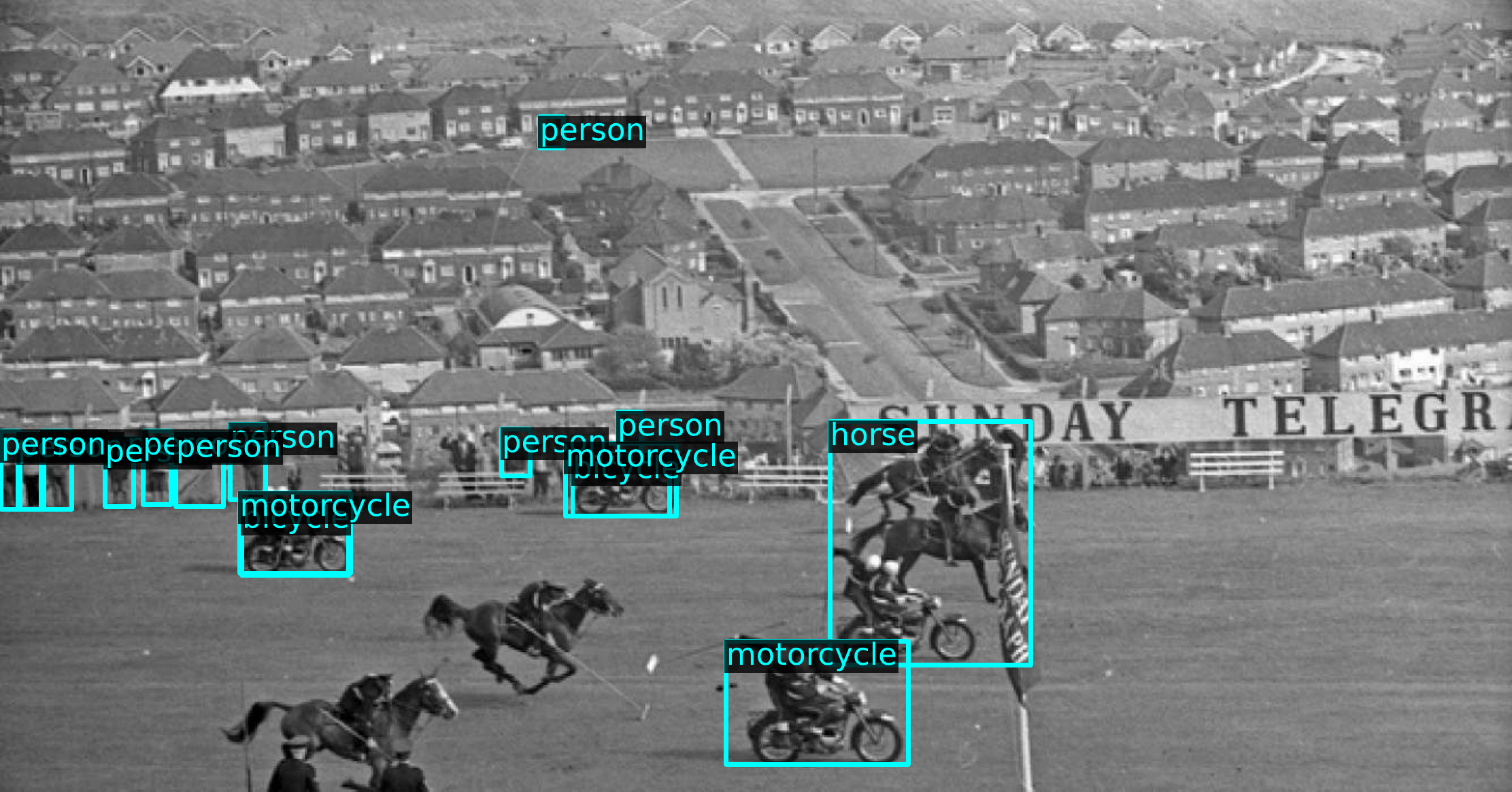}}&
    \makecell{\includegraphics[width=0.255\textwidth]{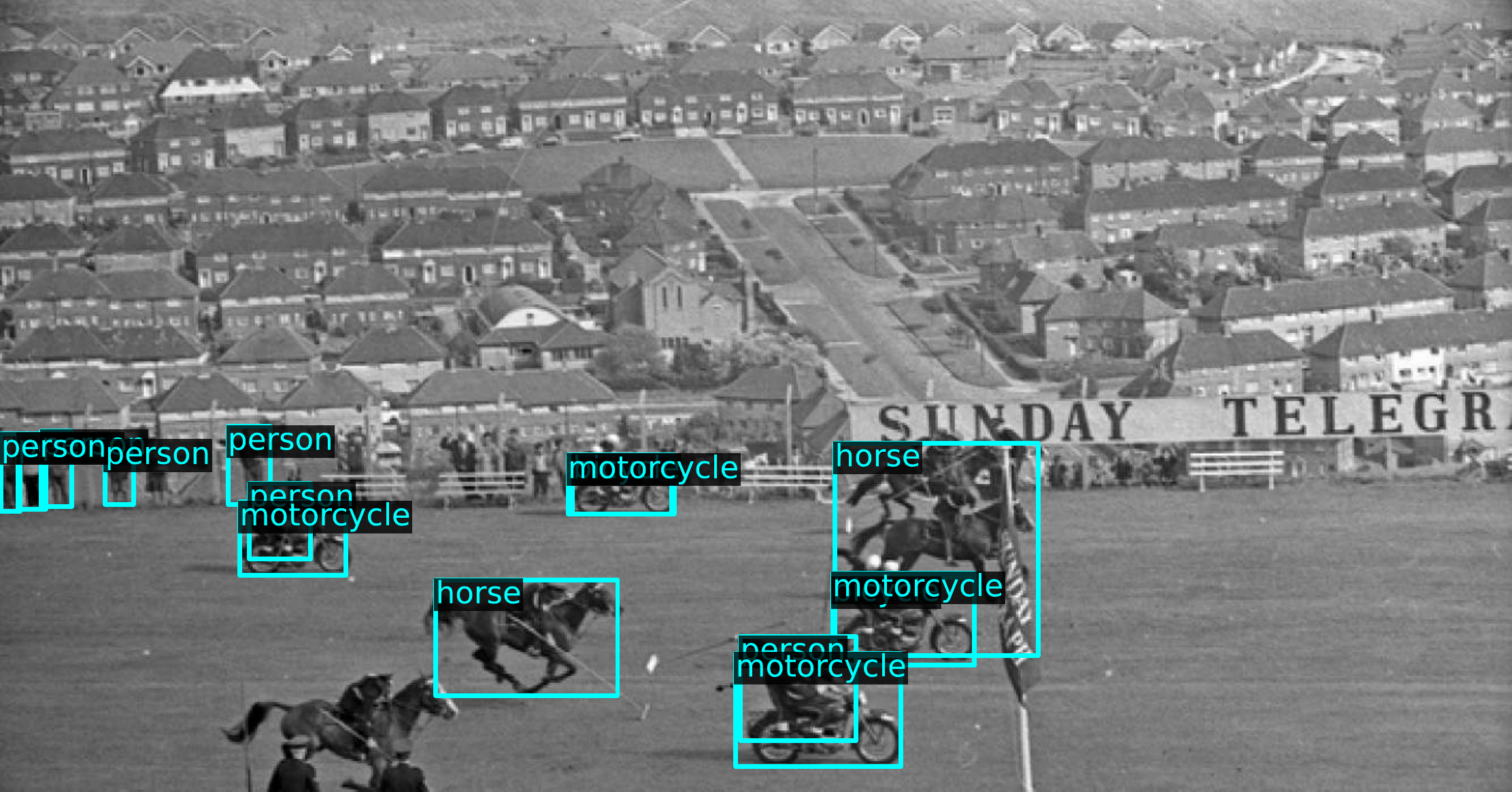}}&
    \makecell{\includegraphics[width=0.255\textwidth]{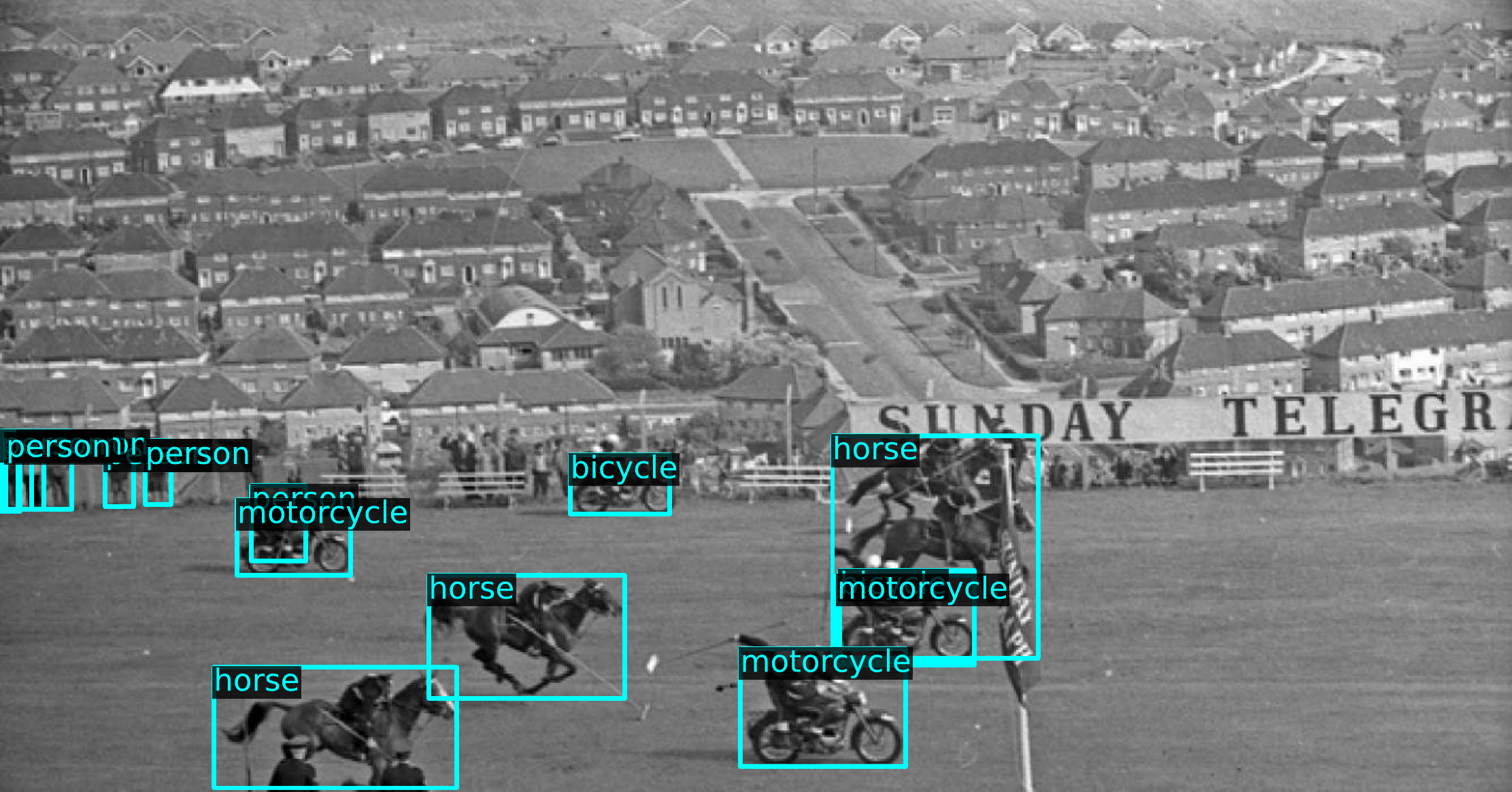}}\\
    \rotatebox[origin=c]{90}{\small 5\% of Labels} \makecell{\includegraphics[width=0.255\textwidth]{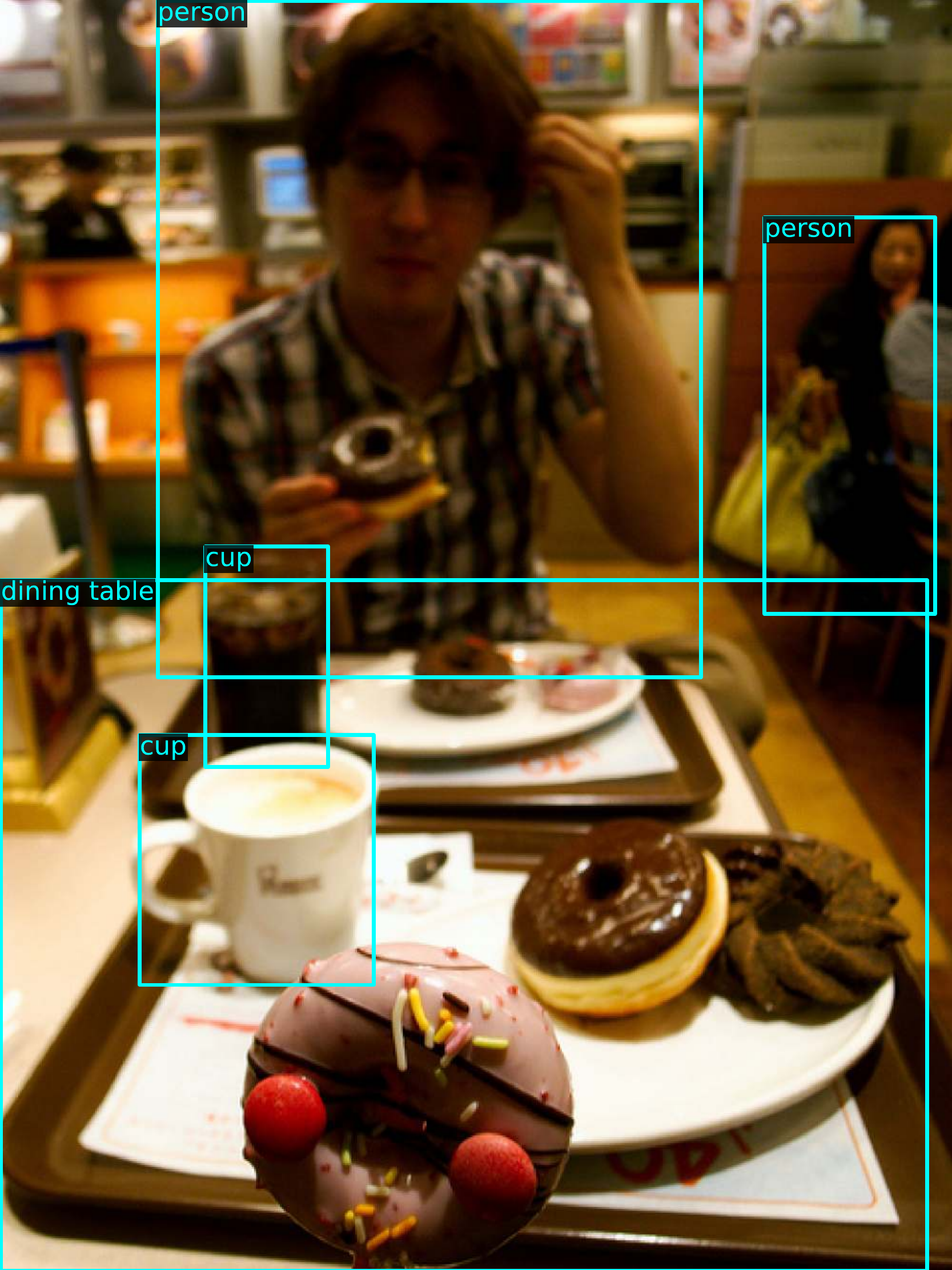}}&
    \makecell{\includegraphics[width=0.255\textwidth]{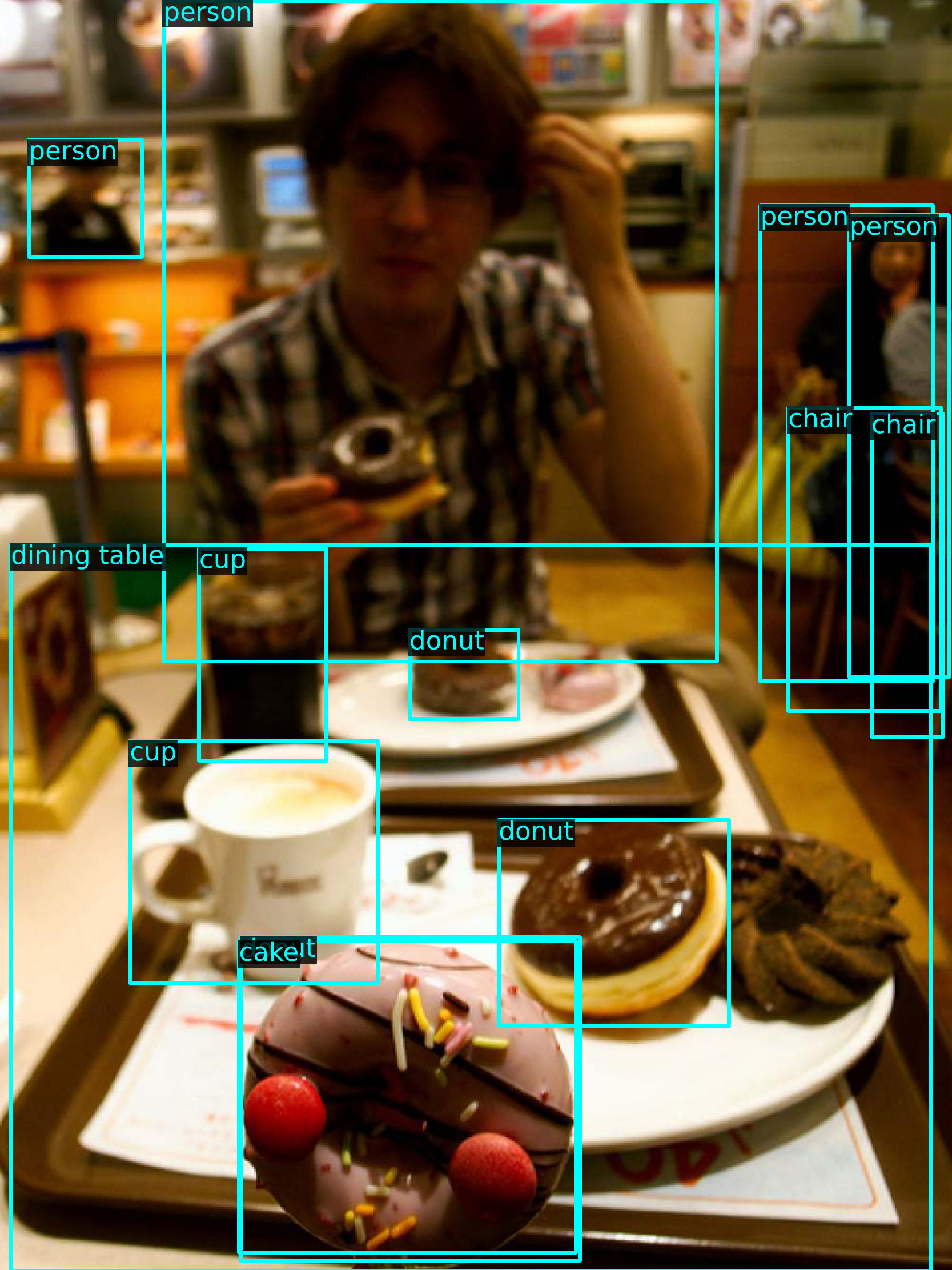}}&
    \makecell{\includegraphics[width=0.255\textwidth]{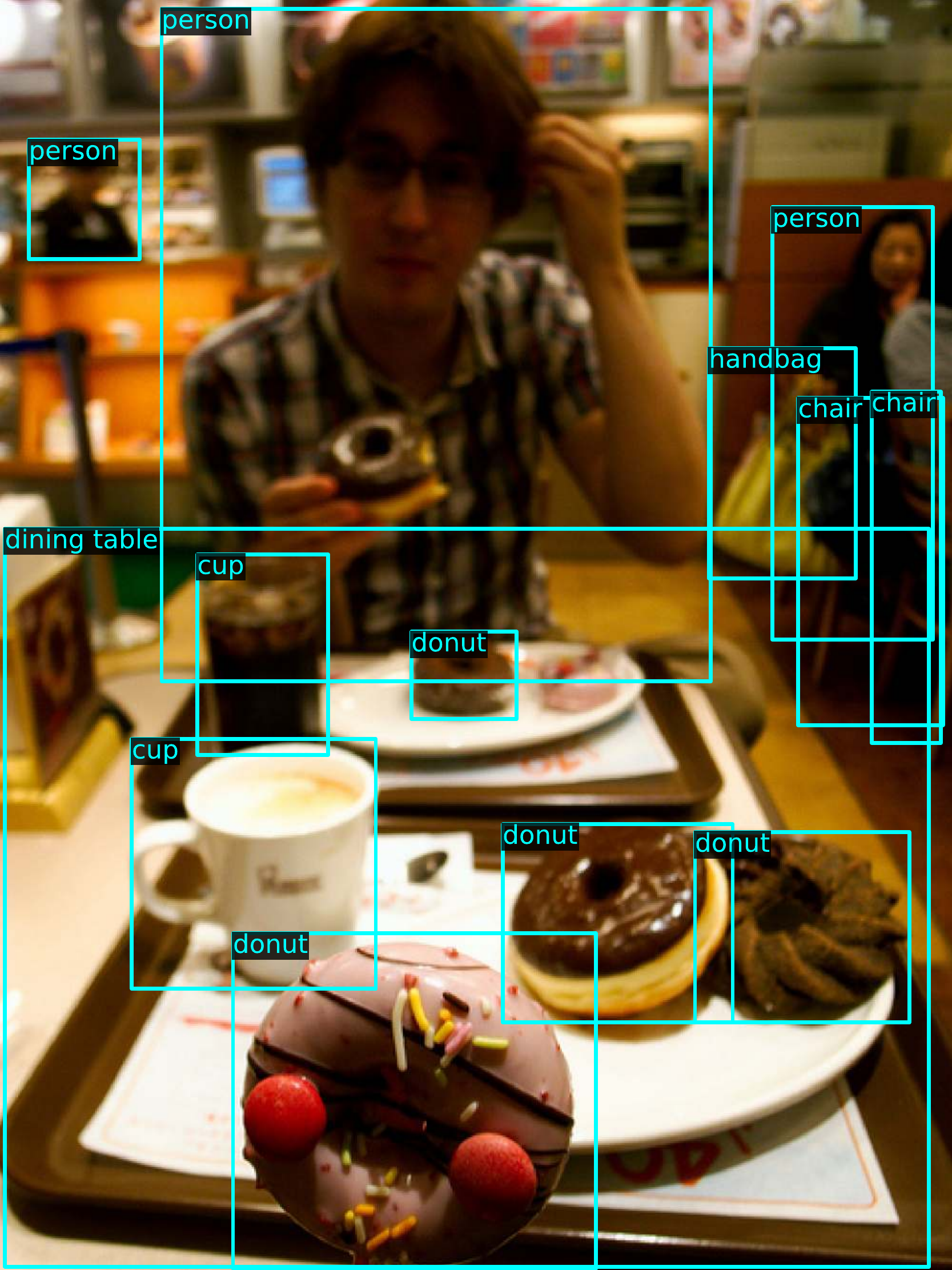}}\\
    \rotatebox[origin=c]{90}{\small 10\% of Labels} \makecell{\includegraphics[width=0.255\textwidth]{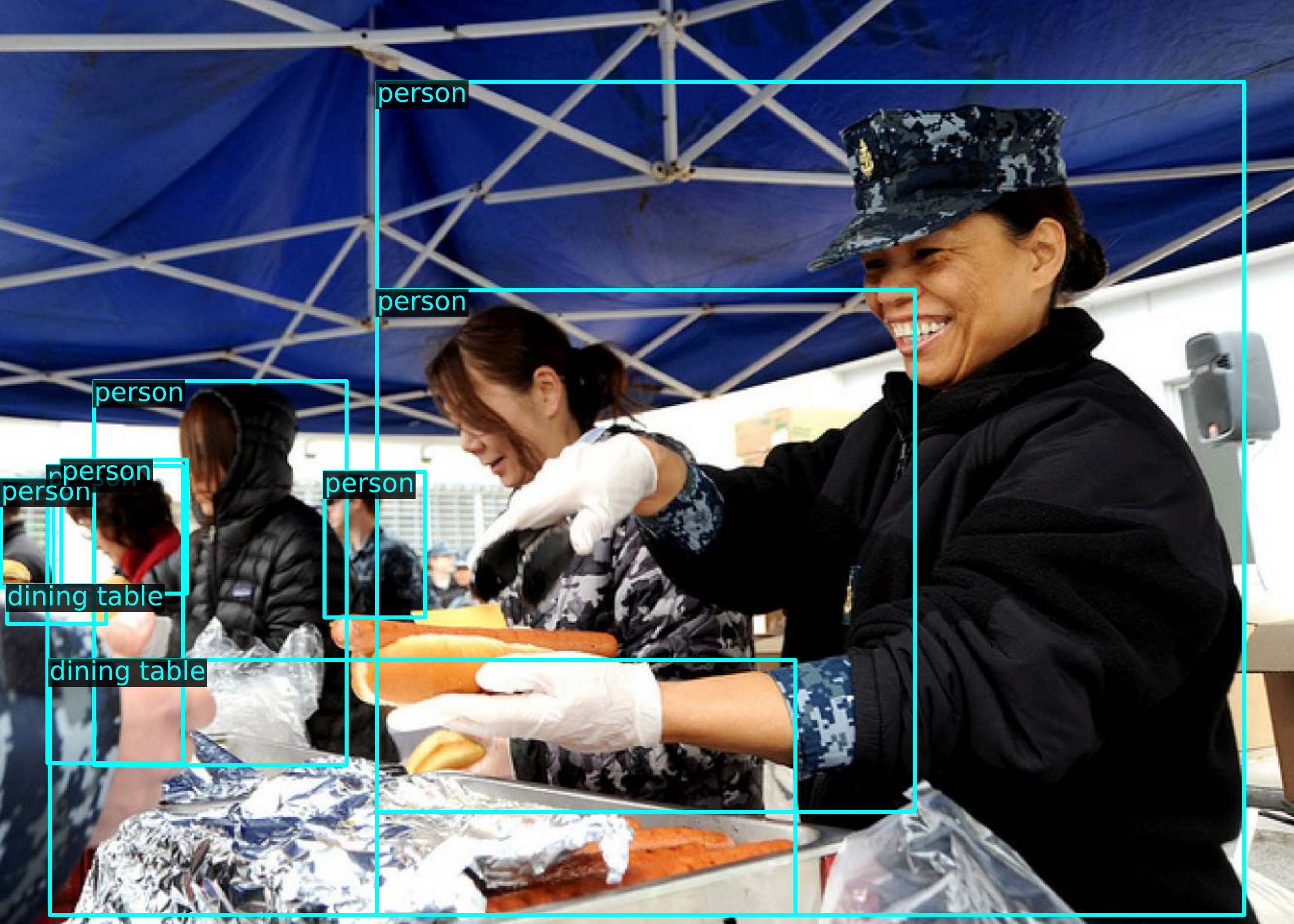}}&
    \makecell{\includegraphics[width=0.255\textwidth]{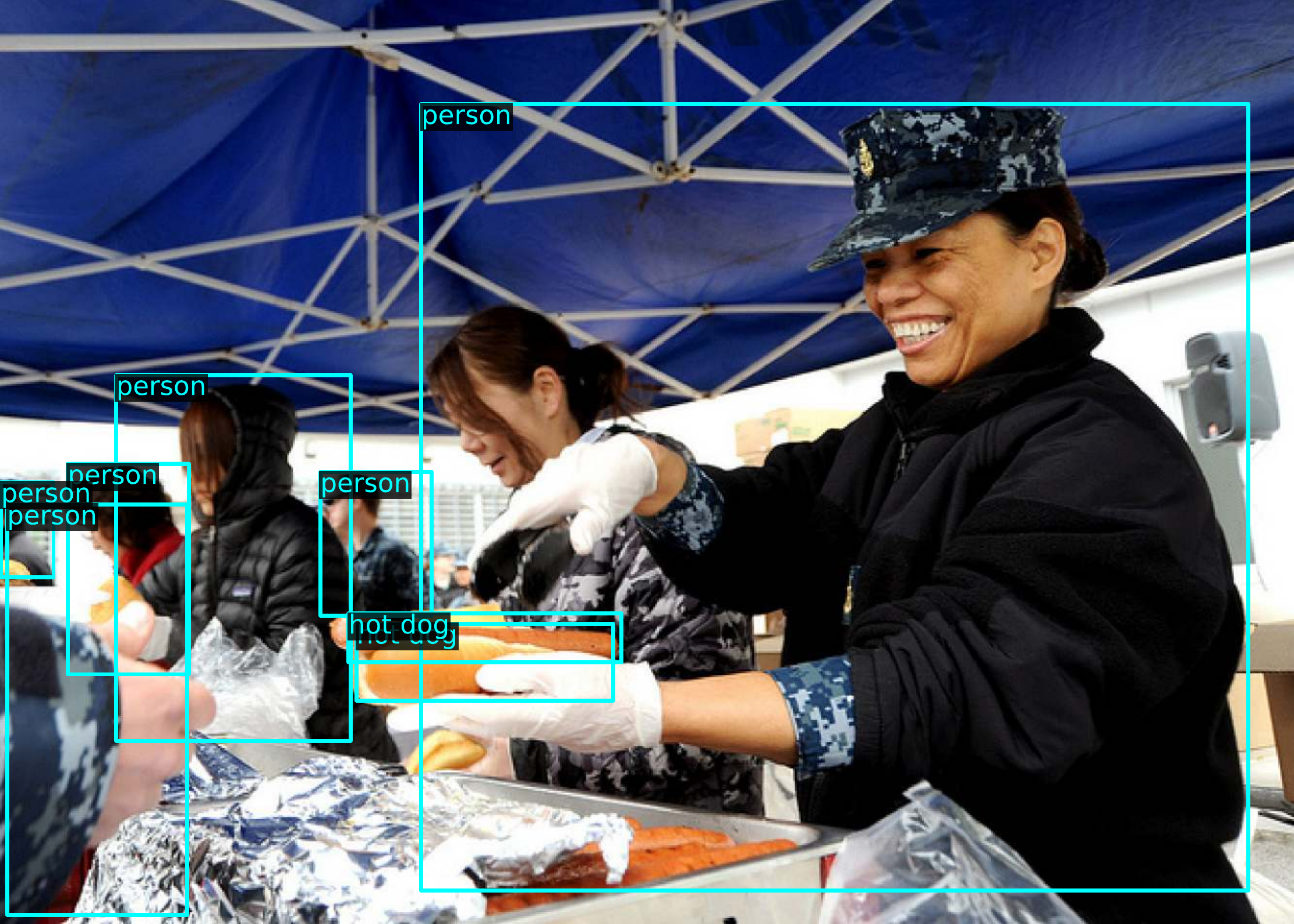}}&
    \makecell{\includegraphics[width=0.255\textwidth]{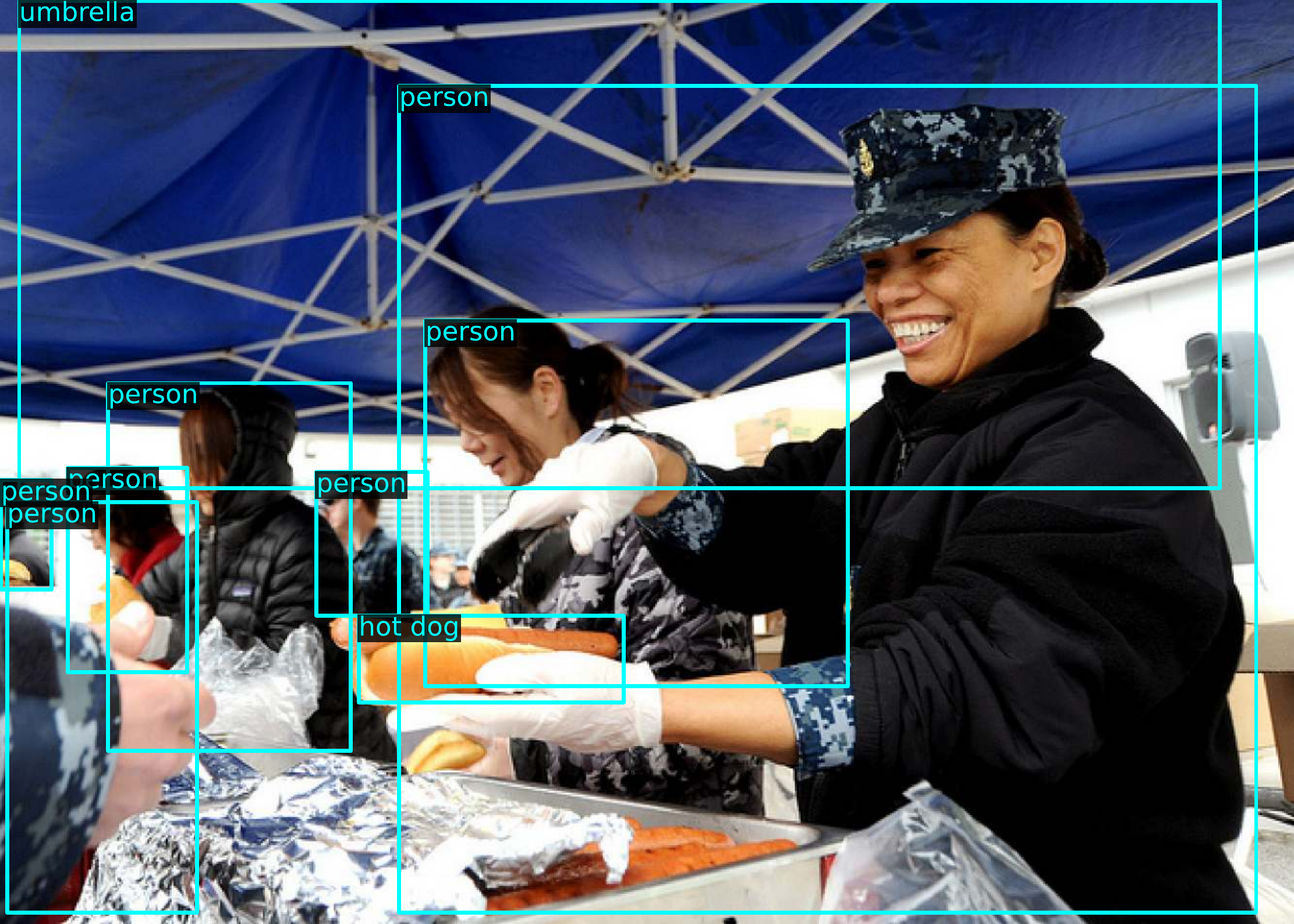}}\\
    \rotatebox[origin=c]{90}{\small 10\% of Labels} \makecell{\includegraphics[width=0.255\textwidth]{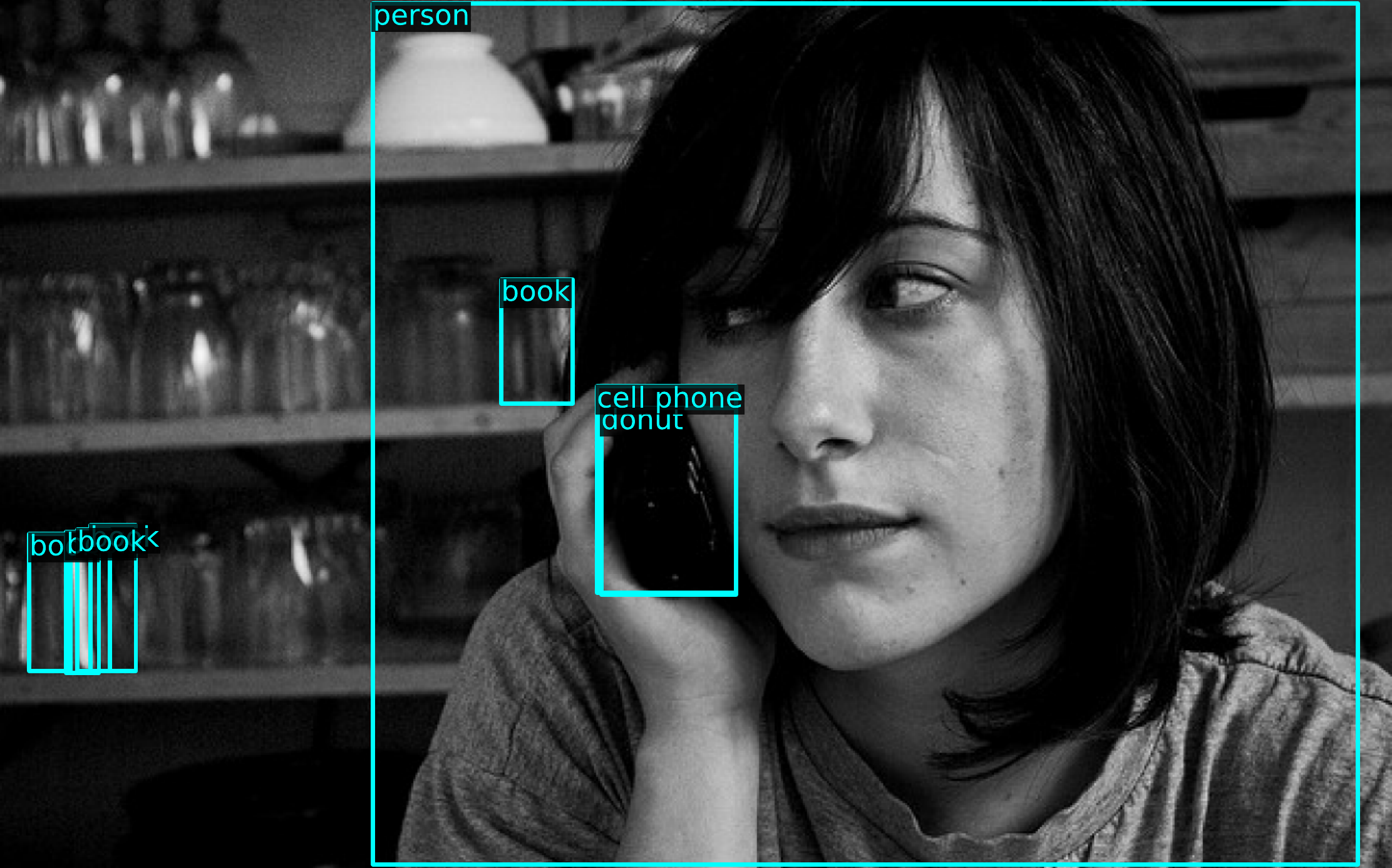}}&
    \makecell{\includegraphics[width=0.255\textwidth]{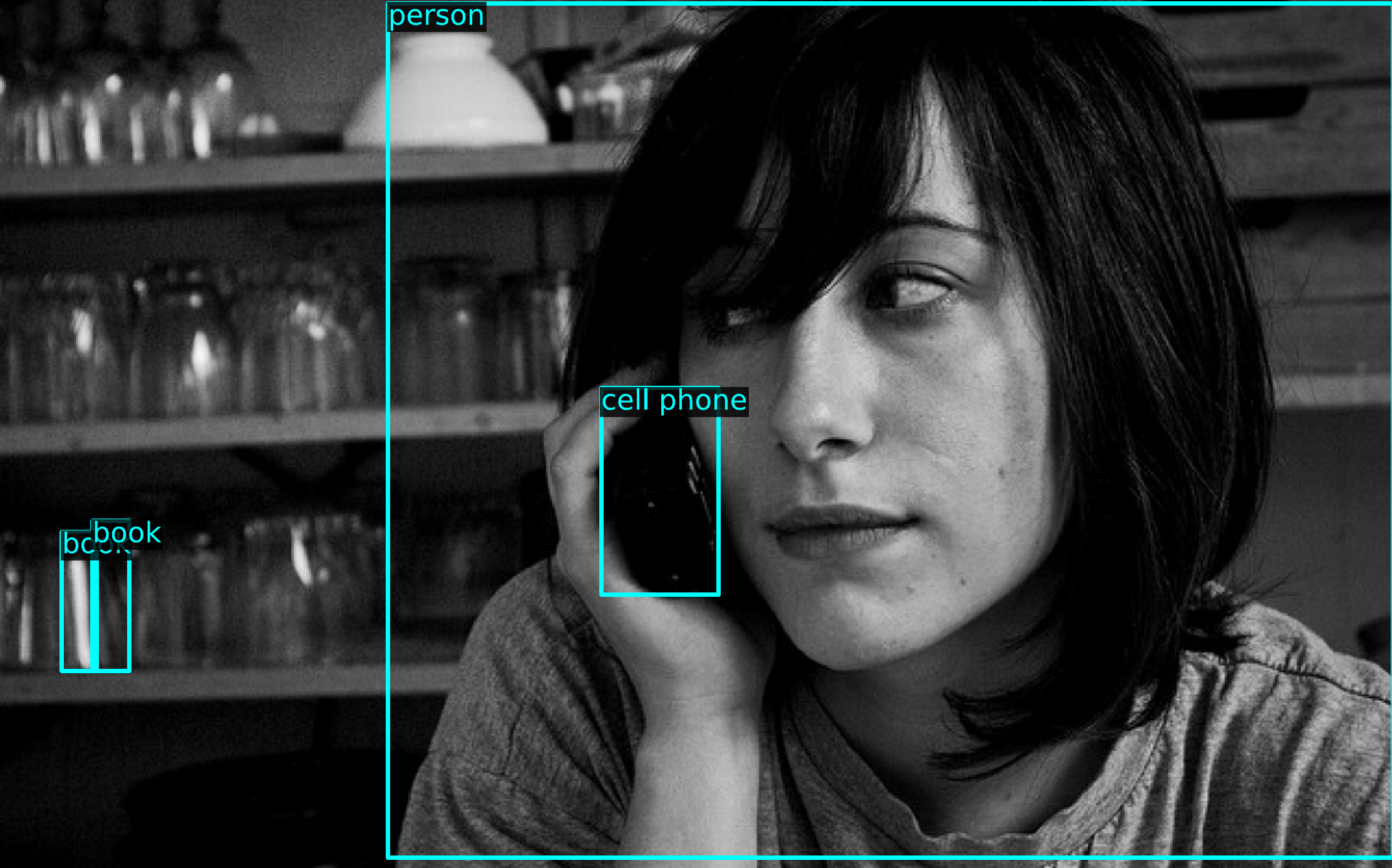}}&
    \makecell{\includegraphics[width=0.255\textwidth]{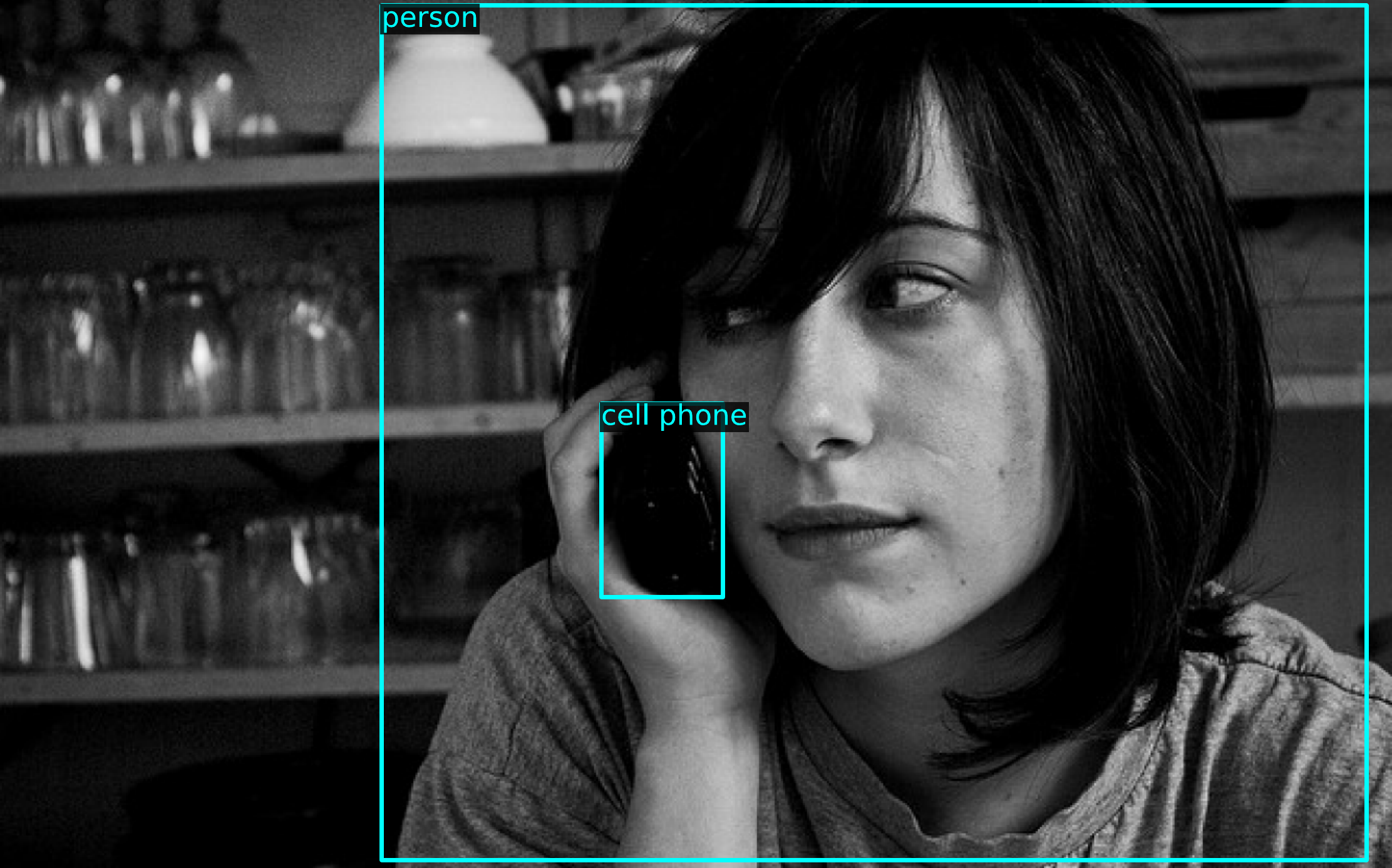}}\\
  \end{tabular}
  \caption{Exemplar detections from models trained on $\{1,5,10\}$ percent of labels sampled from COCO \texttt{train2017} and visualized on \texttt{val2017}. \model captures more object coverage while making fewer false positive detections than its supervised and Soft Teacher counterparts. The enhancements over Soft Teacher are especially pronounced in low-label settings and in crowded scenes with small and ambiguous objects. Best viewed digitally.}
  \label{fig::qualitative-vis-supp}
\end{figure*}

\end{document}